\documentclass[sn-basic,iicol]{sn-jnl} 

\usepackage{graphicx}%
\usepackage{multirow}%
\usepackage{amsmath,amssymb,amsfonts}%
\usepackage{amsthm}%
\usepackage{mathrsfs}%
\usepackage[title]{appendix}%
\usepackage{xcolor}%
\usepackage{textcomp}%
\usepackage{manyfoot}%
\usepackage{booktabs}%
\usepackage{algorithm}%
\usepackage{algorithmicx}%
\usepackage{algpseudocode}%
\usepackage{listings}%
\usepackage{subfigure}%
\usepackage{booktabs}

\usepackage{rotating}
\usepackage{amsmath}%
\usepackage{amssymb}%
\usepackage{caption}%
\usepackage{times} %
\usepackage{helvet} %
\usepackage{courier}  %
\usepackage{tabularray}%
\usepackage{adjustbox}%
\usepackage{hyperref}
\usepackage{natbib}
\usepackage{pdflscape}

\hypersetup{
    colorlinks=true,
    linkcolor=blue,
    filecolor=magenta,      
    urlcolor=black,
    pdftitle={Overleaf Example},
    pdfpagemode=FullScreen,
    }

\theoremstyle{thmstyleone}%
\theoremstyle{thmstyletwo}%

\theoremstyle{thmstylethree}%

\begin{document}

\title[Article Title]{PIE: Physics-inspired Low-light Enhancement}


\author[1,2]{\fnm{Dong} \sur{Liang}}\email{liangdong@nuaa.edu.cn}

\author[1]{\fnm{Zhengyan} \sur{Xu}}\email{xuzhengyan@nuaa.edu.cn}

\author[1]{\fnm{Ling} \sur{Li}}\email{liling@nuaa.edu.cn}

\author*[1]{\fnm{Mingqiang} \sur{Wei}}\email{mingqiang.wei@gmail.com}

\author*[1]{\fnm{Songcan} \sur{Chen}}\email{s.chen@nuaa.edu.cn}

\affil[1]{\orgdiv{MIIT Key Laboratory of Pattern Analysis and Machine Intelligence}, \orgname{College of Computer Science and Technology, Nanjing University of Aeronautics and Astronautics}, \city{Nanjing}, \country{China}}

\affil[2]{\orgdiv{Nanjing Universily of Aeronautics and Astronautics Shenzhen Research Institute}}

\abstract{
In this paper, we propose a physics-inspired contrastive learning paradigm for low-light enhancement, called PIE. 
{PIE primarily addresses three issues:
(i) To resolve the problem of existing learning-based methods often training a LLE model with strict pixel-correspondence image pairs, we eliminate the need for pixel-correspondence paired training data and instead train with unpaired images.
(ii) To address the disregard for negative samples and the inadequacy of their generation in existing methods, we incorporate physics-inspired contrastive learning for LLE and design the Bag of Curves (BoC) method to generate more reasonable negative samples that closely adhere to the underlying physical imaging principle.
(iii) To overcome the reliance on semantic ground truths in existing methods, we propose an unsupervised regional segmentation module, ensuring regional brightness consistency while eliminating the dependency on semantic ground truths.}
Overall, the proposed PIE can effectively learn from unpaired positive/negative samples and smoothly realize non-semantic regional enhancement, which is clearly different from existing LLE efforts. 
Besides the novel architecture of PIE, we explore the gain of PIE on downstream tasks such as semantic segmentation and face detection. 
Training on readily available open data and extensive experiments demonstrate that our method surpasses the state-of-the-art LLE models over six independent cross-scenes datasets. PIE runs fast with reasonable GFLOPs in test time, making it easy to use on mobile devices.
\href{https://github.com/ZhengYanXU/PIE}{\underline{Code available}}}




\maketitle

\newpage
\section{Introduction}
\label{Introduction}
\begin{figure*}[h]
\centering
\subfigure[Input]{
\includegraphics[width=3.9cm]{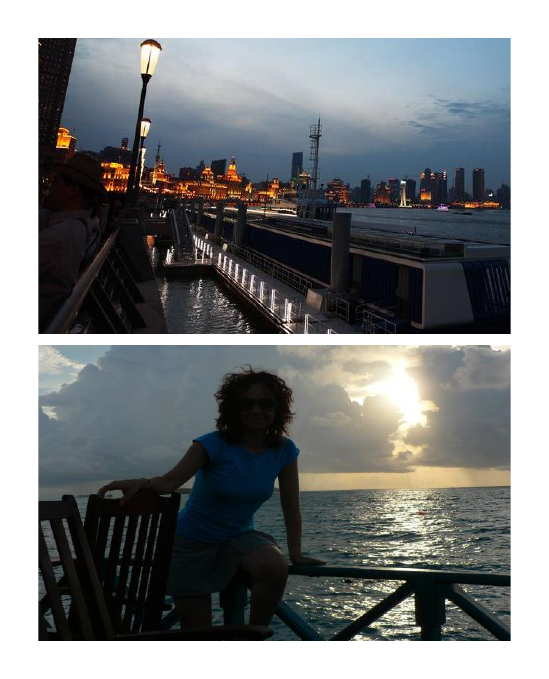}
}
\hspace{-0.3cm}
\subfigure[w/o Neg. samples]{
\includegraphics[width=3.9cm]{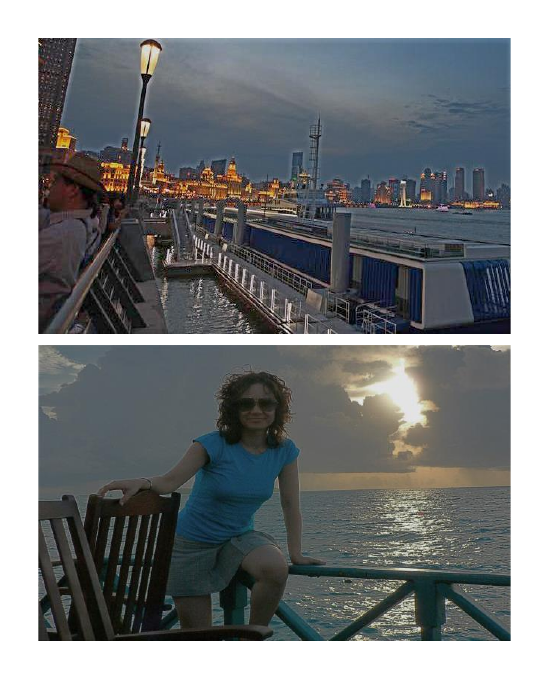}
}
\hspace{-0.3cm}
\subfigure[SCL-LLE \citep{liangaaai}]{
\includegraphics[width=3.9cm]{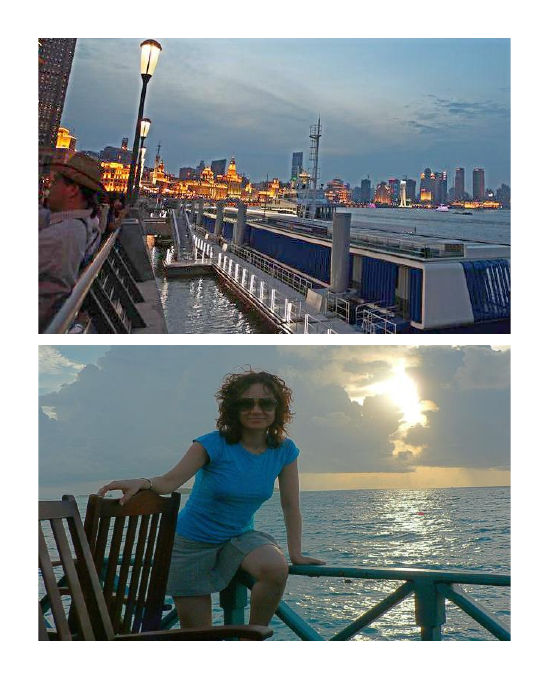}
}
\hspace{-0.3cm}
\subfigure[PIE]{
\includegraphics[width=3.9cm]{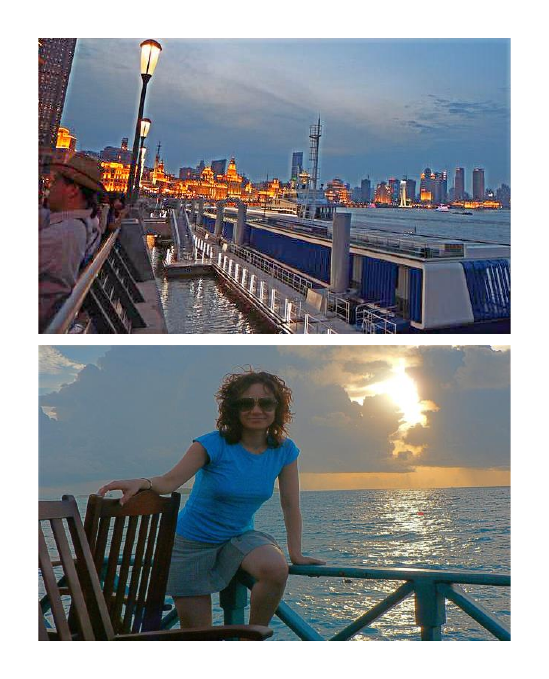}
}
\\
\caption{Impact of training data. The proposed PIE (d), which generates negative samples using physical laws closer to realistic imaging, produces enhanced results with better brightness, color, contrast, and naturalness under extremely dark conditions than SCL-LLE (c) and SCL-LLE without any negative samples (b).
More specifically, the first sample in (d) has a higher dynamic range and better subjective feeling. In the second sample in (d), the saturation of the girl’s T-shirt and the sunset surrounding is much higher and presents a better global stereoscopic atmosphere of the scene. {The comparison between (c) and (d) illustrates the necessity of introducing negative samples. In contrast to (c), the improvement in image quality in (d) reflects the crucial role of negative sample quality in contrastive learning.}
}
\label{img8}
\end{figure*}
Capturing images under low illumination remains a significant source of errors in camera imaging, further leading to image details loss, color under-saturation, low contrast/low dynamic range, and uneven exposure.
Such degeneration severely hinders downstream vision tasks, \emph{e.g.}, semantic segmentation~\citep{wang2022sfnet,cho2020modified,liang2021cross} and object detection~\citep{wu2022edge,al2022comparing,liang2014robust,geng2021dense}, from operating smoothly in vision-based systems.
Existing methods formulate low-light enhancement (LLE) as a mapping problem with three main challenges.

\textbf{First}, the existing learning-based methods in the low-level domain often train a model with strict pixel-correspondence image pairs via strong supervisions~\citep{lore2017llnet, Chen2018Retinex, zhang2019kindling, Xu_2020_CVPR,8692732,ignatov2017dslr,zhou2023improving}. 
However, high-quality pixel-correspondence image pairs are challenging to acquire in practice.
For example, \cite{ignatov2017dslr} proposed to acquire them from a DSLR camera to refine the imaging of a mobile phone camera. It brings complicated registration and pixel-by-pixel calibration to image pairs. 

\begin{figure*}[h]
\centering
\subfigure[Ours]{
\includegraphics[width=4.8cm]{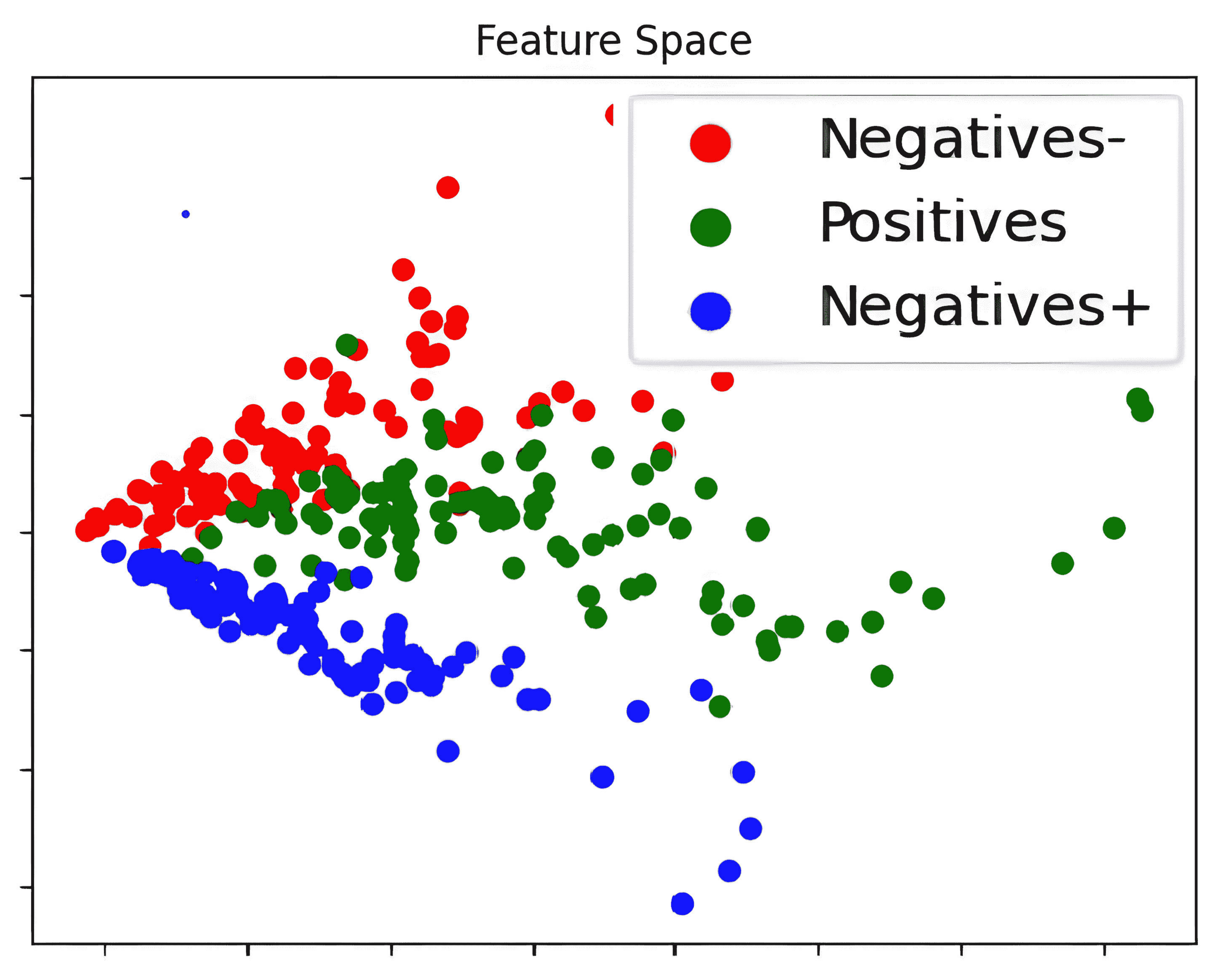}
}
\hspace{-0.3cm}
\subfigure[Manual brightness adjustment]{
\includegraphics[width=4.8cm]{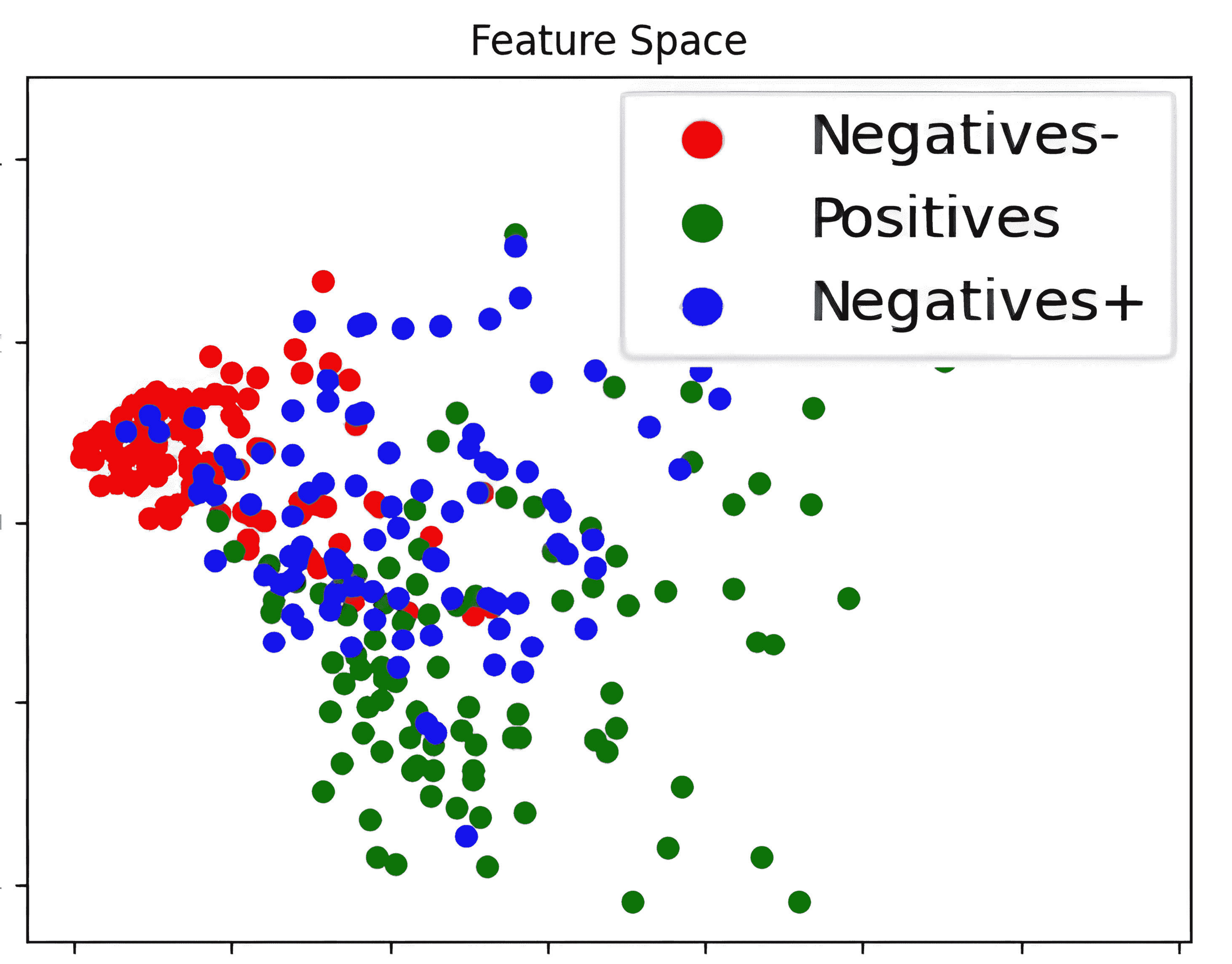}
}
\hspace{-0.3cm}
\subfigure[SICE dataset]{
\includegraphics[width=4.8cm]{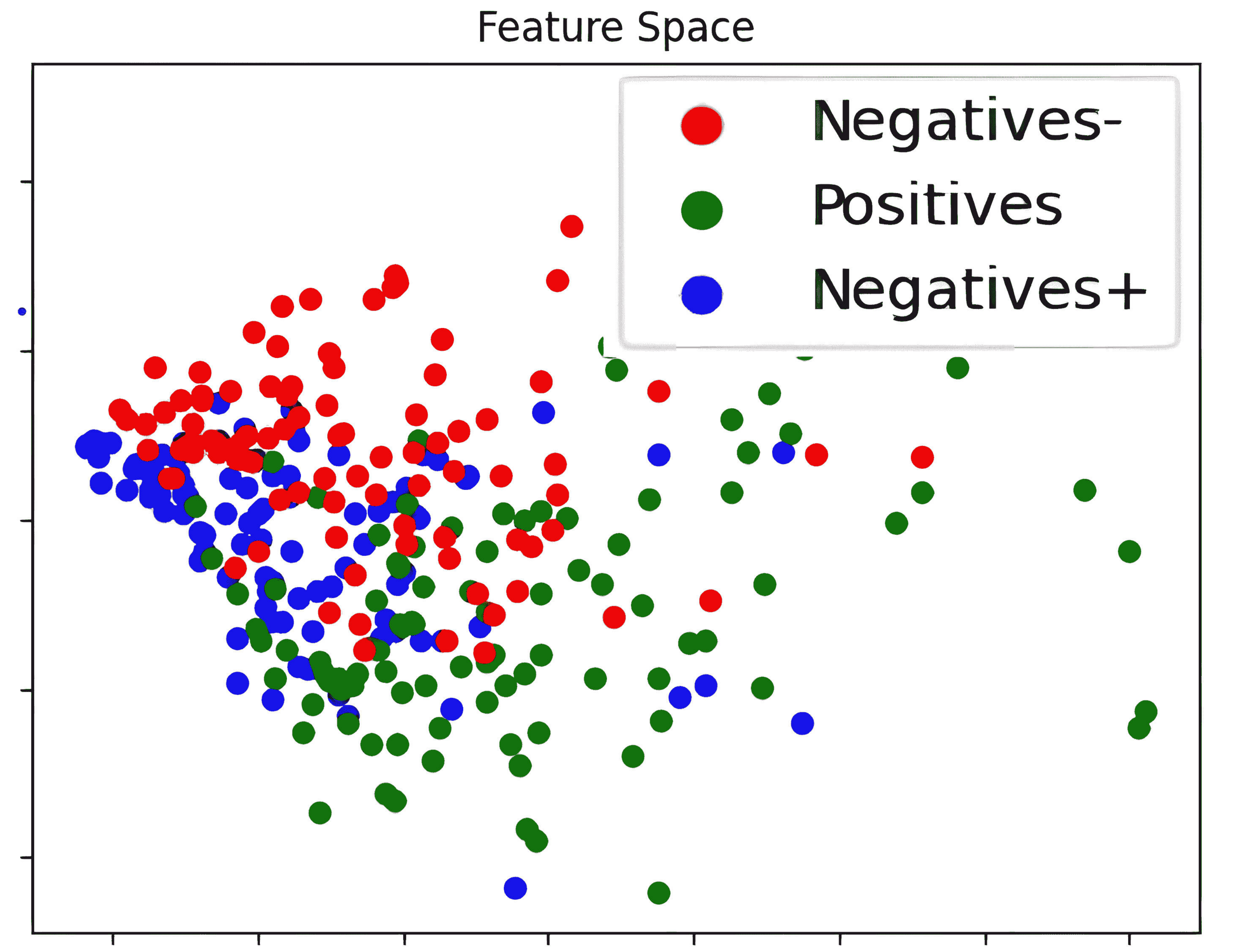}
}

\caption{Feature Visualization of generating negative samples using different methods. Compared to the method of artificially adjusting brightness (b) and low-light dataset (c), our method (a) exhibits a clear boundary between positive and negative samples. The samples in (b) are derived from positive and negative samples in SCL-LLE~\citep{liangaaai}, while the images in (c) are sourced from the SICE~\citep{Cai2018deep} dataset.}
\label{imgneg}
\end{figure*}

\textbf{Second}, to release pixel-correspondence image pairs, some works \citep{huang2023low,shi2022unsupervised} have introduced contrastive learning for LLE,  which adopt the normal-light images as positive samples and the over/underexposed images as negative samples to guide the training. 
As shown in Fig.~\ref{img8}~(b), the selection of negative samples in contrastive learning significantly impacts the results of LLE. 
The quality of negative samples and the specific contrastive learning strategy would be more significant for LLE, if we wish to release the pixel-correspondence image pairs while maintaining consistent performance with the LLE model with the strict image pairs. 
Therefore, another concern of this work is which negative samples to choose for what kind of contrastive learning,
to provide diverse and representative negative samples, squeezing and filling the feature space, enabling the learned LLE model to provide a visual experience closer to the underlying imaging principles. Most existing methods mainly rely on directly using underexposed/overexposed images from existing low-light datasets~\citep{huang2023low} or artificially adjusting image brightness based on empirical experience~\citep{liangaaai} to obtain negative samples. However, due to limitations in the quality of the dataset itself and the constraints of human expertise, the boundaries between negative and positive samples generated by these two methods, as shown in Fig.~\ref{imgneg}~(b) and (c), may not be significant. 

\textbf{Third}, 
the enhancement strategies for the background and foreground should be different. 
In our previous work \citep{liangaaai}, we utilize semantic information to differentiate the enhanced regions and maintain the consistency of brightness within the same semantic category. \cite{wu2023learning} also employs semantic information to maintain consistent brightness for each semantic. However, the introduction of semantic segmentation destroys the universality and flexibility of the method, as semantic segmentation is a full-supervision training setting with massive pixel-level annotation. 

In order to effectively learn from unpaired positive/negative
samples and smoothly realize non-semantic regional enhancement with underlying imaging principles, 
we propose physics-inspired contrastive learning for low-light image enhancement (PIE).  
In contrastive learning, we design the Bag of Curves (BoC) solution by leveraging the Image Signal Processing (ISP) pipeline (\emph{i.e.}, the Gamma correction and Tone mapping) to destroy positive samples but follow the basic imaging rules to generate negative samples. This method generates under/overexposed samples in a way that is more closely aligned with the underlying physical imaging principles.
At the same time, the design of the regional segmentation module maintains regional brightness consistency, realizes region-discriminate enhancement, and releases from semantic labels. 
PIE casts the image enhancement task as a multi-task joint learning problem, where LLE is converted into three constraints -- contrastive learning, regional brightness consistency, and feature preservation, simultaneously ensuring the quality of global/local exposure, texture, and color. 
We also pay more attention to downstream tasks (\emph{i.e.}, semantic segmentation, and face detection) to explore if we can realize performance gain from our LLE scheme. 
We find 
that our method potentially benefits the downstream tasks under dark conditions. 

The contributions are three folds: 
\begin{itemize}
    \item [$\bullet$] 
    A physics-inspired contrastive learning approach for real-world cross-scene LLE, without any paired training images and pixel-level annotation:
    
    1) a physics-inspired approach called “Bag of Curves" generates negative samples for contrastive learning using principles closer to the underlying physical imaging mechanism. 
    
    2) an unsupervised regional segmentation module to maintain regional brightness consistency, realize region-discriminate enhancement, and release from semantic labels.

    3) a multi-task joint learning with three constraints -- contrastive learning, regional brightness consistency, and feature preservation, simultaneously ensuring exposure, texture, and color consistency. 
   
    \item [$\bullet$] PIE is compared with SOTAs via comprehensive experiments on six independent datasets in terms of visual quality, no and full-referenced image quality assessment, and human subjective survey. All results consistently endorse the superiority and efficiency of the proposed approach.
    
    \item [$\bullet$] We demonstrate that our PIE is friendly to downstream high-level vision tasks and easy to joint-learn with them.
\end{itemize}

This work is partially presented in our earlier conference version~\citep{liangaaai}. We have introduced many new findings and improvements compared to the conference version. We have two new core contributions. First, we design a ``Bag of Curves" solution inspired by the physical imaging principle to generate negative samples to replace the manual process. Second, we design an unsupervised regional segmentation module to maintain regional brightness consistency to replace the supervised semantic segmentation module. We also modify our contrastive learning loss for better performance, present more extensive experiments related to the aforementioned improvements compared with more recent methods, and provide more discussion with downstream tasks.

\section{Related Work}
\label{Relatedwork}
\subsection{Low-light Image Enhancement}
\textbf{Conventional Methods.} LLE has been actively studied as an image-processing problem for a long. Early efforts are commonly made towards the use of handcrafted priors with empirical observations~\citep{Pizer1990ContrastlimitedAH,land1977retinex,xu2014novel,guo2016lime} to deal with the LLE problem. 
Histogram equalization~\citep{Pizer1990ContrastlimitedAH} used a cumulative distribution function to regularize the image's pixel values and evenly distribute overall intensity levels. However, this kind of operation naturally makes it easy to cause over/under-exposure. Without local adaptation, the enhancement results in intensive noise and undesirable illumination. Later methods constrained the equalization process with several kinds of priors, \emph{e.g.} mean intensity preservation~\citep{2007Brightness}, noise robustness, and white and black stretching~\citep{2009A}, to improve the overall visual quality of the adjusted image. Retinex model~\citep{land1977retinex} and its multi-scale version~\citep{Jobson1997A} decomposed the brightness into illumination and reflectance and then processed them separately. \cite{Shuhang2013Naturalness} constructed a brightness filter for Retinex decomposition and tried to preserve the naturalness while enhancing details in low-light images. The reflectance component is commonly assumed to be consistent under lighting conditions; thus, light enhancement is formulated as an illumination estimation problem.
The gray-scale transformation~\citep{xu2014novel} is a method based on the spatial domain, which enhanced the image by modifying the distribution and dynamic range of the gray-scale value of the pixels. 
\cite{guo2016lime} introduced a structural prior to refining the initial illumination map and finally synthesized the enhanced image according to the Retinex theory.
However, these handcrafted constraints/priors are not self-adaptive to recover image details and color. This results in washing out details, local under/over-saturation, uneven exposure, or halo artifacts.
\\
\textbf{Data-Driven Methods.} In the past decade, data-driven methods~\citep{li2021low} have achieved significant advancements in the field of low-light image enhancement. \cite{lore2017llnet} proposed a variant stacked sparse denoising autoencoder to enhance the degraded images. RetinexNet~\citep{Chen2018Retinex} leveraged a deep architecture based on Retinex to enhance low-light images. \cite{zhang2019kindling} developed three subnetworks for layer decomposition, reflectance restoration, and illumination adjustment based on Retinex. RUAS~\citep{liu2021retinex} constructed the overall LLE network architecture by unfolding its optimization process. 
The above methods are trained based on image pairs with strict pixel correspondence. \cite{zhou2022lednet} introduced LEDNet, a powerful network designed for simultaneous low-light enhancement and deblurring tasks.
\cite{jiang2021enlightengan} reported an unsupervised method using normal-light images that do not have low-light images as correspondences. Zero-DCE~\citep{Guo_2020_CVPR}, FlexiCurve~\citep{li2023flexicurve}, CuDi~\citep{li2022cudi} and ReLLIE~\citep{Zhang2021ReLLIEDR} reformulated the LLE task as an image-specific curve estimation problem with a fixed default brightness value. 
\cite{FanWY020} used semantic information to guide the reconstruction of the reflection of Retinex. DNF~\citep{jin2023dnf} is a decouple and feedback framework for the RAW-based LLIE. CLIP-LIT~\citep{liang2023iterative} introduced an initial prompt pair, enforcing text prompt and backlit image similarity using CLIP latent space. SCL-LLE~\citep{liangaaai} introduced semantic information to the brightness reconstruction and paid more attention to the dependency among the semantic elements via the interaction of high-level semantic knowledge and low-level signal priors. The proposed PIE maintains the brightness consistency of image regions without relying on semantic ground truths, which is clearly different from existing LLE efforts. 
\subsection{Contrastive Learning for Vision Tasks}
Contrastive learning~\citep{he2020momentum, chen2020simple, sermanet2018time, tian2020contrastive, henaff2020data} is from the self-supervised learning paradigm, which is characterized by using pretext tasks to mine its supervisory information from original data for downstream tasks. 
For a given input, contrastive learning aims to pull it together with the positives and push it apart from the negatives in feature space. 
Previous works have applied contrastive learning to high-level vision tasks because these tasks are inherently suited for modeling the contrast~\citep{he2020momentum,chen2020simple,tian2020contrastive} between positive and negative samples. 
It has also been applied to low-level visual tasks, such as deraining~\citep{chen2022unpaired}, underwater image enhancement~\citep{han2021single}, and dehazing~\citep{wu2021contrastive}. 
\cite{huang2023low,shi2022unsupervised} introduced a contrastive learning module for low-light enhancement. \cite{huang2023low} employed contrastive learning to train a two-stream encoder for feature extraction. \cite{shi2022unsupervised}  used contrastive learning techniques to train the SFE model for extracting structure maps. However, these methods overlooked the importance of the way to select positive and negative samples in contrastive learning.
In addition, most of the existing contrastive learning methods rely heavily on a large number of negative samples and thus require either large batches or memory banks~\citep{li2021triplet}. In our approach, we employ only a couple of negative samples for one positive sample and introduce a random mapping strategy to avoid the risk of overfitting. 
\subsection{Gamma Correction \& Tone Mapping}
The Image Signal Processing (ISP) pipeline is used in modern digital cameras to convert raw camera sensor data into high-quality, human-readable RGB images. 
ISP ~\citep{karaimer2016software} consists of several operations, including image denoising, noise reduction, white balance, color space conversion, Gamma correction, and Tone mapping. 

Gamma correction~\citep{farid2001blind, yuan2012automatic} is a standard step of the image processing pipeline to adjust the brightness of an image for display on different devices. It transformed the pixel value following a  non-linear power-law function.
Tone mapping~\citep{mantiuk2008display} refers to the process of converting high dynamic range (HDR) images to low dynamic range (LDR) images. HDR images in the RAW domain have higher color depth and dynamic range than LDR images, allowing them to better represent the brightness and color details in a scene, but they cannot be fully displayed on conventional RGB displays.  Therefore, Tone mapping is needed for HDR images to present as much HDR image information as possible on LDR displays. \cite{10074176} combined Tone mapping with GAN to adjust the brightness of images. \cite{drago2003adaptive} and \cite{yongqing2013dodging} respectively used two different Tone mapping methods based on different curves to adjust the brightness of the image. Inspired by Gamma correction and Tone mapping, we propose a physics-inspired contrastive learning method and introduce “Bag of Curves” to generate negatives for contrastive learning. 
\begin{figure*}[!htb]
\centering
\includegraphics[width=.95\textwidth]{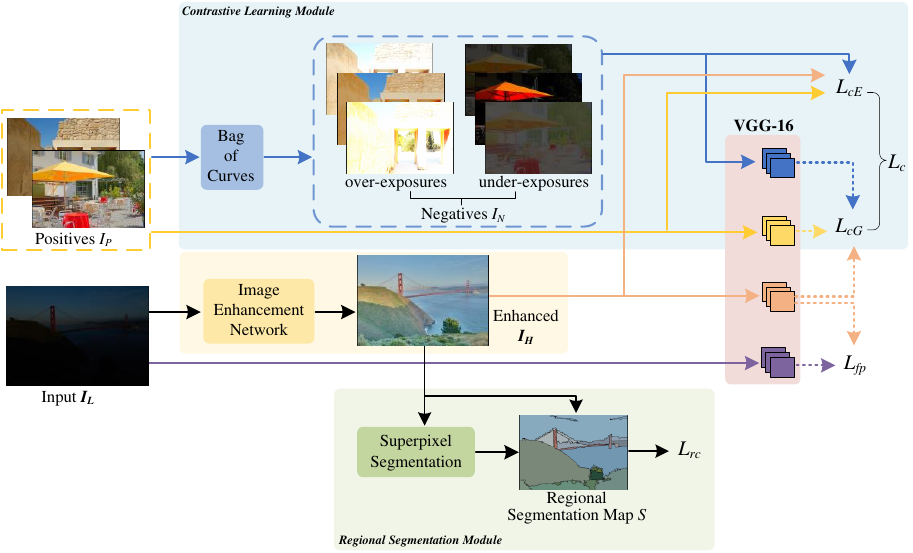}
\caption{The overall architecture of PIE. It includes a low-light image enhancement (LLE) network, a contrastive learning module (the blue block) boosted by Bag of Curves (BoC), a regional segmentation module (the green block), and a VGG-16 feature extractor (the red block). 
PIE jointly minimizes the contrastive learning loss $L_{c}$, which consists of two components, $L_{cE}$ and $L_{cG}$, feature preserving loss $L_{fp}$, and regional brightness consistency loss $L_{rc}$.}
\label{Net}
\end{figure*}
\section{The proposed PIE}
\label{Proposed}
\subsection{Problem Formulation \& Architecture}
Fundamentally, low-light image enhancement can be regarded as seeking a mapping function ${F}$, such that $I_H = {F}(I_L)$ is the desired image, which is enhanced from the input image $I_L$. In our design, we introduce a prior of contrastive learning: the contrastive samples, including the negatives $I_N$, \emph{i.e.}, the under/overexposed images which are generated by our proposed Bag of Curves solution, and the positives $I_P$, \emph{i.e.}, the normal-light images. Therefore, we formulate a new mapping function as follows: 
\begin{equation}
    I_H = {F} (I_L, I_N, I_P)
\end{equation}

As depicted in Fig.~\ref{Net}, PIE consists of a low-light image enhancement network, a contrastive learning module, and a regional segmentation module. Specifically, our approach comprises a low-light image enhancement network, which leverages a U-Net-like backbone~\citep{Guo_2020_CVPR} to generate a pixel correction curve that remaps each pixel. We use VGG-16~\citep{simonyan2014very} as the feature extraction network. Specifically, for a given image $I_L$, it is first input to the image enhancement network. Then, the enhanced image $I_H$ is fed into the regional segmentation module, ensuring brightness consistency within each region. For the contrastive learning module, images enhanced by image enhancement network $I_H$ serve as the anchor for contrastive learning, images under normal lighting $I_P$ serve as positive samples, and negative samples are over/underexposed images $I_N$ obtained from images under normal lighting through Bag of Curves. To optimize our approach, we employ three types of losses corresponding to the framework's three key aspects: the contrastive learning loss $L_{c}$, feature preserving loss $L_{fp}$, and regional brightness consistency loss $L_{rc}$.

In PIE, we solve the challenges of manually dividing positive and negative samples in contrastive learning as on~\citep{liangaaai} and eliminate the dependency on semantic ground truths. 
Specifically, we first propose a ``Bag of Curves" method that combines the physical imaging principle with contrastive learning to generate negative samples, which aids in compressing the feature space and enabling the model to effectively adjust the distance between the anchor and positive/negative samples. Additionally, we introduce an unsupervised regional segmentation module that maintains regional brightness consistency while removing the reliance on semantic ground truths. 
 \subsection{Contrastive Learning Module}
\subsubsection{Bag of Curves}
\begin{figure*}[h]
\centering
\subfigure{
\includegraphics[width=160mm]{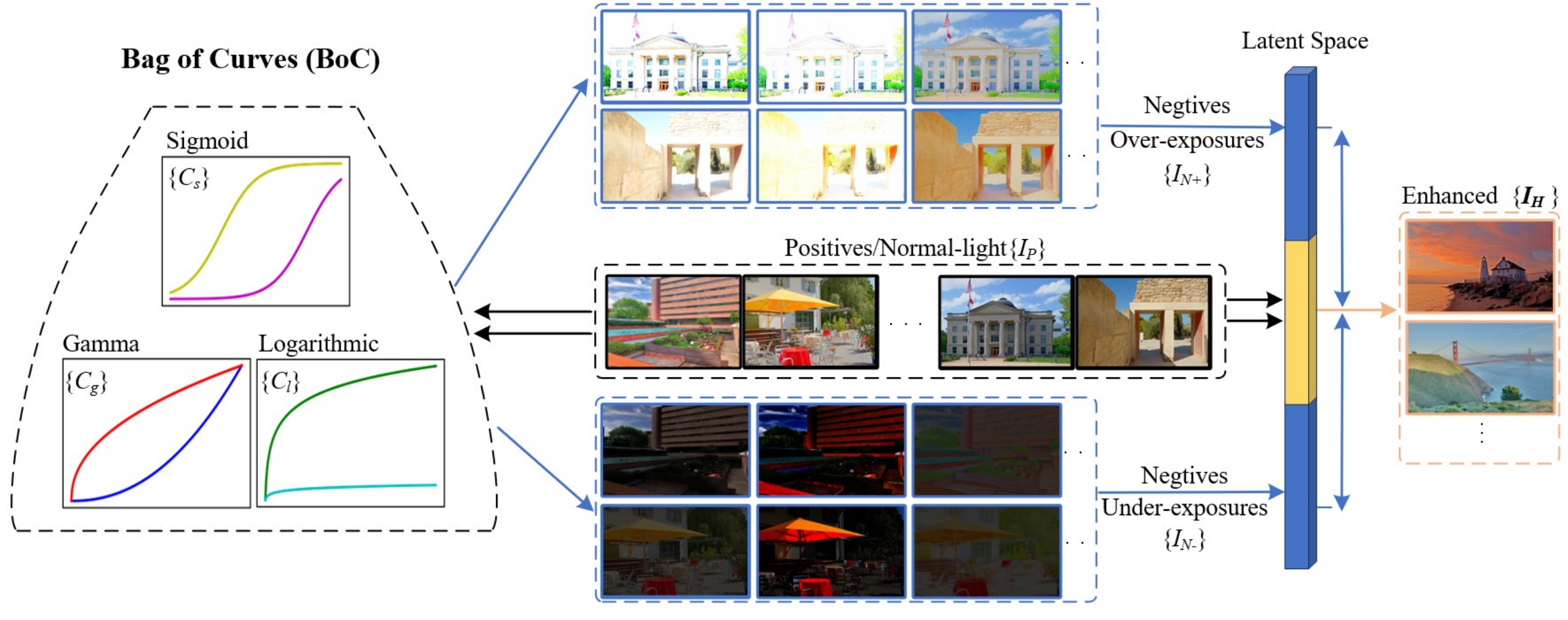}
}
\caption{Bag of Curves. 
There are three different groups of curves: Gamma, Sigmoid, and 
Logarithmic curves enable the model to learn diverse and representative characteristics of the produced negative samples. }
\vspace{-0.3CM}
\label{curves}
\end{figure*}

Choosing appropriate negative samples is crucial for the success of contrastive learning, as it enables the model to learn sample representations that capture the unique characteristics of the data. 
For an LLE task, the construction of negative samples in contrastive learning should better follow the physical laws of imaging as closely as possible and mimic both overexposure and underexposure. 
Our previous work~\citep{liangaaai} has achieved non-paired contrastive learning, but the positive and negative samples used are still provided by the manual process -- manually adjusting the brightness of images to generate a set of under/overexposed negatives. 

We leverage Tone mapping and Gamma correction in ISP for brightness adjustment and utilize this prior knowledge to adjust the brightness of images to generate negative samples that are more consistent with physical imaging laws. We follow the following principles when choosing curves: 1) It should be able to effectively adjust the overall brightness of the image in a reasonable manner. 2) The form of the curve should be as simple as possible for ease of implementation and computation. 3) Priority is given to commonly used curves in existing methods. Taking into account these reasons, we choose Gamma, Logarithmic, and Sigmoid curves to simulate the inverse Gamma correction and Tone mapping process for adjusting the brightness of the image.

\begin{figure*}[h]
\centering
\subfigure[Input]{
\includegraphics[width=2.2cm]{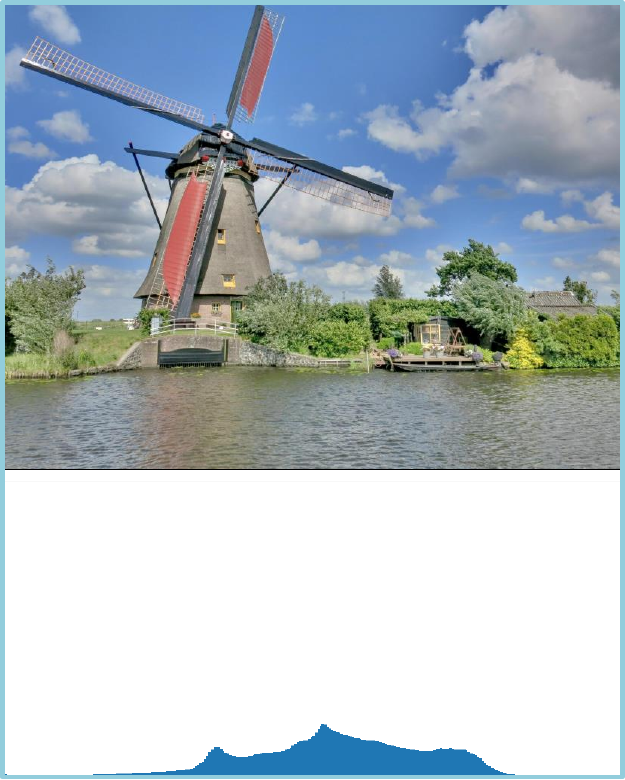}
}
\hspace{-0.3cm}
\subfigure[Gamma (-)]{
\includegraphics[width=2.2cm]{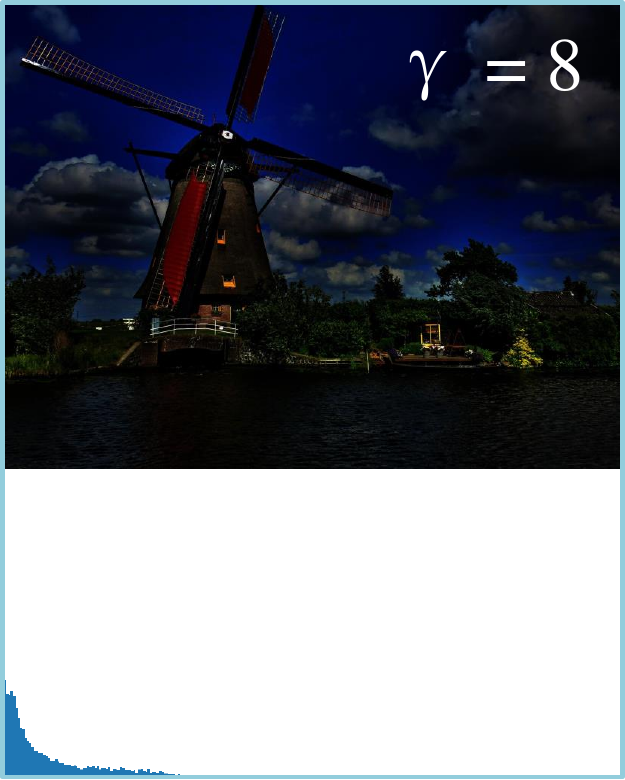}
}
\hspace{-0.3cm}
\subfigure[Gamma (+)]{
\includegraphics[width=2.2cm]{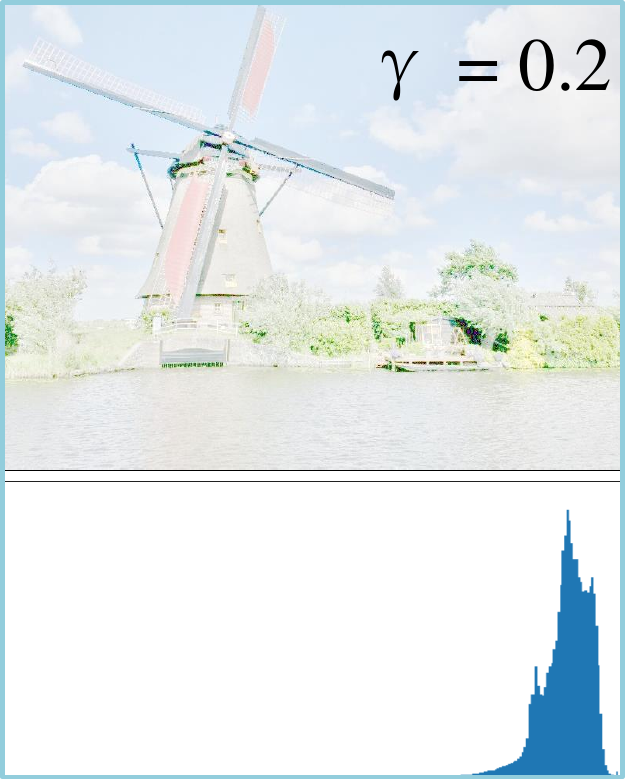}
}
\hspace{-0.3cm}
\subfigure[Sigmoid (-)]{
\includegraphics[width=2.2cm]{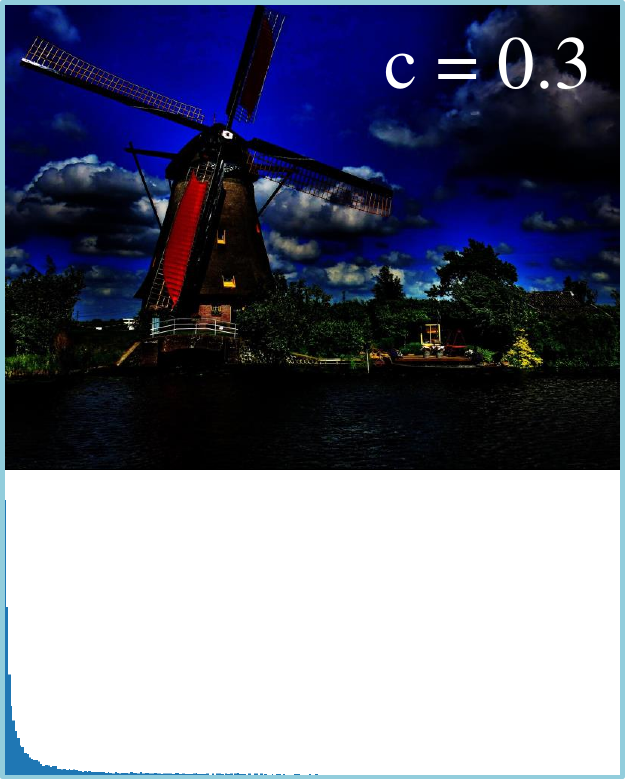}
}
\hspace{-0.3cm}
\subfigure[Sigmoid (+)]{
\includegraphics[width=2.2cm]{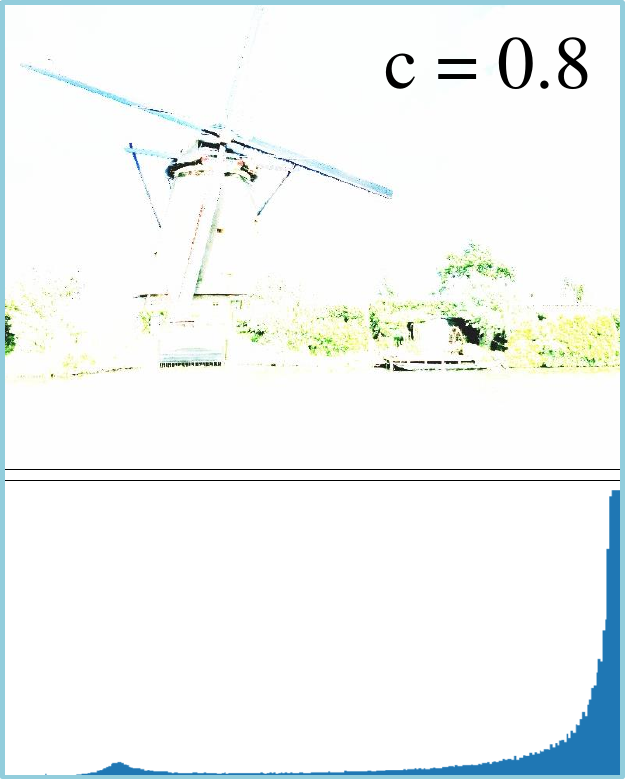}
}
\hspace{-0.3cm}
\subfigure[Logarithmic (-)]{
\includegraphics[width=2.2cm]{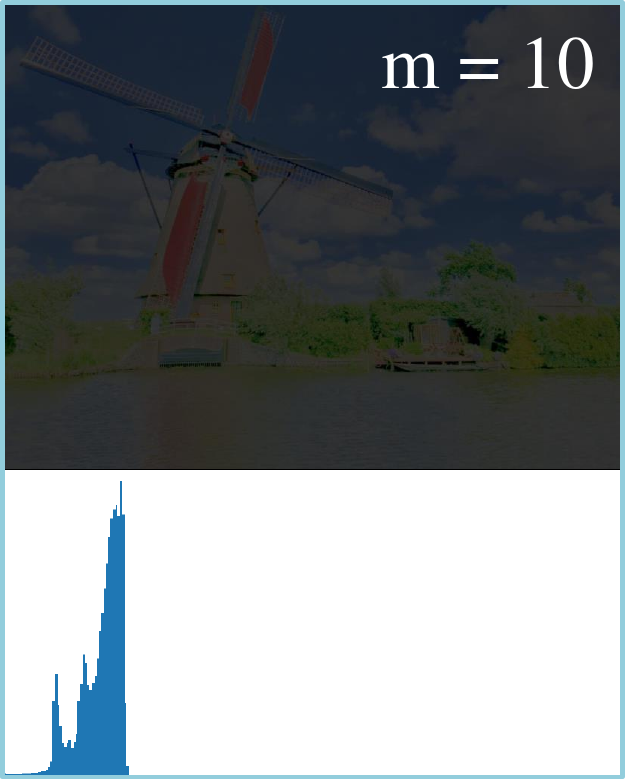}
}
\hspace{-0.3cm}
\subfigure[Logarithmic (+)]{
\includegraphics[width=2.2cm]{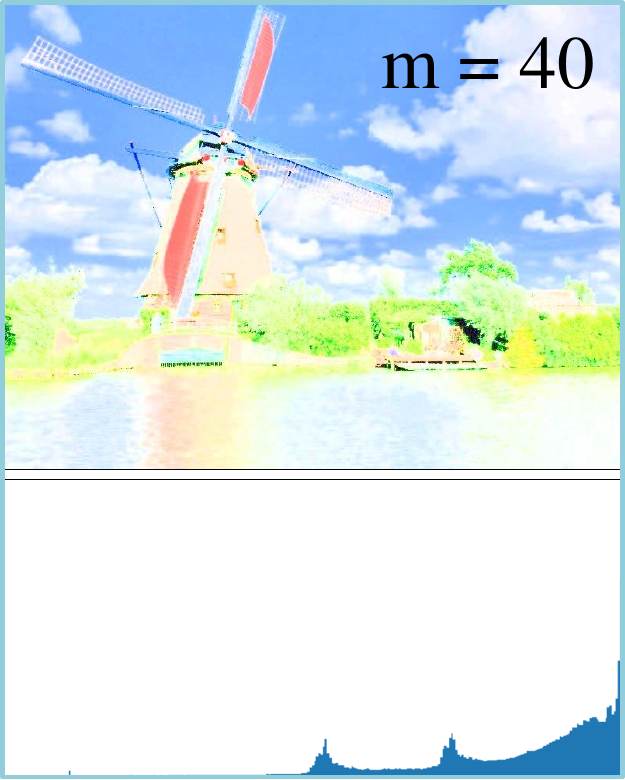}
}
\caption{BoC samples and their histogram from normal lighting sample (positives) to negatives, using Gamma, Sigmoid, and Logarithmic curves, respectively. These three curves effectively make the brightness of the normal illumination image distributed in the under (-)/overexposed (+) areas of the histograms. The (-)/(+) samples have quite different histograms and appearances but follow the physical imaging laws, making the negative samples effective for contrastive learning.
}
\label{img_gamma}
\end{figure*}

Due to the different brightness ranges captured by the human eyes and digital cameras, the brightness and color captured by cameras (when viewed on a standard monitor) look different from what the human eye perceives. 
When rendering high dynamic range (HDR) images, the brightness values can exceed the maximum value that a monitor can show. Therefore, we need to adjust the brightness range of the image to convert HDR images to low dynamic range (LDR) that can be appropriately displayed on a monitor. 
The process of adjusting the brightness of an image is commonly referred to as Tone mapping. 
After Tone mapping, Gamma correction is usually performed to account for humans' non-linear perception of natural brightness and to adapt to the monitor's display characteristics, ultimately outputting the corresponding brightness to the display.
Tone mapping~\citep{mantiuk2008display} and Gamma correction~\citep{farid2001blind} are commonly used on physical devices such as cameras and monitors to adjust the brightness of images, resulting in a photo effect that is more similar to human perception. 
Both Tone mapping and Gamma correction are intended to improve the display of images on LDR devices by transforming the range of brightness values from one distribution to another. Inspired by the above observation, 
and also inspired by Bag of Words (BoW)~\citep{sivic2003video} in feature engineering, we propose Bag of Curves (BoC) to generate negative samples for contrastive learning. 
Specifically, we leverage standard curves in Tone mapping and Gamma correction to realize reversed gone mapping and reversed Gamma correction. This way maps the brightness of $\{I_{p}\}$ to a certain range to generate over/underexposed images as negative samples. This range could destroy the original brightness of $\{I_{p}\}$ but could follow the physical imaging principle.
\\
\begin{equation}
\begin{aligned}
I_{N} \in BoC = \{C_g,C_s,C_l\}
\end{aligned}
\end{equation}
In BoC, we select three curves, the Gamma curve $C_g$ from Gamma correction, the Sigmoid curve $C_s$, and the logarithmic curve $C_l$  in Tone mapping to generate over/underexposed images as negative samples for contrastive learning, which are directly and parallel functioned with the positives ${I_P}$. The curves representation of BoC are shown in Fig.~\ref{curves}. 
\\
\textbf{Gamma curve} 
The Gamma curve $C_g$ is defined as follows:

\begin{equation}
\begin{aligned}
\{C_g\} = \{I_{p}^{ \gamma }\}
\end{aligned}
\end{equation}
A Gamma value $\gamma$  $\textless$ 1 is sometimes called an encoding Gamma, and the process of encoding with this compressive power-law nonlinearity is called Gamma compression; conversely, a Gamma value $\gamma$ $\textgreater$ 1 is called a decoding Gamma, and the application of the expansive power-law nonlinearity is called Gamma expansion. We set $\gamma$ = 0.2 to generate overexposed images and $\gamma$ = 8 to generate underexposed images. 
\textbf{Sigmoid curve} 
The Sigmoid curve $C_s$ is a nonlinear function curve.
\cite{yongqing2013dodging} succeeded in expanding the local dynamic range in dark and bright areas by dodging and burning with the Sigmoid curve operator. 
The formula for the Sigmoid curve $C_s$ is as follows: 
\begin{equation}
{\small
\begin{aligned}
\{C_s\} =  \{\frac{1}{1 + e^{10\cdot (c - I_{p}) } } 	\}
\end{aligned}
}
\end{equation}
where $c$ represents the offset of the Sigmoid curve. By adjusting the values of $c$, the shape of the Sigmoid function can be controlled to achieve different brightness adjustment effects. For underexposure, the value of $c$ can be set to a value smaller than the brightness of $I_{p}$, and 0.3 is selected in our setting. 
Similarly, for overexposure, the value of $c$ is 0.8.
\\
\textbf{Logarithmic curve} 
The Logarithmic curve $C_l$ is approaching HVS's perception of brightness~\citep{drago2003adaptive}. The formula for the Logarithmic curve $C_l$ is as follows:
\begin{equation}
\begin{aligned}
\{C_l\} = \{m \cdot \log_{2}{ (1 + I_{P})}	\}
\end{aligned}
\end{equation}
The smaller the value of constant $m$, the weaker the effect of Logarithmic transformation and the lower the brightness of the image. We set $m$ = 10 and $m$ = 40 to generate under/overexposed negatives. 

As shown in Fig.~\ref{img_gamma}, the three curves can adjust the brightness of the image to guide its brightness distribution on the highlight or low-light side of the brightness histogram in quite different forms, which could be effectively negative samples for contrastive feature learning. 
\begin{figure}[h]
\centering
\subfigure[BoC with fixed parameters]{
\centering
\includegraphics[width=6.5cm]{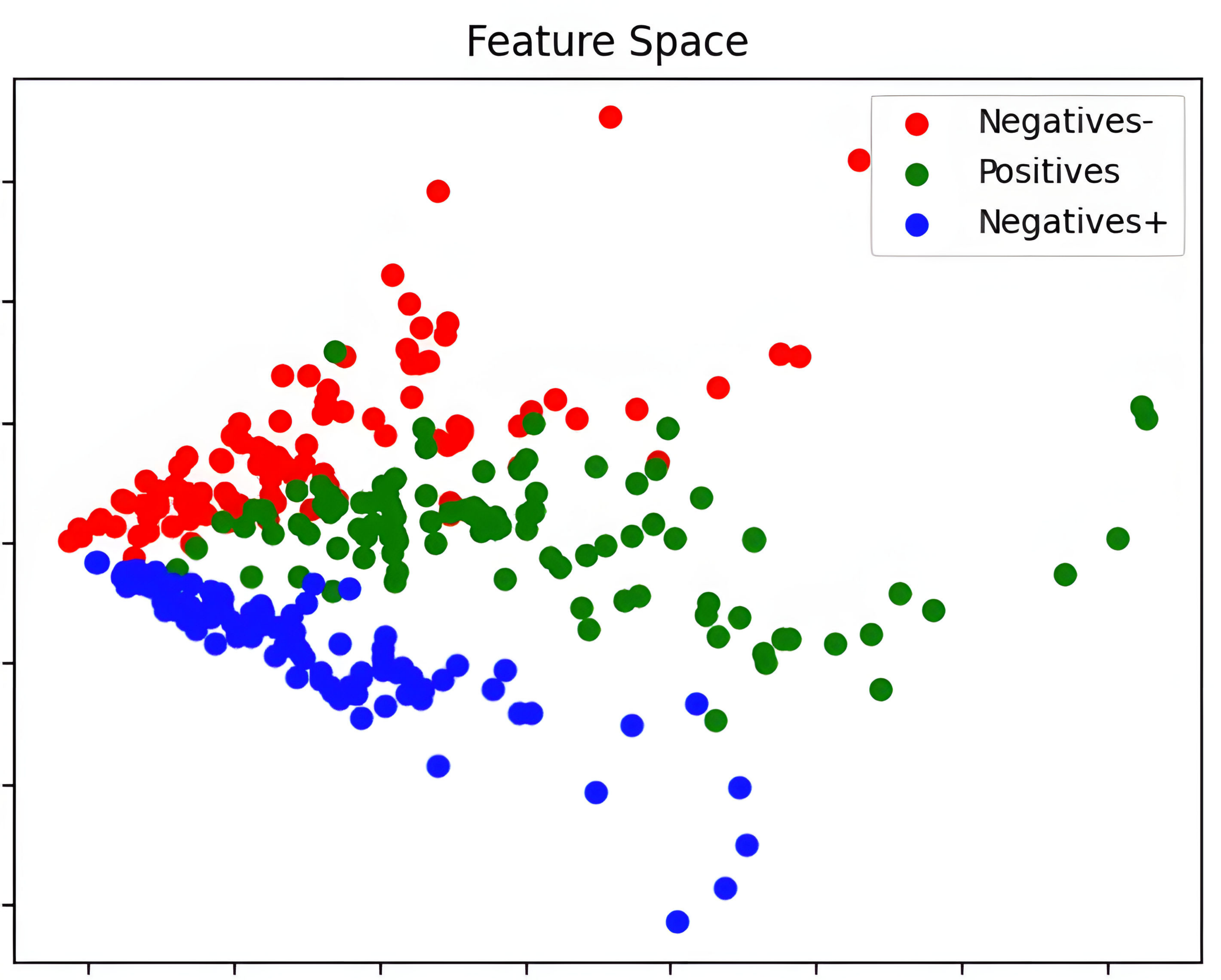}
}
\vspace{-0.2cm}
\subfigure[BoC with parameters within a certain range]{
\centering
\includegraphics[width=6.5cm]{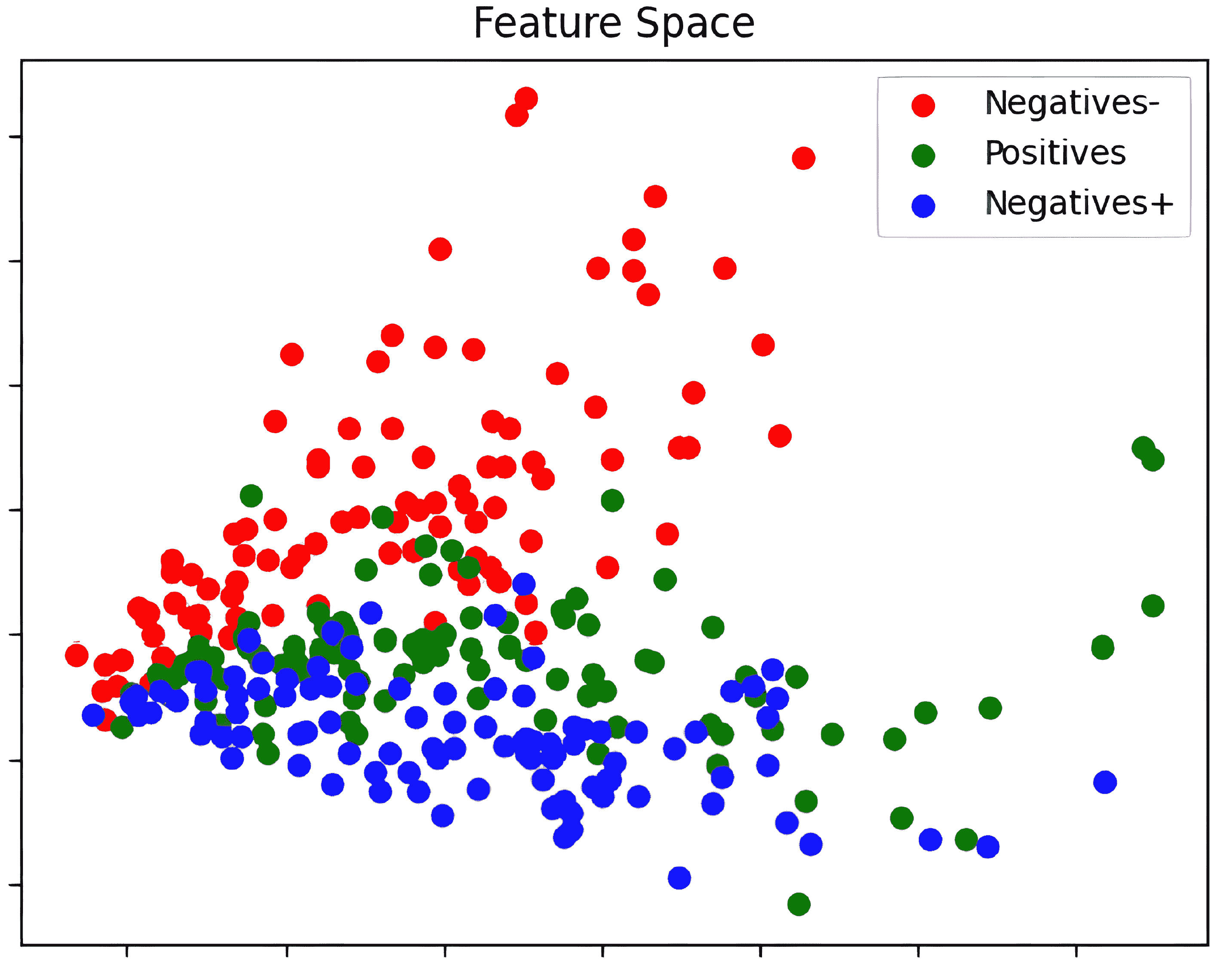}
}
\caption{Visualization of features. In BoC, the three curves included in the paper have fixed parameter values in (a), while in (b), they randomly fall within a certain range. In (a), the boundary between positive and negative samples is more pronounced, indicating a clearer separation between the two classes.}
\label{const}
\end{figure}

In contrastive learning, it is necessary to ensure a clear boundary between negative samples and positive samples in the feature space. Negative samples should also be as clustered as possible in the feature space. Therefore, generating representative negative samples is considered crucial. We achieve this by using fixed values to generate underexposed or overexposed negative samples that represent the entire sample category.

From Fig.~\ref{const}, it can be observed that when the parameter values (\emph{e.g.}, $\gamma$, $c$, $m$) is fixed, the boundary between the normal illumination image (Positives) and the underexposed negative sample (Negatives-) and overexposed negative sample (Negatives+) is more distinct. This results in higher discrimination in the feature space and effective compression of the feature space. However, when the parameter value falls within a certain range, some negative samples may overlap with the normal illumination image in the feature space, causing interference in feature learning.

\subsubsection{Contrastive learning detail and its loss}
We use the images $I_H$ enhanced by a low-light image enhancement network as anchors for contrastive learning. For negative samples, we adjust the brightness of normal-light images using BoC and transform them into over/underexposed images $I_{N}$. Positive samples are the normal-light images $I_{P}$. The positive and negative samples do not pair with each other or the anchor image, \emph{i.e.}, from different scenes.
\\
\textbf{Feature extraction network} 
We incorporate a pre-trained VGG-16 model to extract the feature map $f\in \mathbb{R}^{C\times H\times W}$ for the latent feature space, where $G^{l}_{ij}$ is the inner product between the feature maps $i$ and $j$ in the layer $l$: 
\begin{equation}
    G_{i j}^{l}=\sum_{k} f_{i k}^{l} f_{j k}^{l}
\end{equation}
where $k$ represents the vector length. 
We then get a set of Gram matrices $\left\{G^{1}, G^{2}, \ldots, G^{L}\right\}$ from layers $1, . . . , L$ in the feature extraction network. 
The Gram matrix $G$ is a quantitative description of latent image features. Contrastive learning aims to learn a feature space in which samples of the same category should be closer to each other while samples of different categories should be farther away. 
\\
\textbf{Contrastive loss} 
A reasonable contrastive loss is necessary to pull the anchors into the positive samples and push away the anchors from the negative samples in the latent space. 
Triplet loss~\citep{Hermans_Beyer_Leibe_2017}, N-pair loss~\citep{Sohn_2016}, and InfoNCE loss~\citep{Gutmann_Hyvärinen_2010} are commonly used loss functions in contrastive learning. 
The choice of which loss function to use depends on the specific task and dataset we discuss in Sec.~\ref{secDIS}. In PIE, 
we mix the triple loss and infoNCE loss in contrastive learning to design the contrastive learning loss.We utilize triplet loss for the Gram matrix $G$, aiming to:
\begin{equation}
  d (G_{I_{H}}, G_{I_{P}})\ll d (G_{I_{H}}, G_{I_{N}})
\end{equation}
where $d$ represents the distance between features. Unlike Gram matrix $G$, we use infoNCE loss for the expectation value $E$, our goal is:

\begin{equation}
d (E_{I_{H}}, E_{I_{P}})\ll d (E_{I_{H}}, E_{I_{N}})
\end{equation}\\
We wish that the distance $d$ between features $I_{H}$ and $I_{P}$ is smaller than the distance between features $I_{H}$ and $I_{N}$. 
Based on the aforementioned objectives, we designed $L_{cG}$ and $L_{cE}$ as two components of contrastive learning loss $L_c$.\\
For Gram matrix $G$:  
\begin{equation}
{
\begin{aligned}
 L_{cG}&=max\left \{ d (G_{I_{H}}, G_{I_{P}})-d (G_{I_{H}}, G_{I_{N}}) + \alpha , 0 \right \}
\end{aligned}
}
\end{equation}
$\alpha$ is a hyperparameter, and we set it to 0.3.\\
For the expectation $E$:
\begin{equation}
{
\begin{aligned}
  L_{cE}&= -log\frac{exp(d (E_{I_{H}}, E_{I_{P}}))}{exp(d (E_{I_{H}}, E_{I_{P}}))+exp(d (E_{I_{H}}, E_{I_{N}}))} \\
\end{aligned}
}
\end{equation}
Therefore, the contrastive loss function in PIE is expressed as follows:
\begin{equation}
\begin{aligned}
  L_{c} = L_{cG} + L_{cE}
\end{aligned}
\end{equation}
\\
\textbf{The numbers of positive and negative samples} 
In a theoretical work~\citep{li2021triplet}, the author argued that a 1:1 rate of positive to negative samples is sufficient for triplet loss. The author also observed significant benefits in contrastive learning of visual representations from randomness.
Inspired by this work, our method involves using one underexposed and one overexposed sample as negatives for each scene. Positive and negative samples are randomly selected during each iteration of training to enhance the model's robustness. Our positive and negative samples are obtained from the SICE dataset~\citep{Cai2018deep}, which consists of 589 scenes (360 scenes in Part1 and 229 scenes in Part2) with a total of 4413 multi-exposure images. In our method, all 360 standard images in all scenes of Part1 are used as positive samples, while negative samples are generated by applying BoC to the standard images to produce under/overexposed images.

In Sec.~\ref{secDIS}, we investigate the impact of different rates of positive and negative samples  (1:1, 1:5, 5:1, and 5:5 in a batch) on low-light enhancement results. Additionally, we consider the average training time for each epoch, which involves training on all samples in the training set once.

\subsection{Regional Segmentation Module}
\subsubsection{Unsupervised super-pixel segmentation}
\begin{figure}[h]
\centering
\subfigure[The enhanced images $\{I_H\}$]{
\centering
\includegraphics[width=7.5cm]{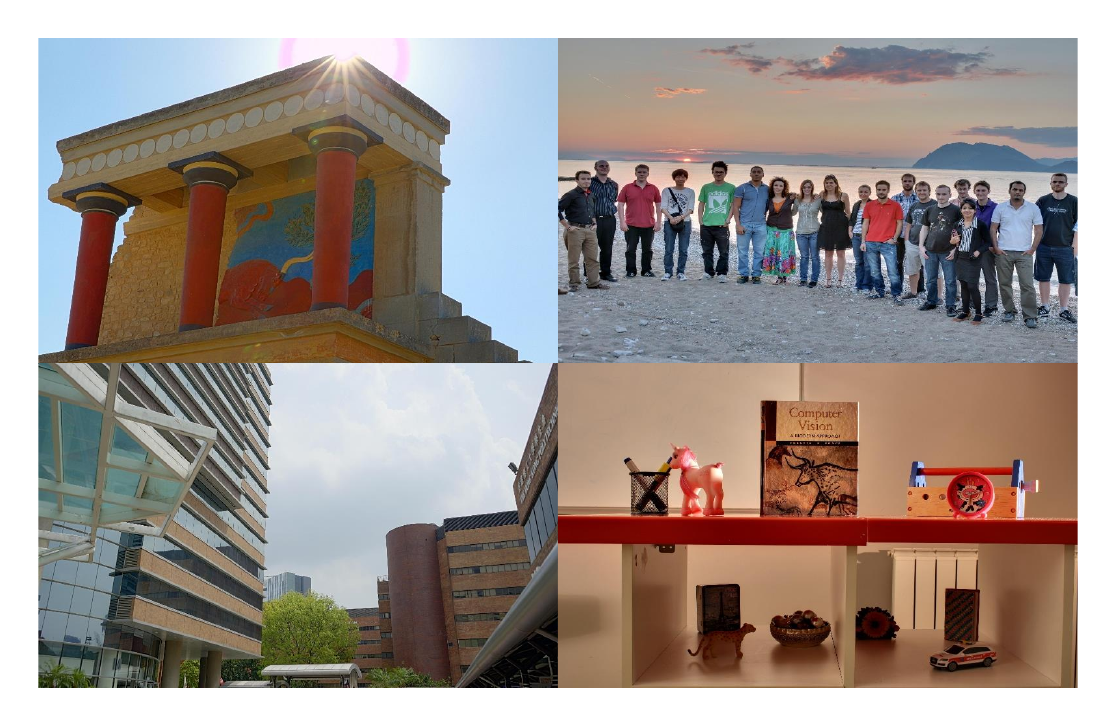}
}
\vspace{-0.2cm}
\subfigure[Graph-based pixel-level segmentation ${S}$]{
\centering
\includegraphics[width=7.5cm]{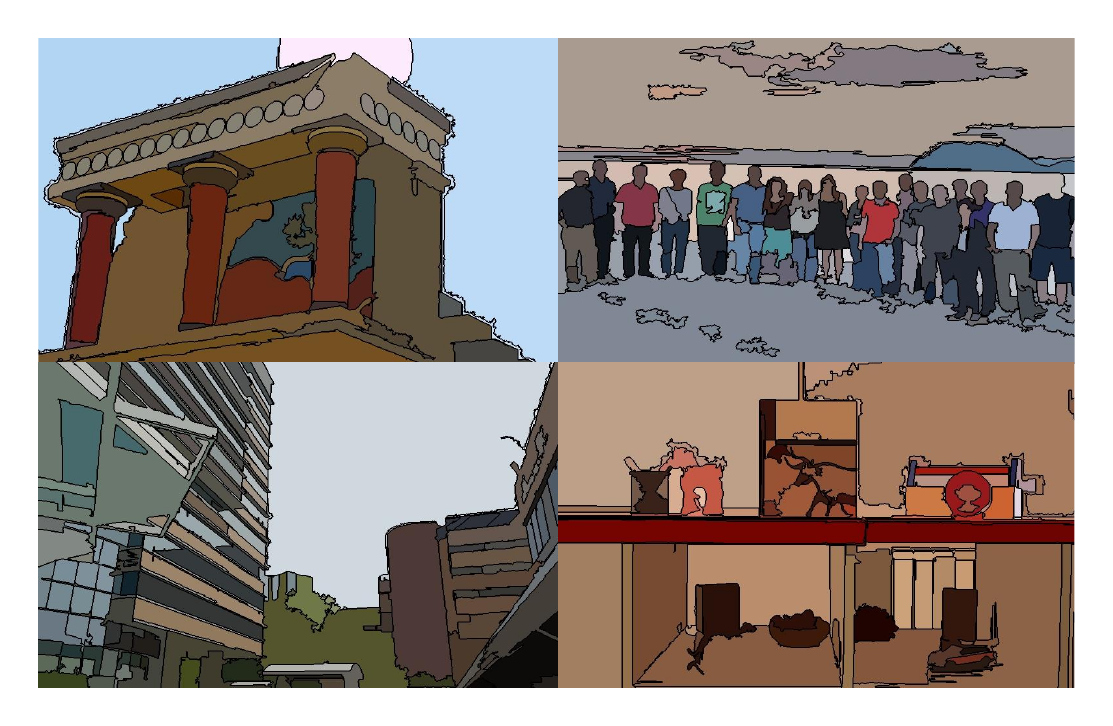}
}
\caption{Demonstration of the enhanced images $\{I_H\}$ and segmentation results after Graph-based pixel-level segmentation ${S}$.}
\label{segmentation}
\end{figure}

In real-world scenes, it is expected that the same region of an object should have uniform brightness, while the enhancement strategies applied to the background and foreground should be different. To address this issue, \cite{liangaaai} incorporates a semantic segmentation module to prevent local overexposure or underexposure. However, the use of a semantic segmentation module resulted in the model's dependence on semantic ground truth. PIE introduces an unsupervised regional segmentation module that uses a super-pixel segmentation to maintain regional brightness consistency and enable region-discriminate enhancement while avoiding reliance on semantic labels. For this purpose, we employ a Graph-based supervised super-pixel segmentation method~\citep{felzenszwalb2004efficient} as illustrated in Fig.~\ref{segmentation}. We first use super-pixel segmentation to divide an image into super-pixel blocks. Then, we utilize the regional brightness consistency loss $L_{rc}$ to maintain the consistency of brightness within each region.


The output is a segmentation component $S=\{c_1,c_2...\}$.
During the Graph-based super-pixel segmentation process, it compares the inter-domain difference ${Dif}\left (c_{i}, c_{j}\right)$ between two different regions $c_{i}$ and $c_{j}$, with the minimum intra-domain difference $Mint\left (c_{i}, c_{j}\right)$ between the smallest regions, $c_{i}$ and $c_{j}$, within these two segmentation regions. If the difference between components is larger than the minimum internal difference, it indicates that there is a boundary between these two regions; otherwise, these two regions are merged while other regions remain unchanged. The judgment method is as follows: 
\begin{small}
\begin{equation}
\begin{split}
\mathrm{D}\left (c_{i}, c_{j}\right)=\left\{\begin{array}{ll}
\text { True } & Dif\left (c_{i}, c_{j}\right)>Mint\left (c_{i}, c_{j}\right) \\ 
\text { False } & \text { otherwise }
\end{array}\right.
\end{split}
\end{equation}
\end{small}
\subsubsection{Regional brightness consistency loss}
The application of super-pixel segmentation frees our method from dependence on semantic ground truths information. 
We define an average value $B$ of the brightness level of the overall pixels in each super-pixel block $c \in S$ as follows:
\begin{equation}
  B_{c}=\frac{1}{n}\sum_{i\in {\theta_{c}}}^{} (B_{I_{H}}^{i})
\end{equation}  
where $c$ represents the $c$-th super-pixel block, and we can attain multiple averages representing individual super-pixel block separately $\left\{B_{1}, B_{2}, \ldots \right\}$. $n$ represents the number of pixels in this super-pixel block. We denote $\theta_{c}$ as the pixel index collection belonging to block $c$, $B_{I_{H}}^{i}$ as the brightness level in the enhanced image $I_{H}$ at the block $c$. The regional brightness consistency loss $L_{rc}$ is defined as:
\begin{equation}
  L_{rc}=\sum_{c=1}^{C}\sum_{i\in {\theta_{c}}}^{} (B_{I_{H}}^{i}-B_{c})^{2}
\end{equation}
where $C$ is the number of the super-pixel blocks. 
\subsection{Other Details}
\subsubsection{Feature preservation loss}
Many low-level visual tasks~\citep{ledig2017photo,Kupyn_2018_CVPR,johnson2016perceptual} use the perceptual loss to make desired images and their features and ground truth perceptually consistent. We also leverage perceptual loss as our feature retention loss to preserve the image features before and after enhancement. The feature retention loss $L_{fr}$ is defined as:
\begin{equation}
L_{fr}=\frac{1}{C_{l}W_{l}H_{l}} (f^{l} (I_{L})-f^{l} (I_{H}))^{2}
\end{equation}
where $f^{l} (I_{L})$ denotes the feature map $f\in \mathbb{R}^{C\times H\times W}$ of the input image $I_{L}$ in the layer $l$, and $f^{l} (L_{H})$ is the feature map of the enhanced image $I_{H}$ in the layer $l$.

Since the color naturalness is one of the significant concerns of LLE, we add a color constancy term $L_{cc}$ incorporating with the feature retention term, following the way reported in~\citep{Guo_2020_CVPR}. Based on the gray-world color constancy hypothesis~\citep{buchsbaum1980spatial}, the pixel averages of the three channels tend to be the same value. $L_{cc}$ constrains the ratio of three channels to prevent potential color deviations in the enhanced image. In addition, to avoid aggressive and sharp changes between neighboring pixels, an illumination smoothness penalty term is also embedded in $L_{cc}$. The formulation of $L_{cc}$ can be expressed as:
\begin{equation}
\begin{aligned}
    L_{cc} &=\sum_{\forall (p,q)\in \xi } (J^p-J^q)^2 \\
            &+ \lambda \frac{1}{M}\sum_{m=1}^{M}\sum_{p\in \xi } (\left |{\triangledown _x{A}_{m}^{p}} \right|+\left |{\triangledown _y{A}_{m}^{p}} \right|), \\
            &\xi=\left \{ R,G,B \right \}
\end{aligned}
\end{equation}
where $J^p$ denotes the average intensity value of $p$ channel in the enhanced image, $ (p,q)$ represents a pair of channels, $M$ is the number of the iterations, and $\triangledown _x$ and $\triangledown _y$ denote the horizontal and vertical gradient operations, respectively. $A$ is a parameter map with the same size as the image. Each pixel has a corresponding higher-order curve parameter generated in multiple iterations. $A_{m}^{p}$ denotes the parameter map of channel $p$ in $m$-th iteration. We set $\lambda$ to 200 in our experiments for the best outcome.

The feature preservation loss $L_{fp}$ is the sum of $L_{fr}$ and $L_{cc}$.
\subsubsection{Efficient training details}
In our implementation, the feature extraction network is pre-trained on ImageNet~\citep{ILSVRC15}, the CBDNet is pre-trained on BSD500~\citep{937655}, Waterloo~\citep{7752930}, MIT-Adobe FiveK~\citep{5995332} and RENOIR dataset~\citep{ANAYA2018144}. We train PIE end-to-end while fixing the weights of the feature extraction network. The back-propagated operation only updates the weights in the image enhancement network. Hence, most network computation is done in the image enhancement network, which efficiently learns $I_H$ from $ (I_L, I_N, I_P)$ to recover the enhanced image with various scenes. 
We resize the training images to the size of 384 $\times$ 384. As for the numerical parameters, we set the maximum epoch as 10 and the batch size as 2. Our network is implemented with PyTroch on an NVIDIA 1080Ti GPU. The Adam optimizer optimizes the model with a fixed learning rate $1e^{-4}$. 

\subsubsection{Downstream task-driven LLE}
We aim to explore whether PIE can benefit downstream tasks. We evaluated LLE on three tasks: semantic segmentation, and face detection. 
\\ 
\textbf{Semantic segmentation}
Our earlier work~\citep{liangaaai} with a semantic brightness consistency loss has demonstrated the effectiveness of LLE in improving downstream semantic segmentation. In this study, 
to validate the gain of PIE on semantic segmentation, we replace the regional segmentation module in PIE with the same semantic segmentation module used in ~\cite{liangaaai}. Additionally, we replace the regional brightness consistency loss $L_{rc}$ with the semantic brightness consistency loss $L_{sc}$. The semantic segmentation network we use here is the popular DeepLabv3+~\citep{chen2018encoder}, and we train our network on the training images of the Cityscapes~\citep{cordts2016cityscapes} dataset. 
\\
\textbf{Face detection}
For face detection, we replace the regional segmentation module in the PIE with the RetinaFace~\citep{deng2020retinaface} trained on the WIDER FACE dataset~\citep{yang2016wider} and replace the regional brightness consistency loss $L_{rc}$ with a face detection loss $L_{det}$. The face detection loss $L_{det}$ includes two components: $L_{cls}$ and $L_{box}$. $L_{cls}$ is the softmax loss for binary classification (face/not face), while $L_{box}$ is the face box regression loss which is based ond~\citep{Girshick_2015_ICCV}. The PIE network for face detection, called PIE$_{det}$, is fine-tuned using 4000 images from the DARK FACE dataset~\citep{DARKFACE}. During training, the parameters of the RetinaFace are fixed, and the RetinaFace is introduced only to calculate the detection loss to guide the optimization of the low-light enhancement model. 

More details regarding the downstream task-driven LLE will be presented in the following experiments outlined in Sec. \ref{Gain4Down}.
\begin{figure*}[!htb]
\centering
\subfigure[Input]{
\includegraphics[width=3.0cm]{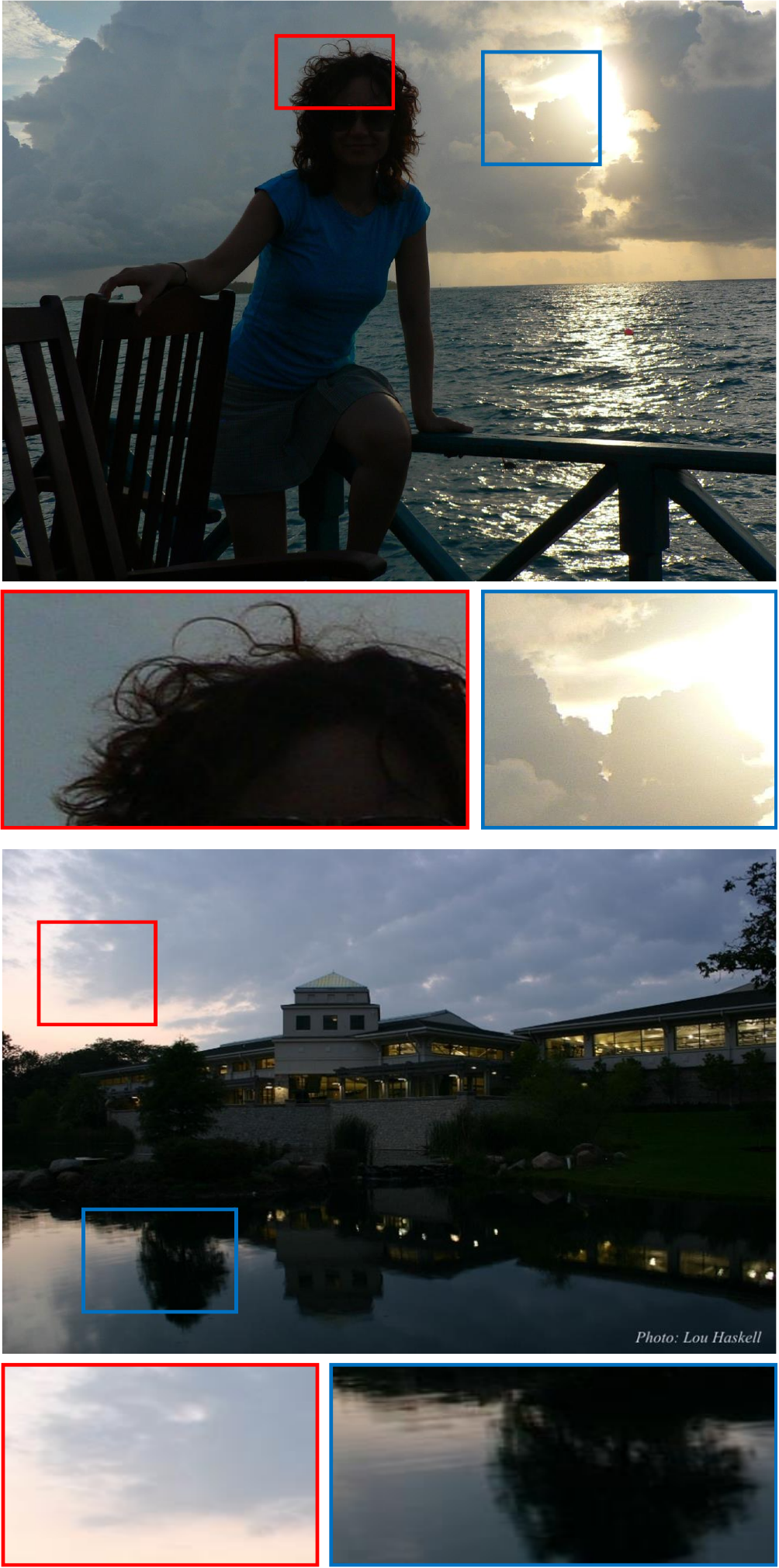}
}\hspace{-1mm}
\subfigure[LIME]{
\includegraphics[width=3.0cm]{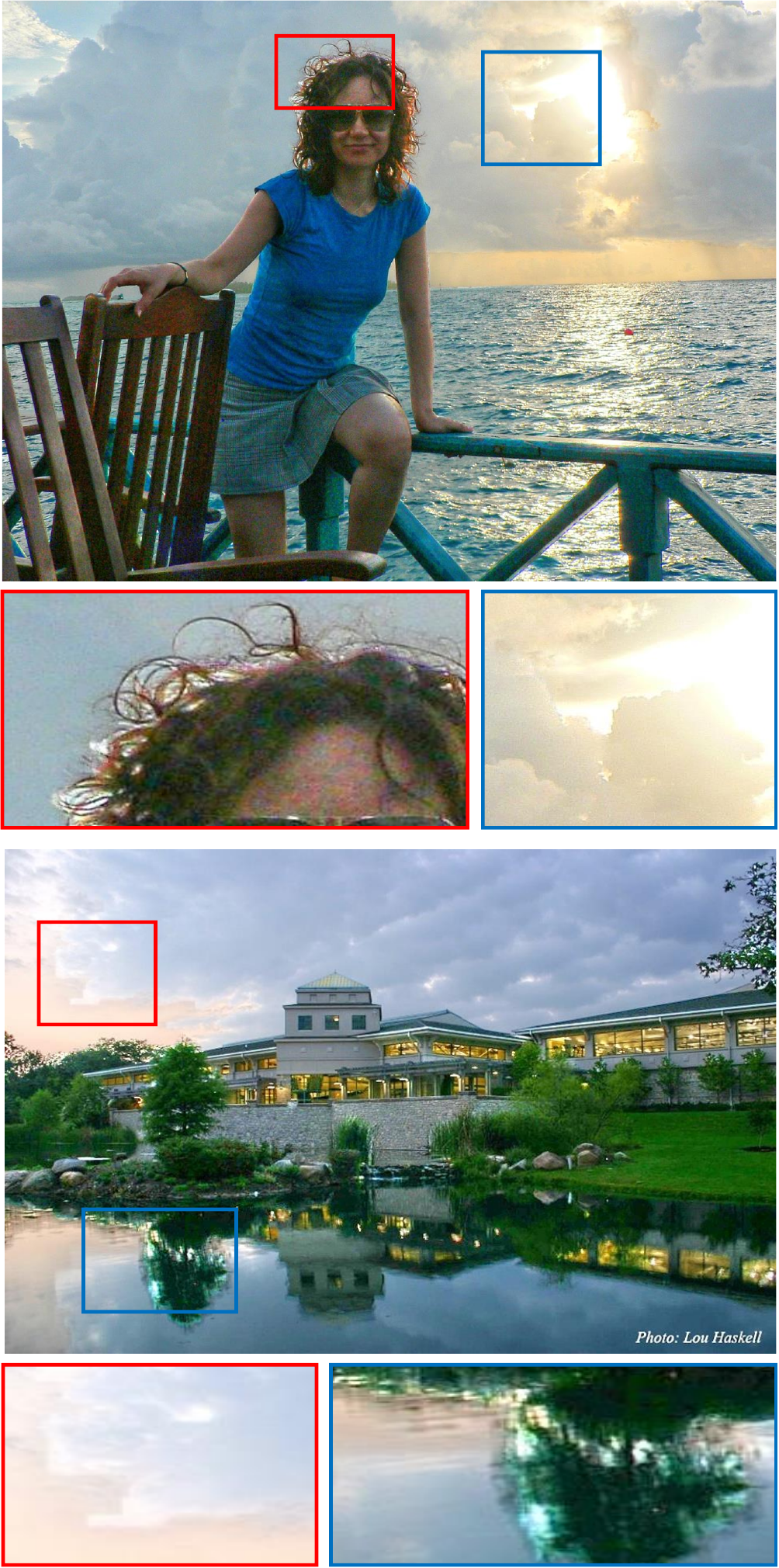}
}\hspace{-1mm}
\subfigure[RetinexNet]{
\includegraphics[width=3.0cm]{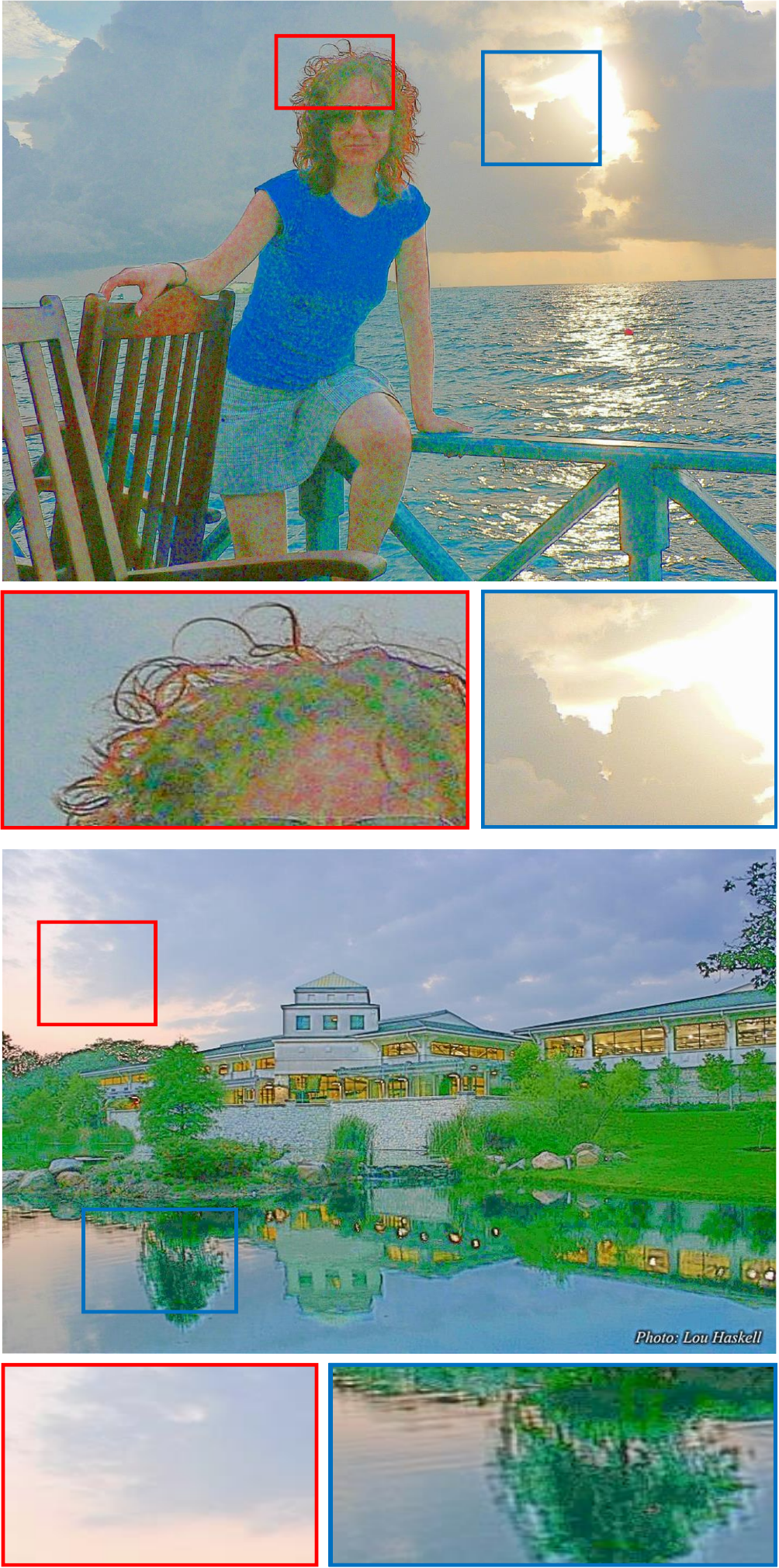}
}\hspace{-1mm}
\subfigure[ISSR]{
\includegraphics[width=3.0cm]{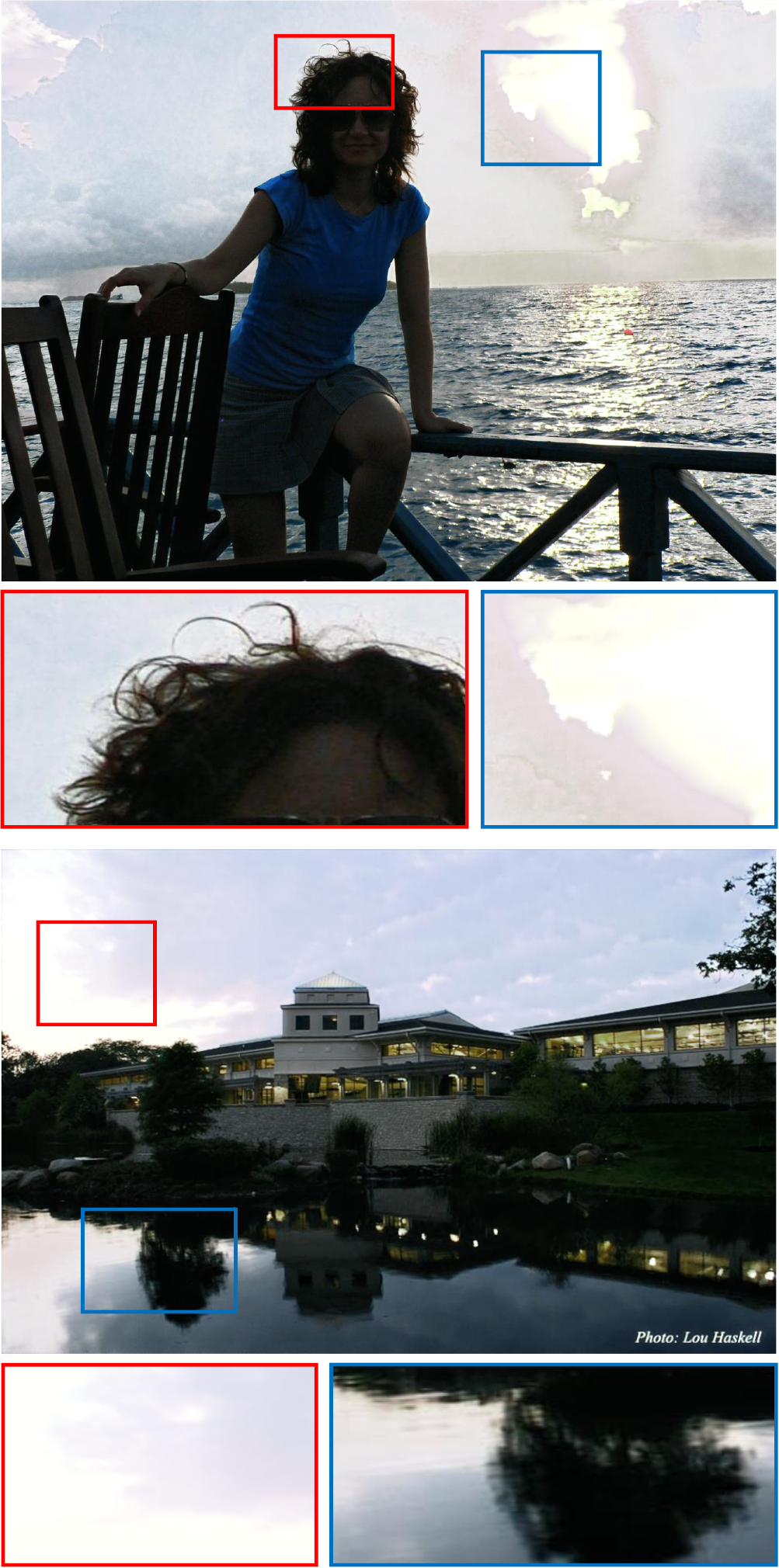}
}\hspace{-1mm}
\subfigure[Zero-DCE]{
\includegraphics[width=3.0cm]{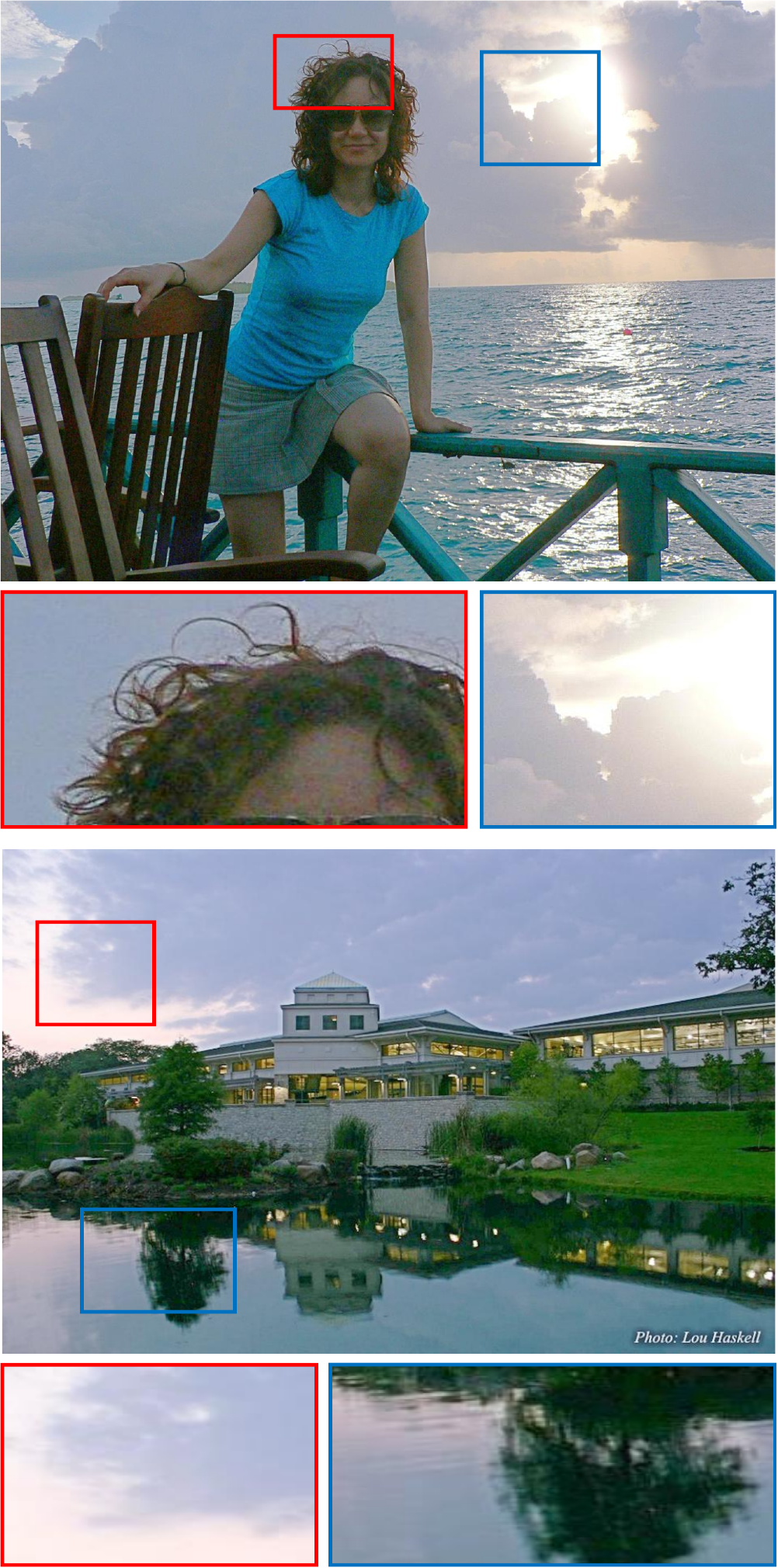}

}\hspace{-1mm}\quad
\\
\subfigure[EnlightenGAN]{
\includegraphics[width=3.0cm]{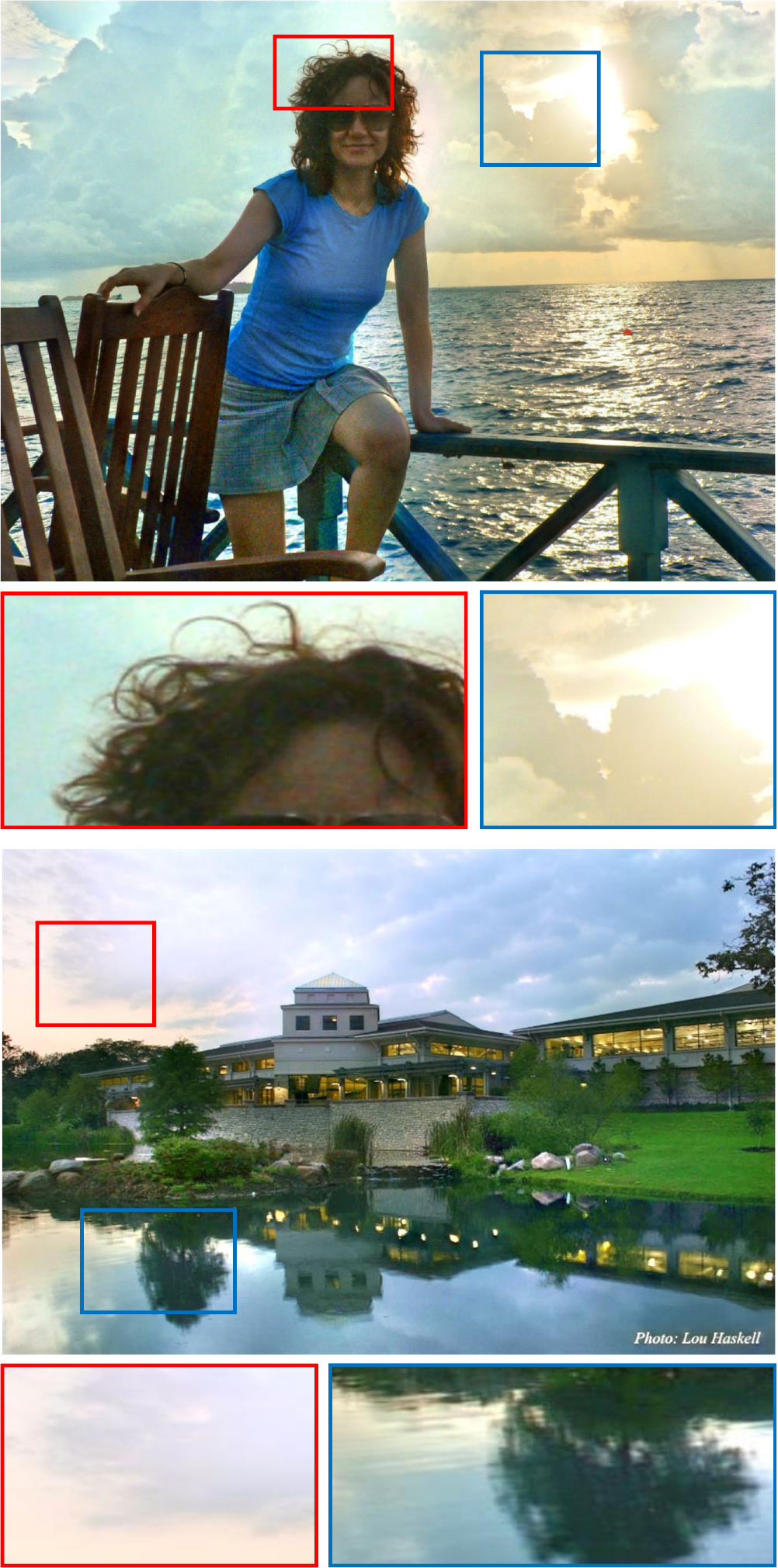}
}\hspace{-1mm}
\subfigure[RUAS]{
\includegraphics[width=3.0cm]{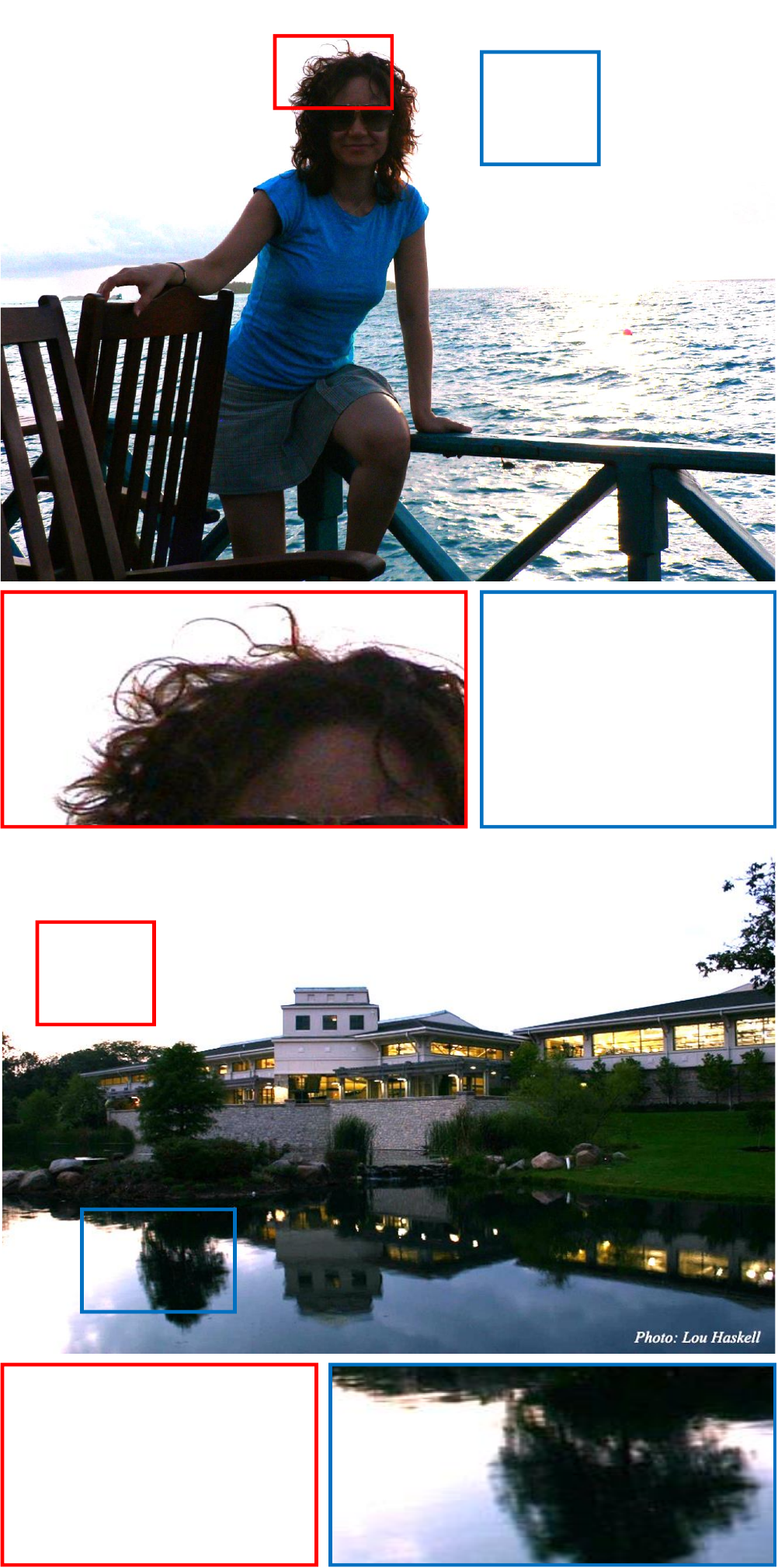}
}\hspace{-1mm}
\subfigure[ReLLIE]{
\includegraphics[width=3.0cm]{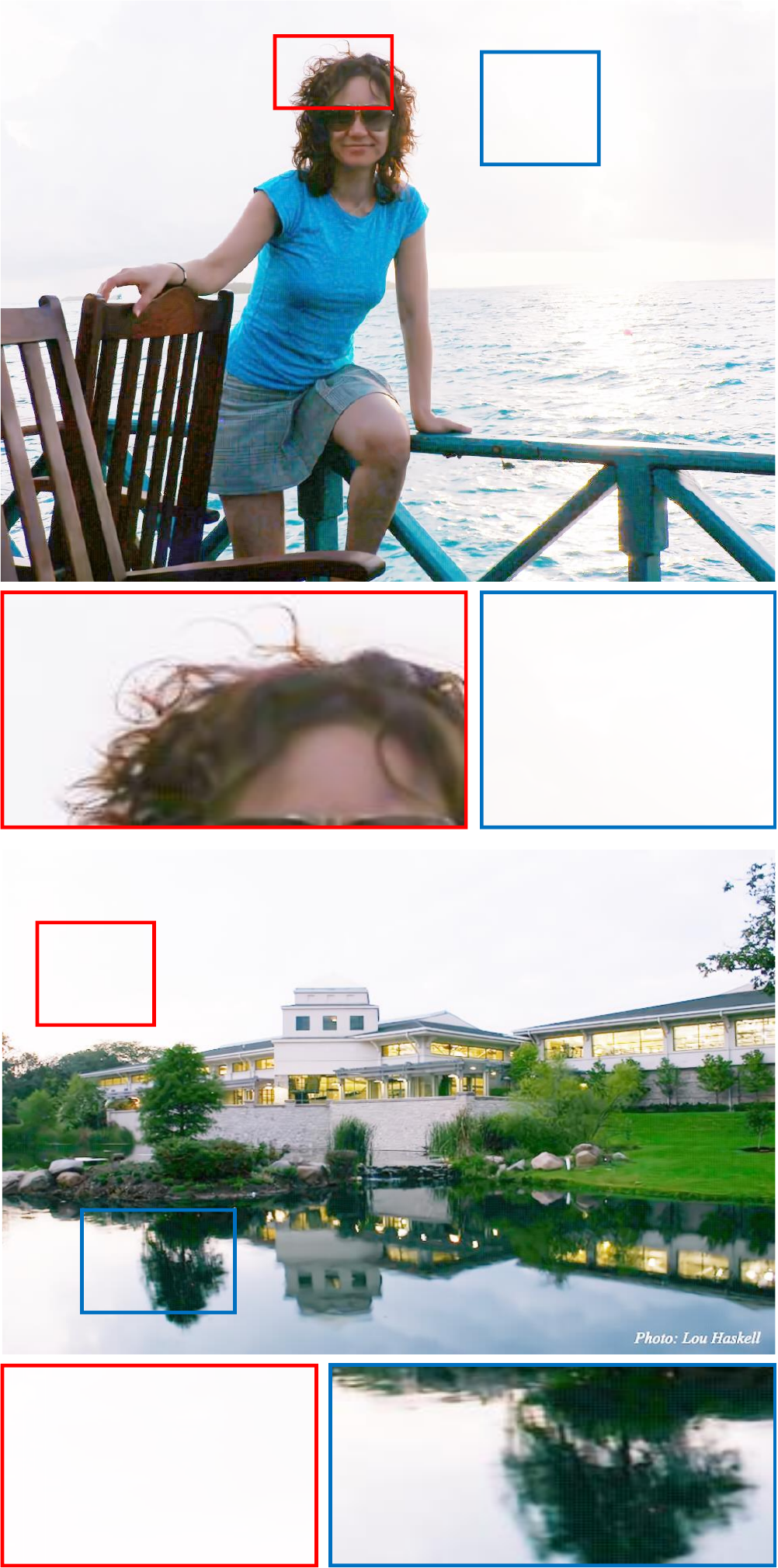}
}\hspace{-1mm}
\subfigure[SCL-LLE]{
\includegraphics[width=3.0cm]{SCL-LLE2.pdf}
}\hspace{-1mm}
\subfigure[PIE]{
\includegraphics[width=3.0cm]{PIE2.pdf}
}\hspace{-1mm}\quad
\caption{Demonstration of PIE and the state-of-the-art methods over VV (the first sample) and DICM (the second sample) datasets with zoom-in regions. 
PIE  enables the enhanced images to look more realistic and recovers better details and richer color in both foreground and background. 
}
\label{DICM}
\end{figure*}
\begin{figure*}[!htb]
\centering
\subfigure[Input]{
\includegraphics[width=3.0cm]{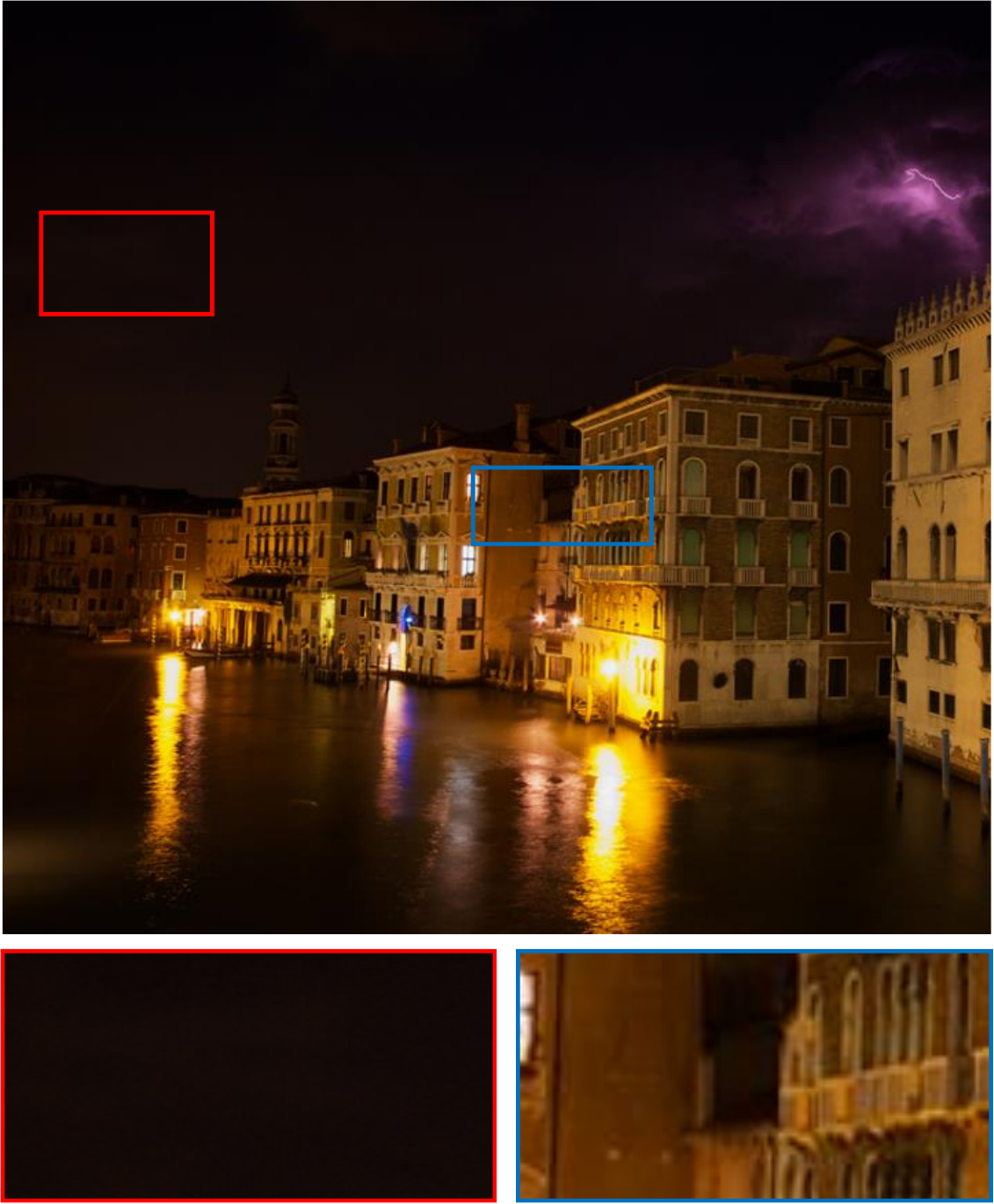}
}\hspace{-1mm}
\subfigure[LIME]{
\includegraphics[width=3.0cm]{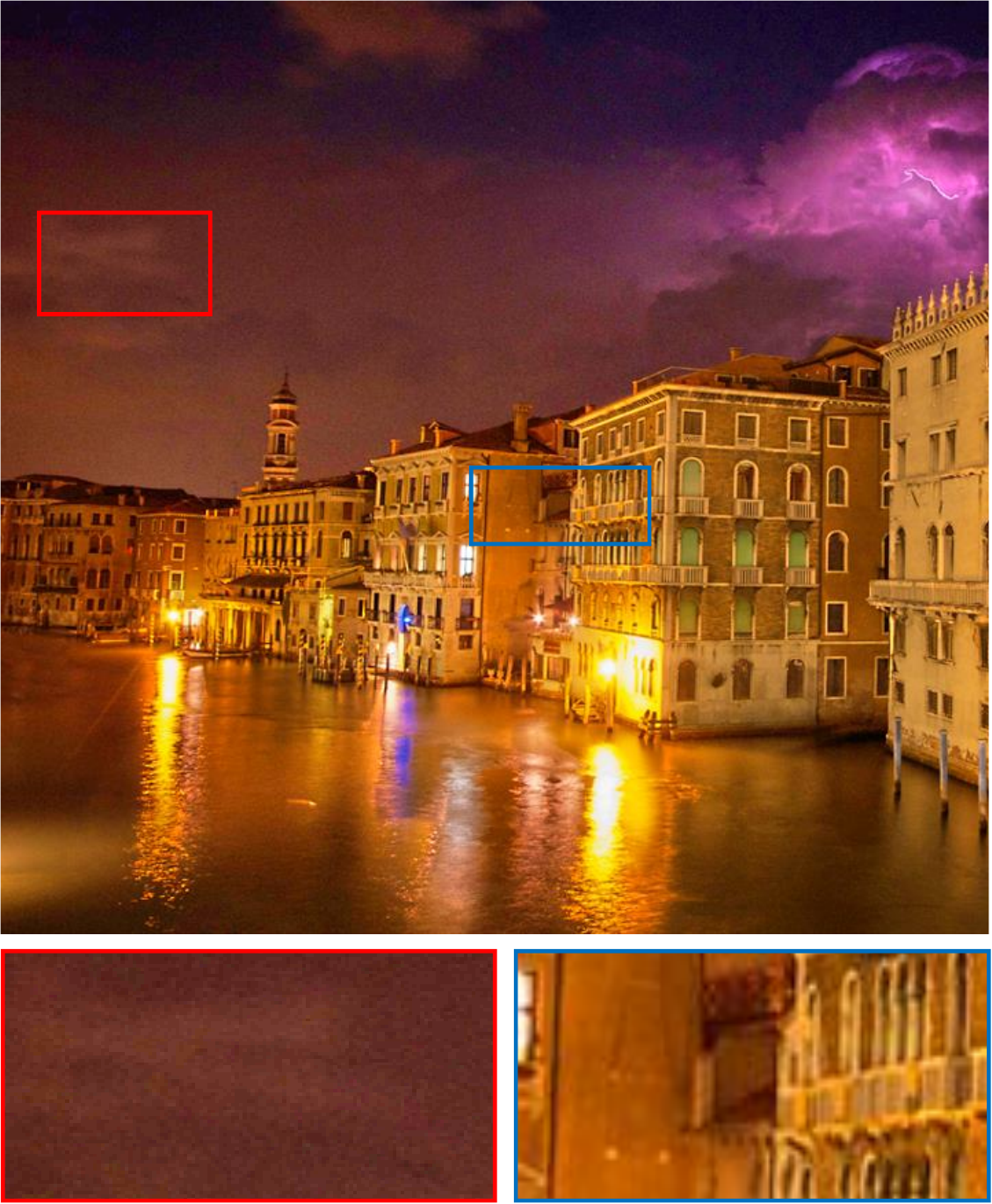}
}\hspace{-1mm}
\subfigure[RetinexNet]{
\includegraphics[width=3.0cm]{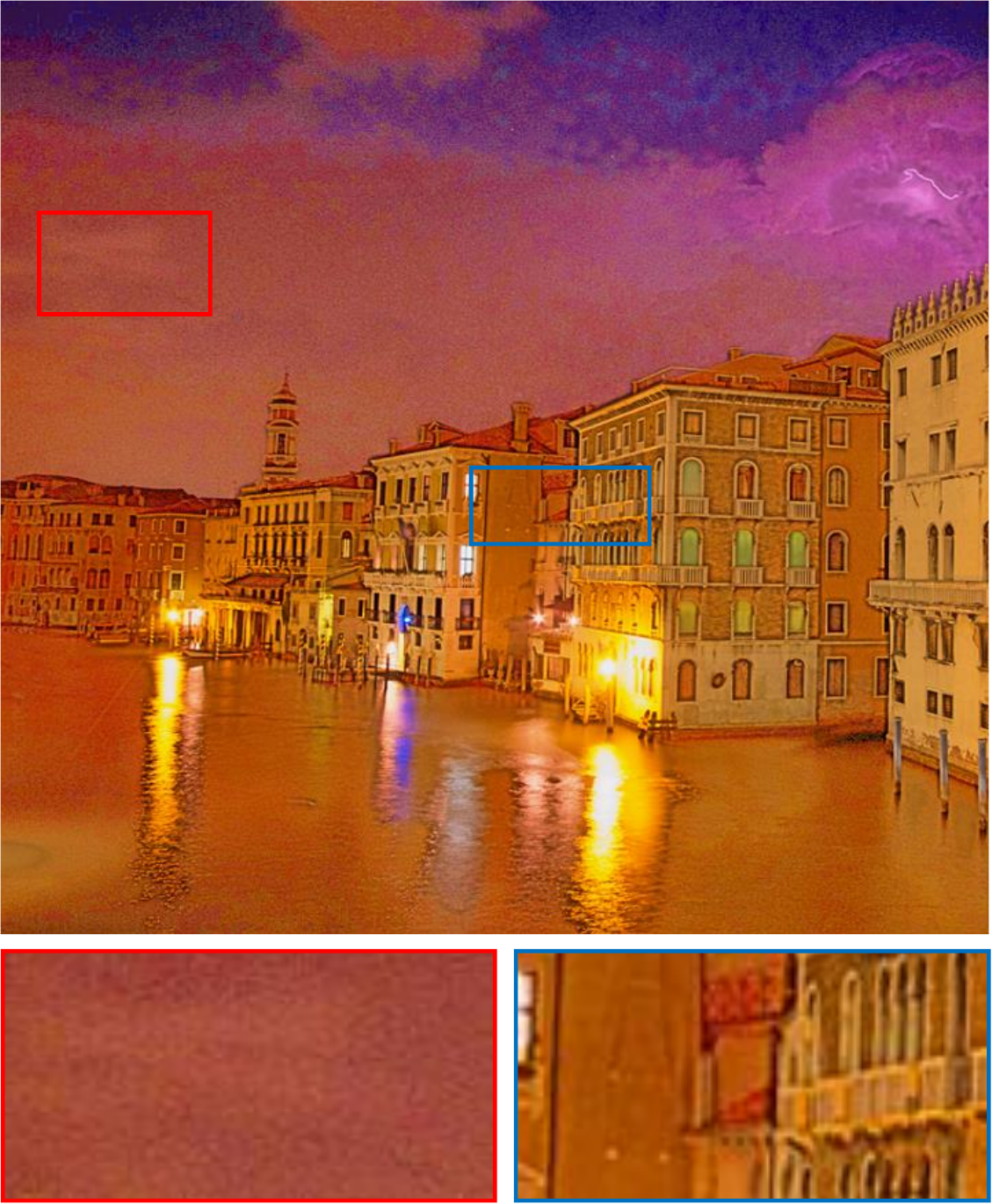}
}\hspace{-1mm}
\subfigure[ISSR]{
\includegraphics[width=3.0cm]{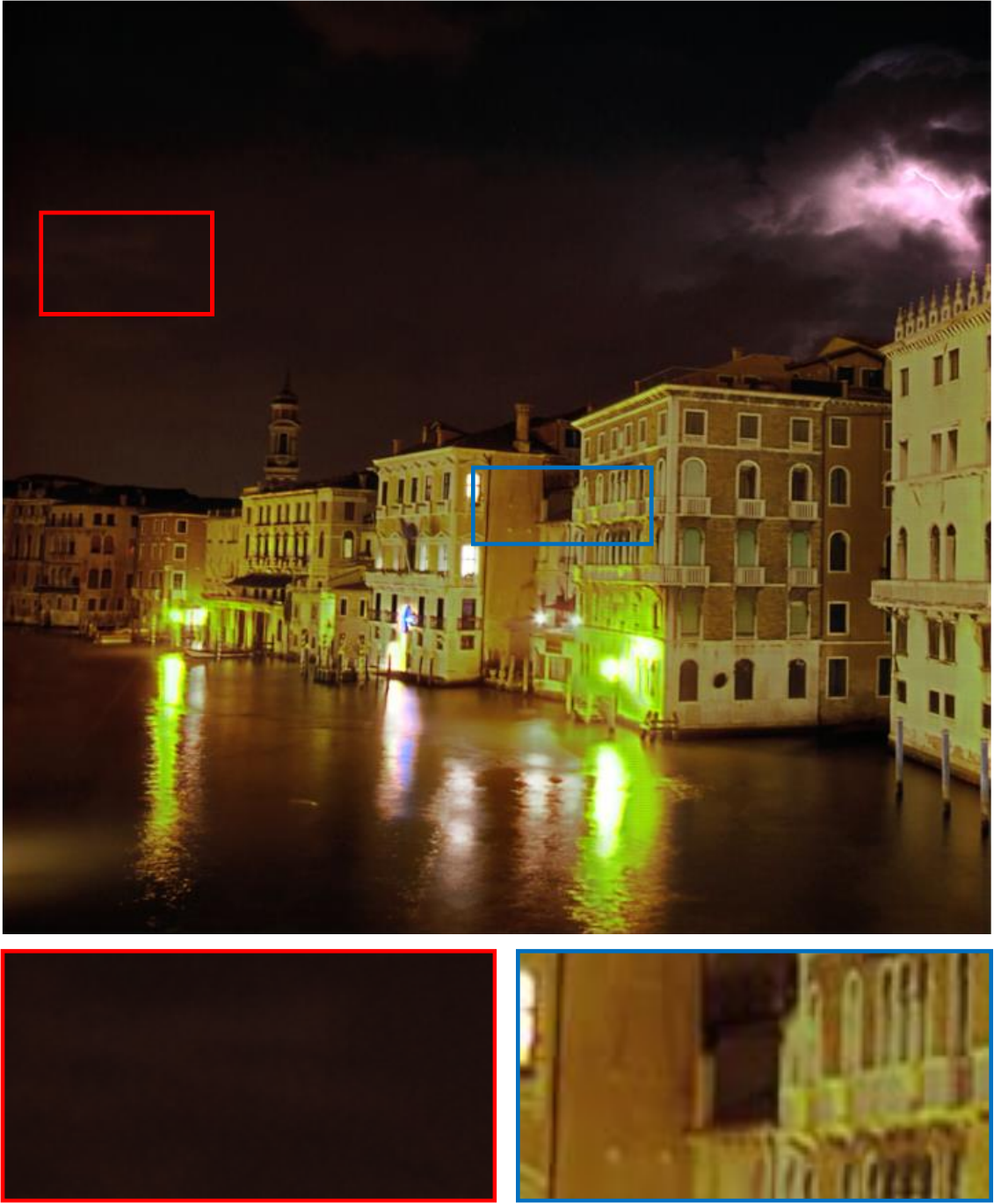}
}\hspace{-1mm}
\subfigure[Zero-DCE]{
\includegraphics[width=3.0cm]{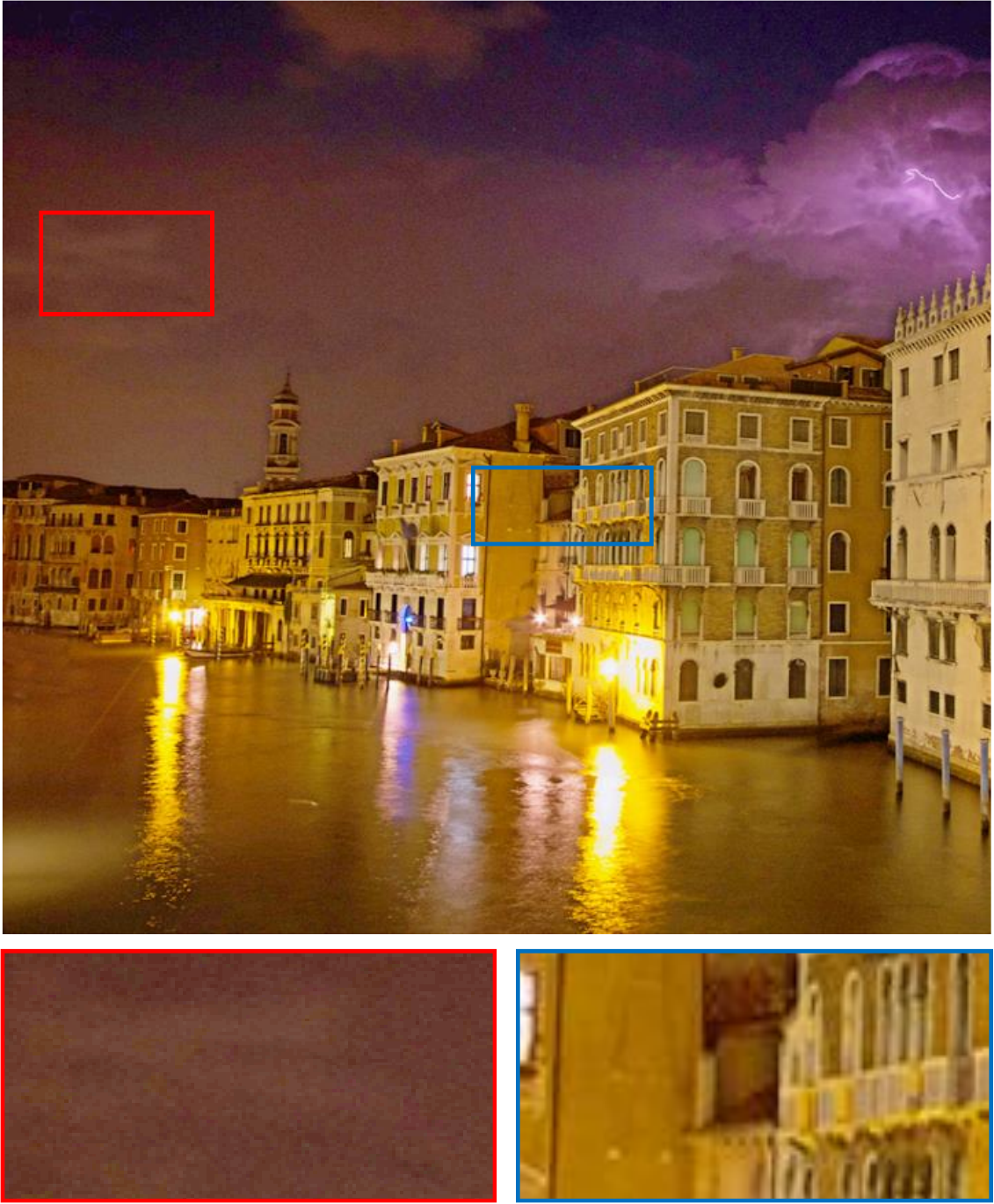}
}\hspace{-1mm}\quad
\\
\subfigure[EnlightenGAN]{
\includegraphics[width=3.0cm]{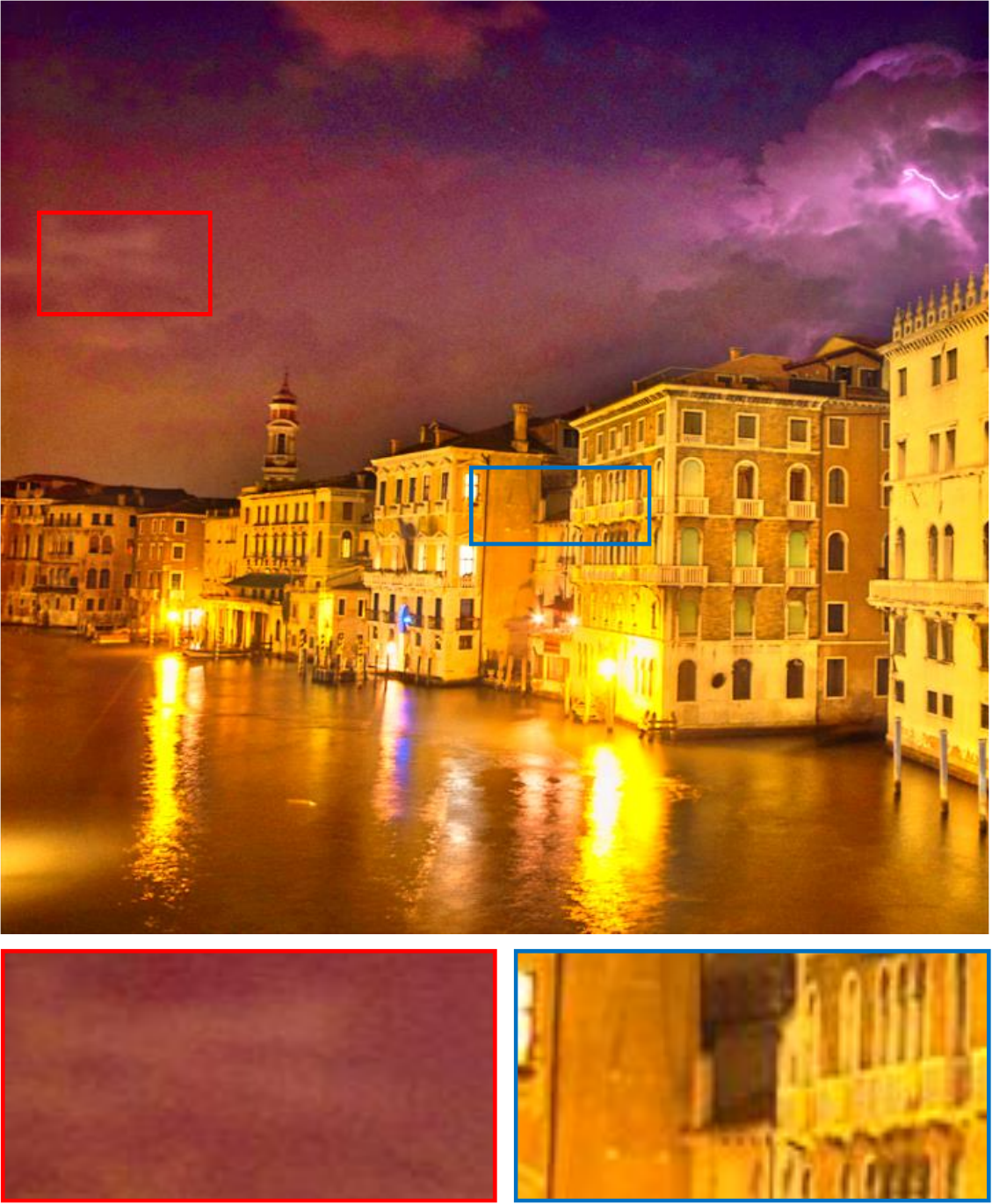}
}\hspace{-1mm}
\subfigure[RUAS]{
\includegraphics[width=3.0cm]{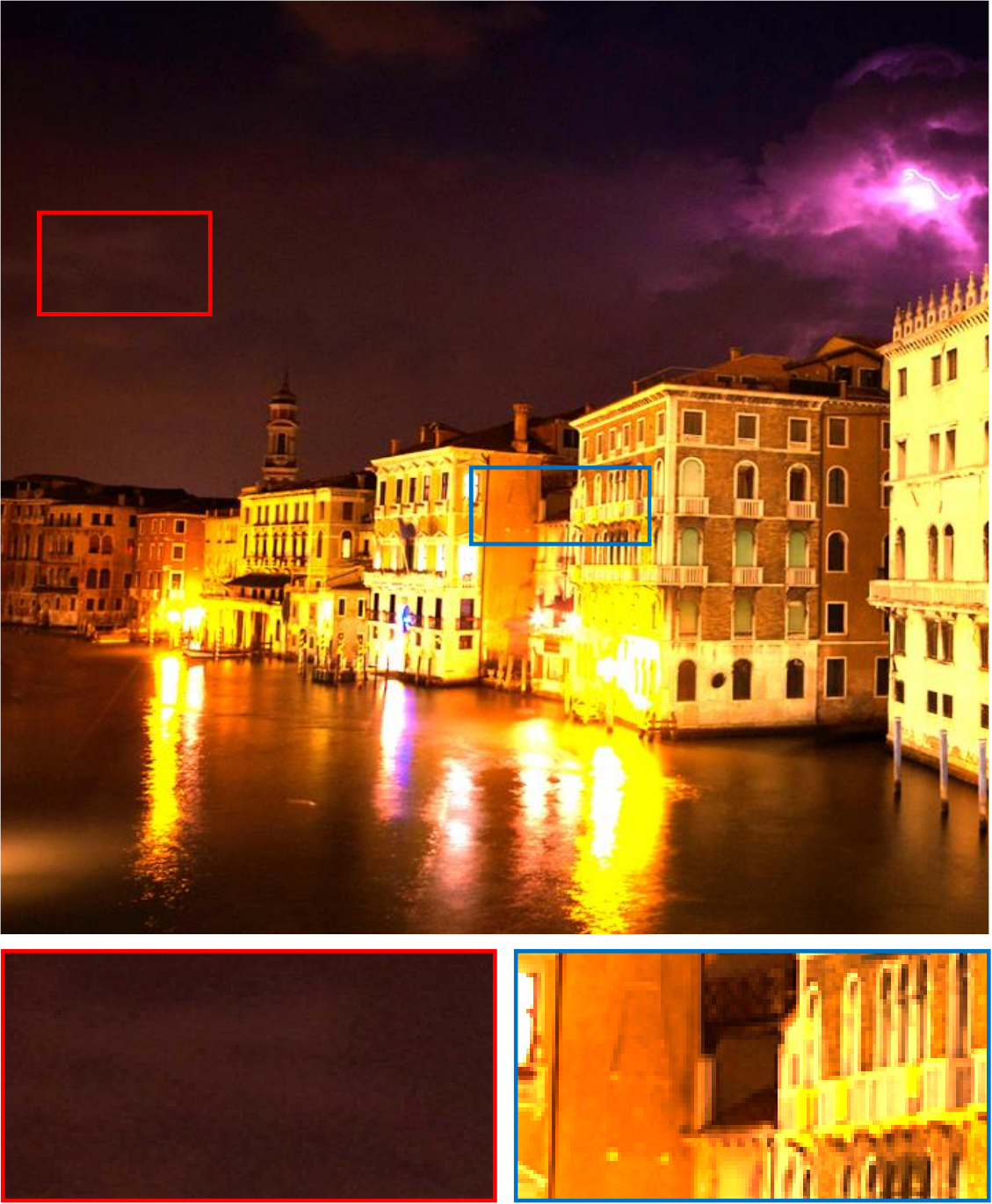}
}\hspace{-1mm}
\subfigure[ReLLIE]{
\includegraphics[width=3.0cm]{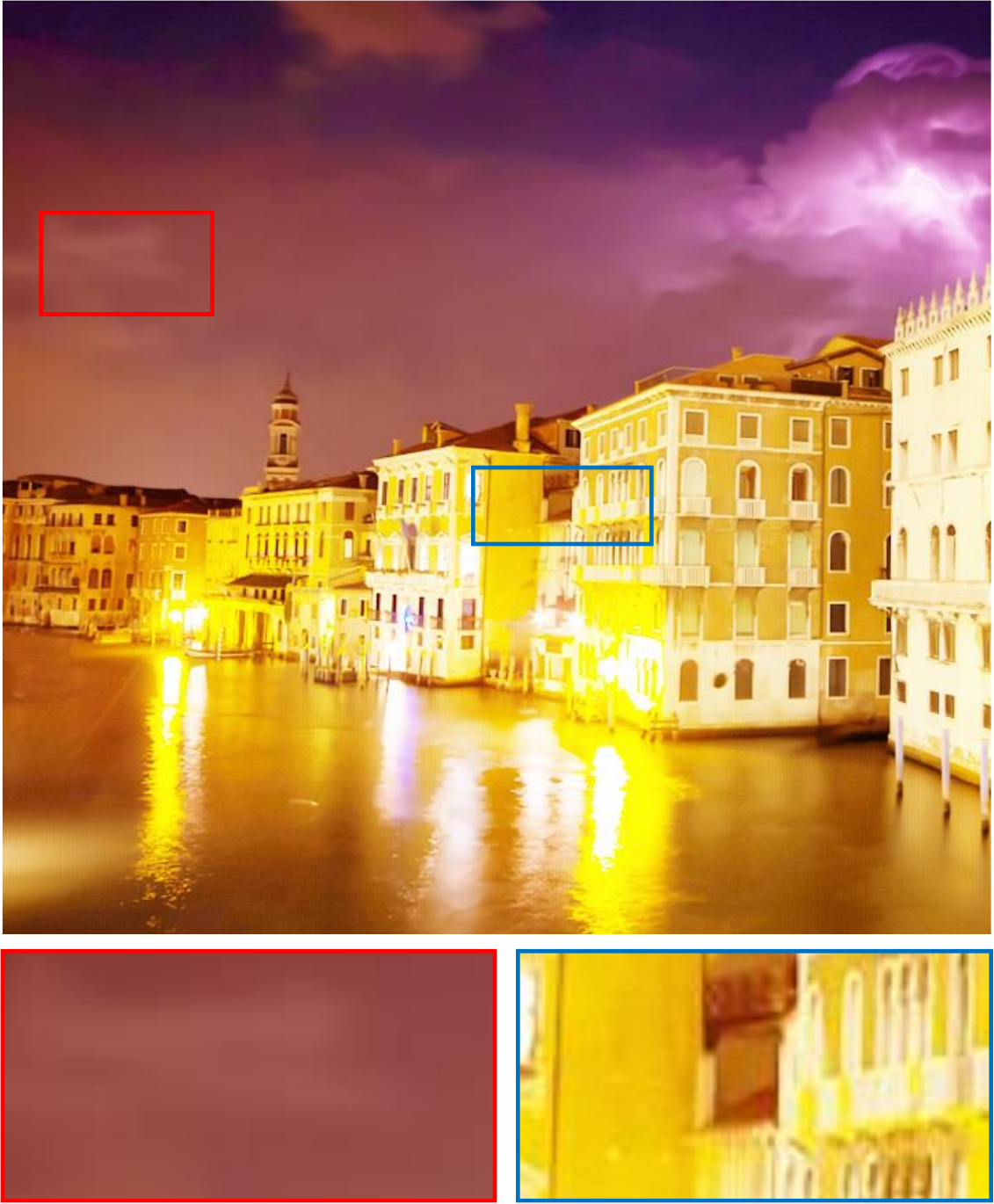}
}\hspace{-1mm}
\subfigure[SCL-LLE]{
\includegraphics[width=3.0cm]{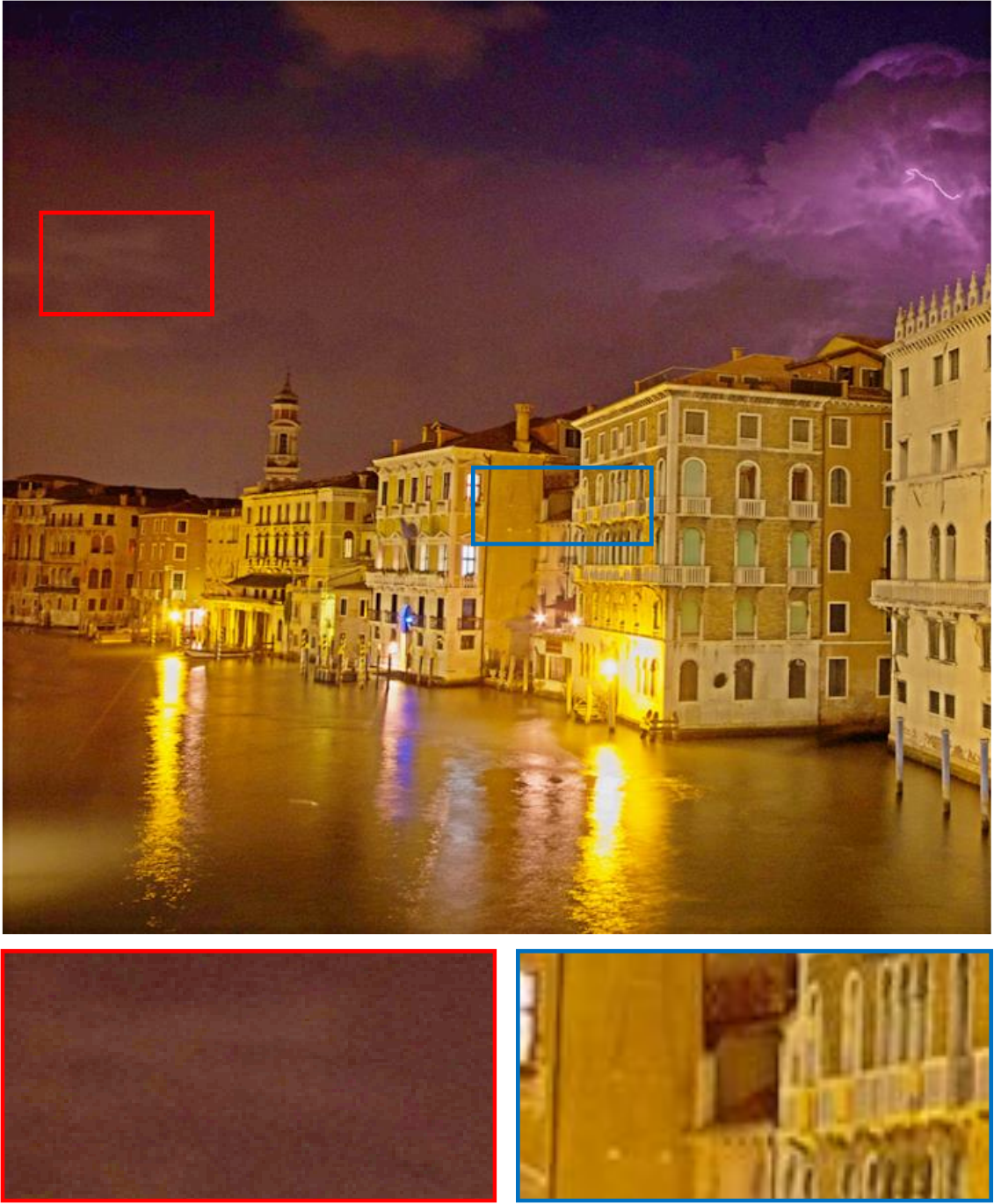}
}\hspace{-1mm}
\subfigure[PIE]{
\includegraphics[width=3.0cm]{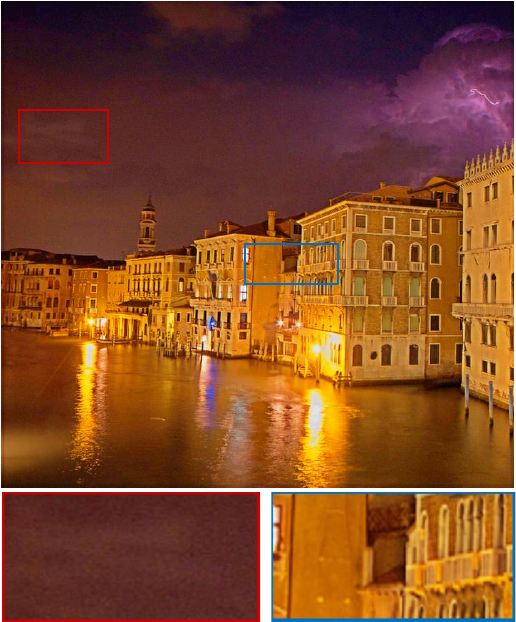}
}\hspace{-1mm}\quad
\caption{Demonstration of PIE and the state-of-the-art methods with a night-view sample in LIME dataset with zoom-in regions.
PIE enables enhanced images with more realistic color and details in both foreground and background and retouching with a negligible noise level on the dark background (the sky). 
}
\label{LIME}
\end{figure*}
\section{Experiments}
\label{Experiments}
\subsection{Cross-dataset Peer Comparison}
For testing images, we use six publicly available low-light image datasets from other reported works, \emph{i.e.}, DICM~\citep{lee2012contrast}, MEF~\citep{ma2015perceptual}, LIME~\citep{guo2016lime}, NPE~\citep{wang2013naturalness}, VV\footnote{https://sites.google.com/site/vonikakis/datasets} and the Part2 of SICE~\citep{Cai2018deep}. 
DICM, LIME, MEF, VV, NPE, and SICE are ad hoc test datasets, including 64, 10, 17, 24, 8, and 229 images, respectively. They are widely used in LLE testing: SCL-LLE \citep{liangaaai}, EnlightenGAN \citep{jiang2021enlightengan}, Zero-DCE \citep{Guo_2020_CVPR} \emph{et al.}
Images in these datasets are diverse and representative:  DICM is mainly landscaped with extreme darkness; LIME focuses on dark street landscapes; MEF focuses on dark indoor scenes and buildings; VV is mostly backlit and portraits; NPE mainly includes natural scenery in low light; SICE is a large-scale multi-exposure image dataset that contains high-resolution image sequences of multiple scenes. 
Note that all the images in the six datasets are independent cross-scene images without any overlapped scene of the input image and the positive/negative samples. 

We compare our proposed method with 13 representative state-of-the-art methods for heterogeneous image enhancement. These included the conventional method LIME~\citep{guo2016lime}, the GAN-based method EnlightenGAN~\citep{jiang2021enlightengan}, and four Retinex-based methods: RetinexNet~\citep{Chen2018Retinex}, RUAS~\citep{liu2021retinex}, ISSR~\citep{FanWY020}, and Zero-DCE~\citep{Guo_2020_CVPR}, which leverages the same backbone enhancement network as our proposed method. We further include two reinforcement learning based method ReLLIE~\citep{Zhang2021ReLLIEDR} and  ALL-E~\citep{li2023all}, four more recent methods Uretinex-net~\citep{wu2022uretinex}, SCI~\citep{ma2022toward}, PairLLE~\citep{fu2023learning}, and IRN~\citep{zhao2021deep}, and our earlier conference version~\citep{liangaaai}. We reproduce the results of these methods using recommended test settings and publicly available models.


\subsubsection{Visual quality comparison}
We first examine whether the proposed methods can achieve visually pleasing results in brightness, color, contrast, and naturalness. We observe from Fig.~\ref{DICM} and Fig.~\ref{LIME} that all the SOTAs sacrifice over/under/uneven exposure in global or local areas. Specifically, LIME~\citep{guo2016lime} leads to color artifacts in strong local edges (\emph{e.g.}, hair and sky, and inverted reflection in the water); RetinexNet~\citep{Chen2018Retinex} and EnlightenGAN~\citep{jiang2021enlightengan} cause global color distortions with details missing; ISSR~\citep{FanWY020} and RUAS~\citep{liu2021retinex} generate severe global and local over/underexposure; ReLLIE~\citep{Zhang2021ReLLIEDR} suffers from over-enhancement and over-smoothing. In contrast, PIE recovers more details and better contrast in both foreground and background, thus enabling the enhanced images to look more realistic with vivid and natural color mapping.

\begin{table*}[!htb]
\setlength\tabcolsep{2pt} 
\centering
\caption{NIQE $\downarrow$, UNIQUE (UN.) $\uparrow$ and User Study (U.S.) $\downarrow$ scores on DICM, LIME, MEF, VV, and NPE datasets.
}
\resizebox{2.1\columnwidth}{!}
{
\begin{tabular}{c|ccc|ccc|ccc|ccc|ccc|ccc}
\hline
 &\multicolumn{3}{c|}{DICM} &\multicolumn{3}{c|}{LIME} &\multicolumn{3}{c|}{MEF} &\multicolumn{3}{c|}{VV} &\multicolumn{3}{c|}{NPE} &\multicolumn{3}{c}{\textbf{Average}} \\ 

Methods   & NIQE $\downarrow$ & UN. $\uparrow$ & U.S. $\downarrow$ & NIQE $\downarrow$ & UN. $\uparrow$ & U.S. $\downarrow$ & NIQE $\downarrow$ & UN. $\uparrow$ & U.S. $\downarrow$ & NIQE $\downarrow$ & UN. $\uparrow$ & U.S. $\downarrow$ & NIQE $\downarrow$ & UN. $\uparrow$ & U.S. $\downarrow$ & NIQE $\downarrow$ & UN. $\uparrow$ & U.S. $\downarrow$ \\
\hline
Input     & 4.26              & 0.72           & 3.33              & 4.36              & 0.70           & 4.30              & 4.26              & 0.72           & 4.41              & 3.52              & \textbf{0.74}  & 3.38              & 4.32              & 1.17           & 3.92              & 4.13              & 0.75           & 3.67              \\
LIME~\citep{guo2016lime}      & 3.75              & 0.78           & 3.44              & 3.85              & 0.53           & 2.10              & 3.65              & 0.65           & 3.82              & \textbf{2.54}     & 0.44           & 2.75              & 4.44              & 0.93           & 3.75              & 3.55              & 0.69           & 3.40              \\
Retinex-Net~\citep{Chen2018Retinex}     & 4.47              & 0.75           & 3.59              & 4.60              & 0.52           & 4.00              & 4.41              & 0.97           & 4.06              & 2.70              & 0.36           & 2.88              & 4.60              & 0.81           & 4.13              & 4.13              & 0.69           & 3.75              \\
ISSR~\citep{FanWY020}      & 4.14              & 0.59           & 3.13              & 4.17              & \textbf{0.83}           & 3.40              & 4.22              & 0.87           & 4.47              & 3.57              & 0.62           & 3.00              & 4.02              & 0.99           & 3.96              & 4.03              & 0.68           & 3.49              \\
Zero-DCE~\citep{Guo_2020_CVPR}    & 3.56              & 0.82           & 2.77              & 3.77              & 0.73           & 2.10              & 3.28              & 1.22           & 3.18              & 3.21              & 0.48           & 2.50              & 3.93              & 1.07           & 2.50              & 3.50              & 0.81           & 2.70              \\
EnlightenGAN~\citep{jiang2021enlightengan}     & 3.55              & 0.63           & 2.81              & \textbf{3.70}     & 0.49           & \textbf{2.00}     & \textbf{3.16}     & 1.03           & 3.29              & 3.25              & 0.58           & 2.12              & 3.95              & 1.07           & 2.85              & 3.47              & 0.69           & 2.72              \\
RUAS~\citep{liu2021retinex}      & 5.21              & -0.17          & 3.44              & 4.26              & 0.34           & 2.30              & 3.83              & 0.73           & 4.11              & 4.29              & -0.04          & 3.75              & 5.53              & 0.13           & 4.17              & 4.78              & 0.04           & 3.60              \\
ReLLIE~\citep{Zhang2021ReLLIEDR}    & 4.44              & 0.41           & 2.94              & 5.22              & 0.52           & 3.40              & 5.22              & 1.07           & 4.05              & 3.51              & 0.33           & 2.50              & 5.14              & 0.37           & 3.71              & 4.48              & 0.49           & 3.25              \\
Uretinex-net~\citep{wu2022uretinex}        & 3.95              & 0.85           & 2.99              & 4.34              & 0.93           & 2.12              & 3.79              & 1.18           & 3.16              & 3.01              & 0.51           & 2.63              & 4.69              & 0.99           & 2.55              & 3.83              & 0.84           & 2.71              \\
SCI~\citep{ma2022toward}   & 4.11              & 0.11           & 3.46              & 4.21              & 0.35           & 2.83              & 3.63              & 1.04           & 3.95              & 2.92              & 0.05           & 2.98              & 4.47              & 0.21           & 3.92              & 3.87              & 0.35           & 3.72              \\
IRN~\citep{zhao2021deep} & 3.68              & 0.91           & 2.83              & 4.16              & 0.79           & 2.09              & 3.83              & 0.97           & 3.22              & 3.01              & 0.57           & 2.46              & 3.91              & 1.06           & 2.31              & 3.61              & 0.84           & 2.53              \\

ALL-E~\citep{li2023all}                & 3.49     & 0.88           & 2.80              & 3.78              & 0.80           & 2.18              & 3.32              & 1.27  & 2.95              & 3.08              & 0.49           & 2.38              & 3.89              & 1.10           & 2.45              & 3.45              & 0.88           & 2.56              \\
PairLIE~\citep{fu2023learning}                & 4.08     & 0.67           & 3.39              & 4.52              & 0.78           & 3.86              & 4.17              & 1.08  & 4.23              & 3.66              & 0.51           & 2.68             & 4.21              & 0.94           & 2.49              & 4.05              & 0.72           & 3.35             \\

SCL-LLE~\citep{liangaaai}             & 3.51              & 0.87           & \textbf{2.73}              & 3.78              & 0.76           & 2.20              & 3.31              & 1.25           & 2.47              & 3.16              & 0.49           & 1.63              & 3.88              & 1.08           & 2.08              & 3.46              & 0.85           & 2.46              \\ 
\hline
The proposed PIE            & \textbf{3.47}     & \textbf{0.99}  & \textbf{2.73}     & 3.78              & \textbf{0.83}  & 2.16              & 3.22              & \textbf{1.32}           & \textbf{2.40}     & 2.98              & 0.58           & \textbf{1.62}     & \textbf{3.72}              & \textbf{1.23}  & \textbf{2.07}     & \textbf{3.38}     & \textbf{0.95}  & \textbf{2.44}                                         \\
\hline
\end{tabular}}
\label{tab12}
\end{table*}
\subsubsection{No-referenced IQA}
For testing using no-referenced image quality assessment (IQA), we adopt Natural Image Quality Evaluator (NIQE)~\citep{Mittal2013MakingA}, a well-known no-reference image quality assessment for evaluating image restoration without ground truth and providing quantitative comparisons. Since some work criticizes that NIQE correlates poorly with subjective human opinion, we also adopt UNIQUE~\citep{9369977} for No-referenced IQA. Smaller NIQE and larger UNIQUE indicate more naturalistic and perceptually favored quality. The NIQE and UNIQUE results on five datasets (DICM, LIME, MEF, VV, and NPE) are reported in Table~\ref{tab12}. 
Compared with other state-of-the-art methods, PIE achieves the best results for the NIQE in two of the five datasets, achieves the best results for the UNIQUE in four of the five datasets, and the average results on these five datasets are the best. 
\begin{table*}[!htb]
\centering
\caption{PSNR $\uparrow$ and SSIM $\uparrow$ on the Part2 of the SICE dataset.
}
\resizebox{2.1\columnwidth}{!}
{
\begin{tabular}{c|c|c|c|c|c|c|c|c|c|c|c|c|c|c}
\hline
Methods &LIME &R.Net &ISSR &Z.-DCE&E.GAN &RUAS&ReLLIE&U.-net&SCI&IRN&ALL-E&PairLIE&SCL-LLE& PIE\\ \hline
PSNR $\uparrow$& 13.67& 16.98  &15.01&14.78 &{17.82} &10.62 &18.17 &\textbf{20.61}&12.04&13.62&8.40&15.82&17.95&\underline{19.79}\\
SSIM $\uparrow$& 0.62& {0.66} &0.65 &0.62 &{0.66} &0.44 &\underline{0.67} & 0.66&0.64&0.52&0.30&0.65&\textbf{0.68} &\textbf{0.68}\\ 
\hline
\end{tabular}}
\label{Full}
\end{table*}
\begin{figure*}[!]
\centering
\includegraphics[width=16cm]{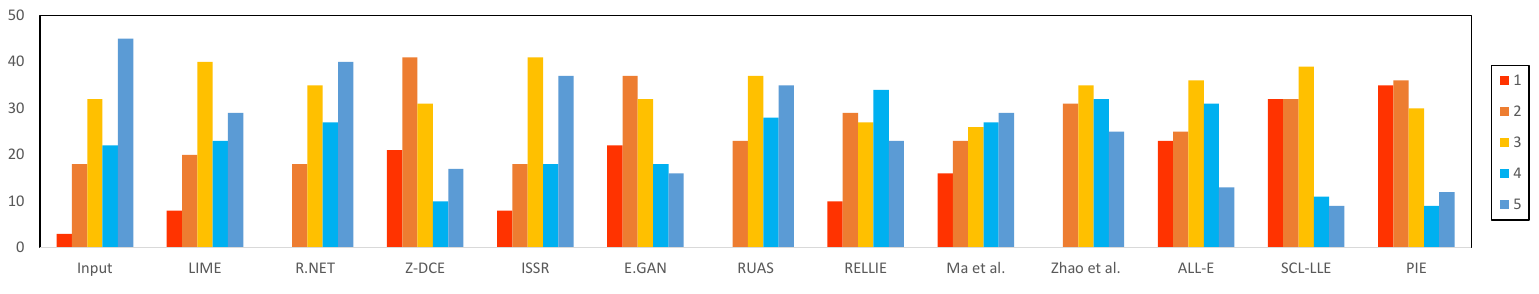}\\
\caption{The results in the human subjective survey. The color-changing from hot to cool means the quality transition from best to worst; the y-axis denotes the number of images in each ranking index.}
\label{US}
\end{figure*}
\begin{figure*}[h]
\centering
\subfigure[Input]{
\includegraphics[width=5.0cm]{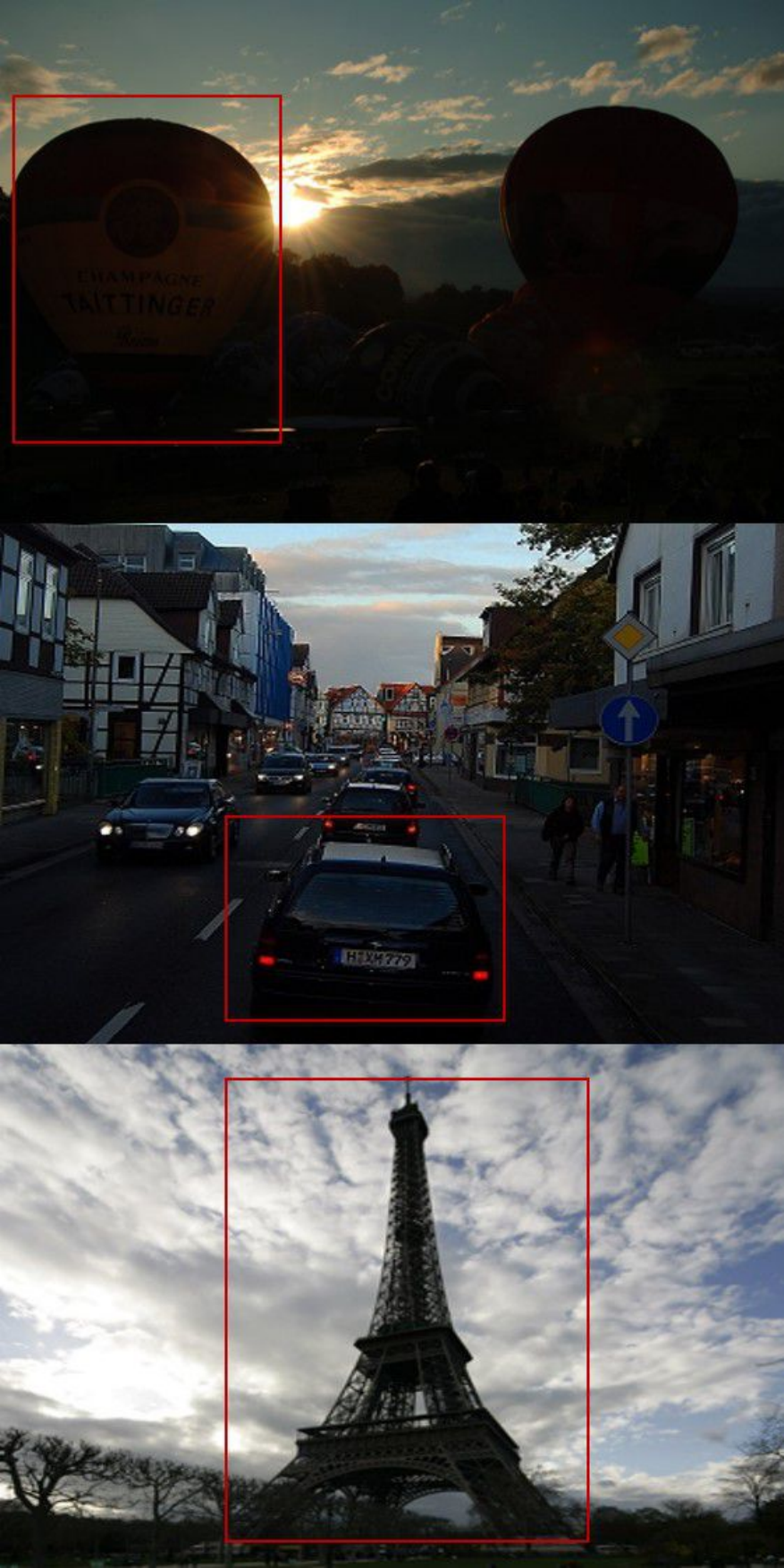}
}\hspace{-2mm}
\subfigure[w/o $L_{rc}$ ]{
\includegraphics[width=5.0cm]{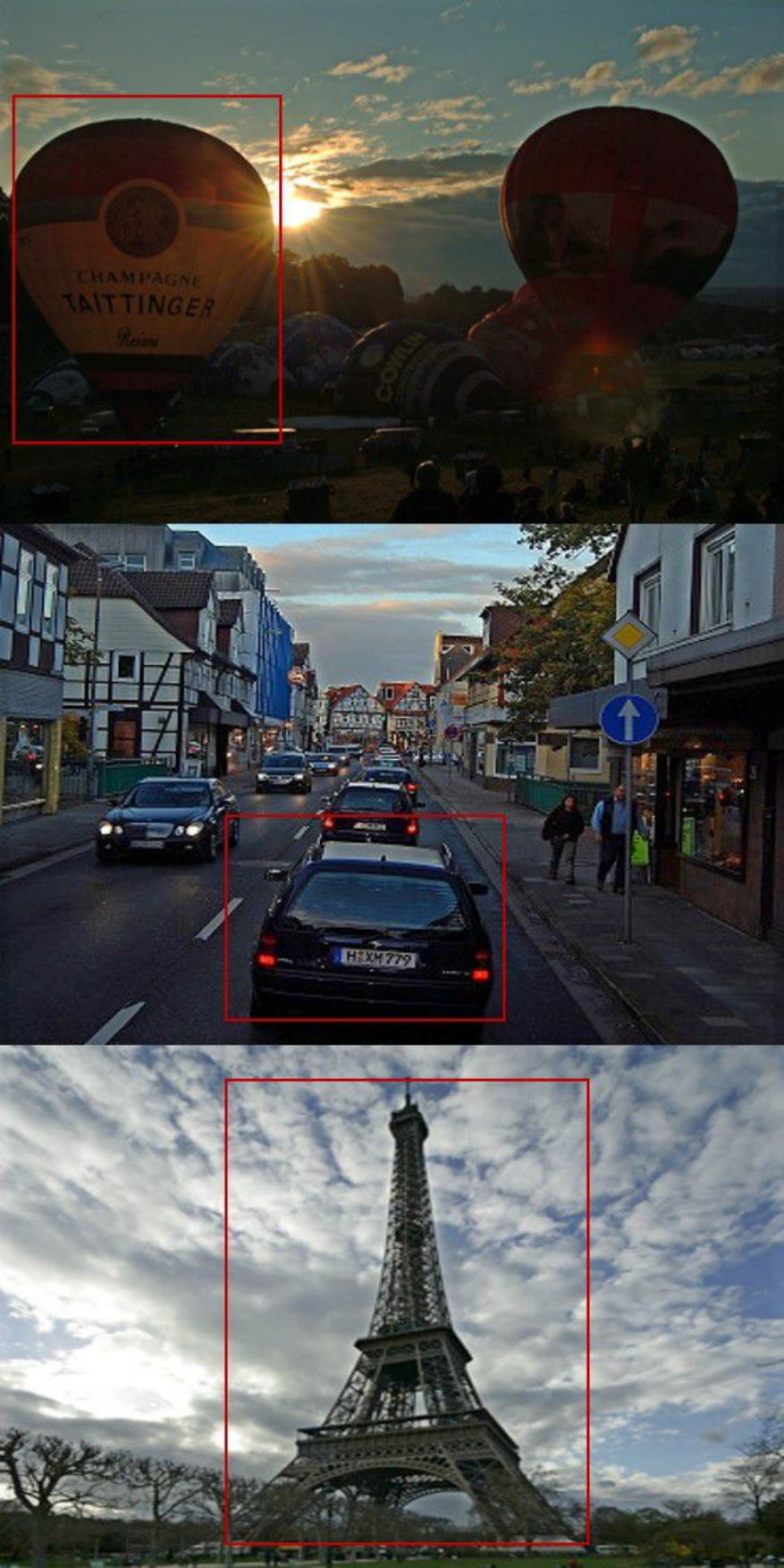}
}\hspace{-2mm}
\subfigure[w $L_{rc}$]{
\includegraphics[width=5.0cm]{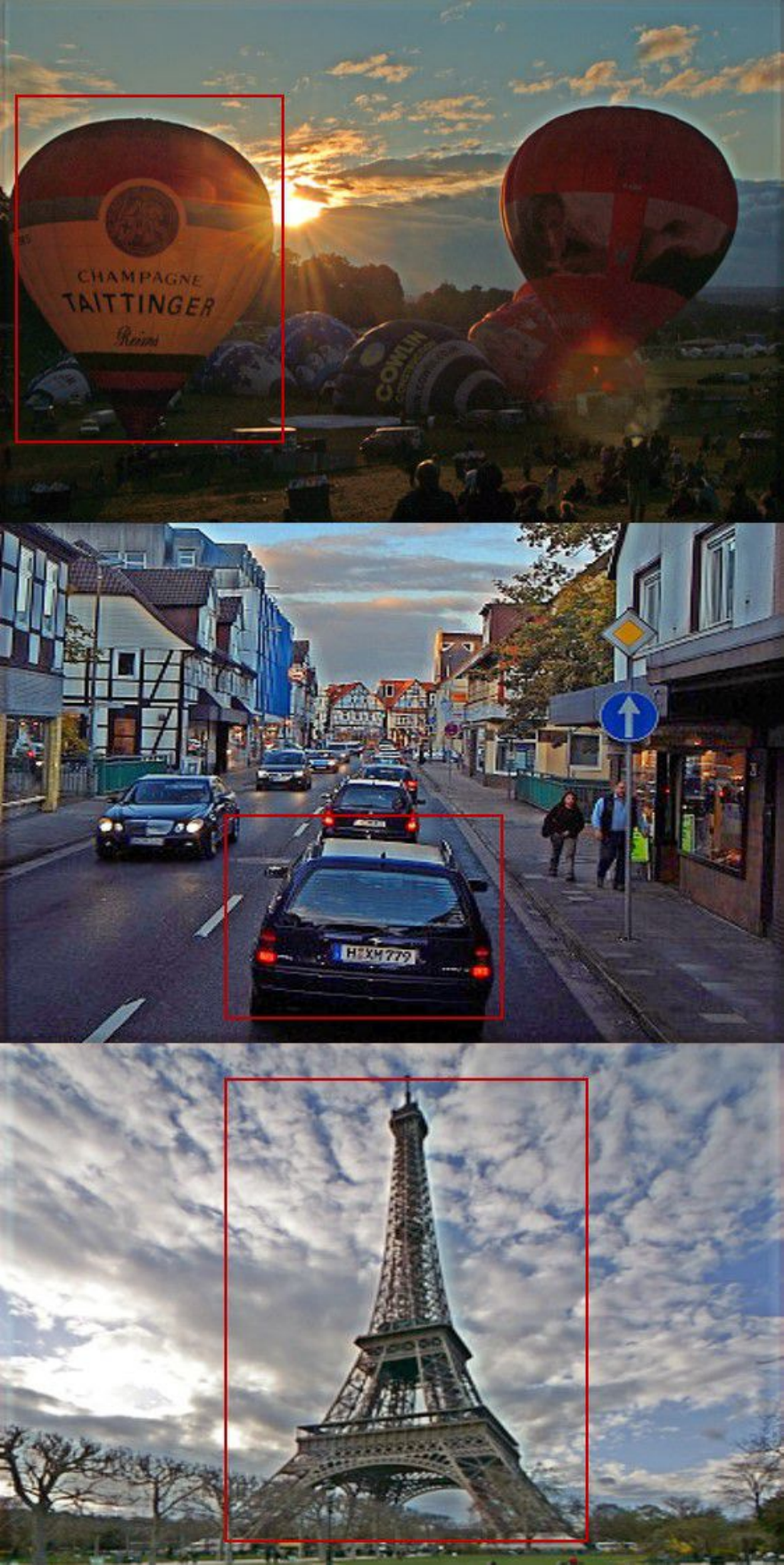}     
}\hspace{-2mm}

\caption{Ablation study on the contribution of the regional brightness consistency loss $L_{rc}$. 
 }
\label{imgseg}
\end{figure*}
\subsubsection{Full-referenced IQA}
For full-reference image quality assessment, we utilize the Peak Signal-to-Noise Ratio (PSNR, dB) and Structural Similarity (SSIM) metrics to compare the performance of various methods quantitatively. PSNR is commonly used in low-level vision tasks, and its value is always non-negative. A higher PSNR value indicates better quality. On the other hand, SSIM measures image similarity based on image brightness, contrast, and structure. Since the five datasets used in the previous test do not contain standard images, we use Part2 of the SICE dataset~\citep{Cai2018deep} without overlapping the training data. PIE demonstrates excellent performance on both PSNR and SSIM metrics, achieving the best performance on the SSIM metric and the second-highest PSNR score, only behind Uretinex-net~\citep{wu2022uretinex}, as shown in Table~\ref{Full}.

\subsubsection{Human subjective survey}
We conduct a human subjective survey (user study) for comparisons. For each image in the five test datasets (DICM, LIME, MEF, VV, and NPE) enhanced by thirteen methods (LIME, Retinex-Net, Zero-DCE, ISSR, EnlightenGAN, RUAS, ReLLIE, SCL-LLE, SCI, IRN, ALL-E, and PIE), we ask 11 human subjects to rank the enhanced images. These subjects are instructed to consider: \\
1) Whether or not the images contain visible noise. \\
2) Whether the images have overexposed or underexposed artifacts. \\
3) Whether the images show non-realistic color or texture distortion. \\%
We assign a score to each image on a scale of 1 to 5, with lower values indicating better image quality.

Each image is assigned a score ranging from 1 to 5, with lower scores indicating better image quality. The final results are presented in Table~\ref{tab12} and Fig.~\ref{US}. Among all the methods evaluated, PIE achieves the best image quality.

\begin{figure*}[h]
\centering
\subfigure[Input]{
\includegraphics[width=3.0cm]{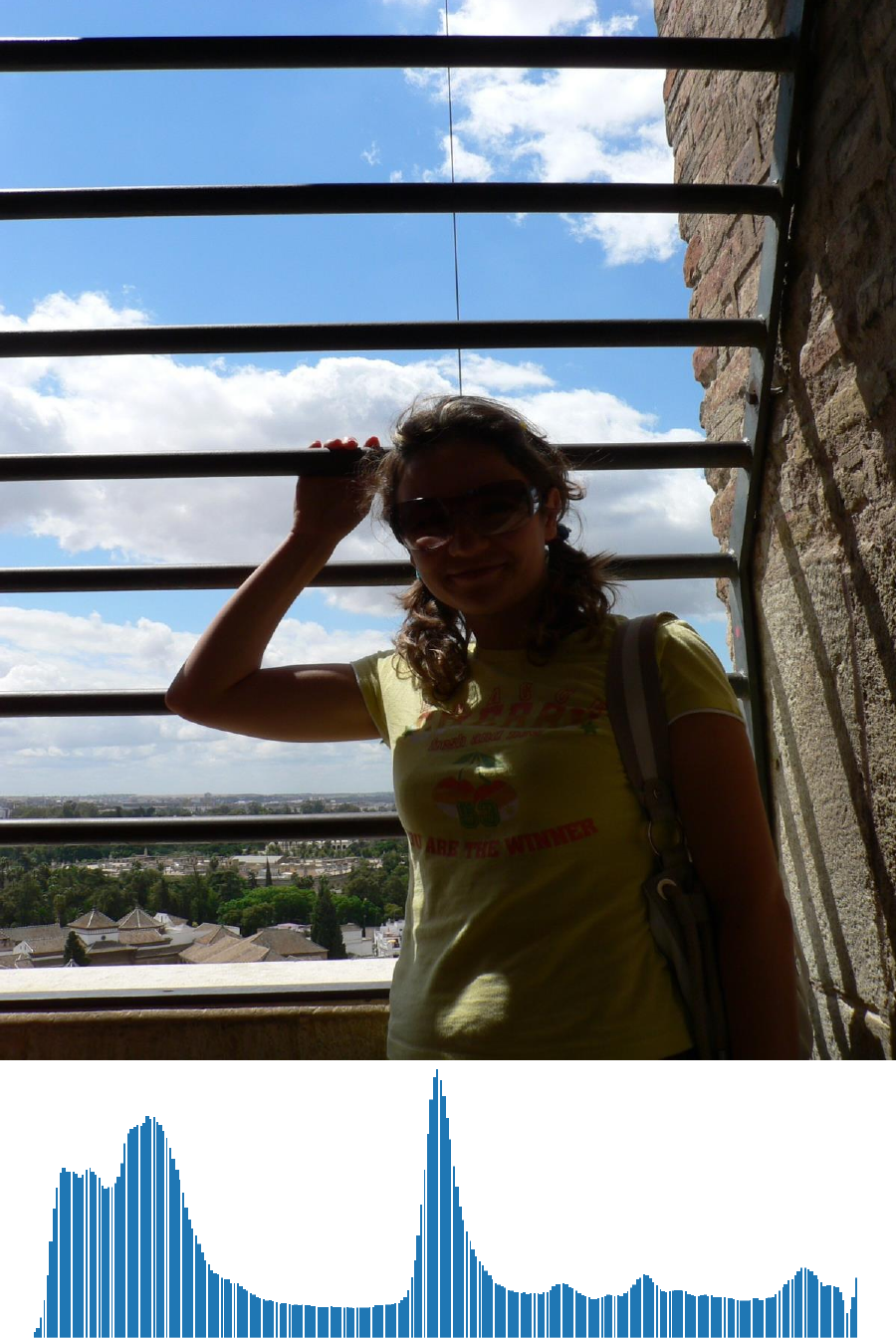}
}\hspace{-2mm}
\subfigure[w/o $L_{c}$ (SCL-LLE)]{
\includegraphics[width=3.0cm]{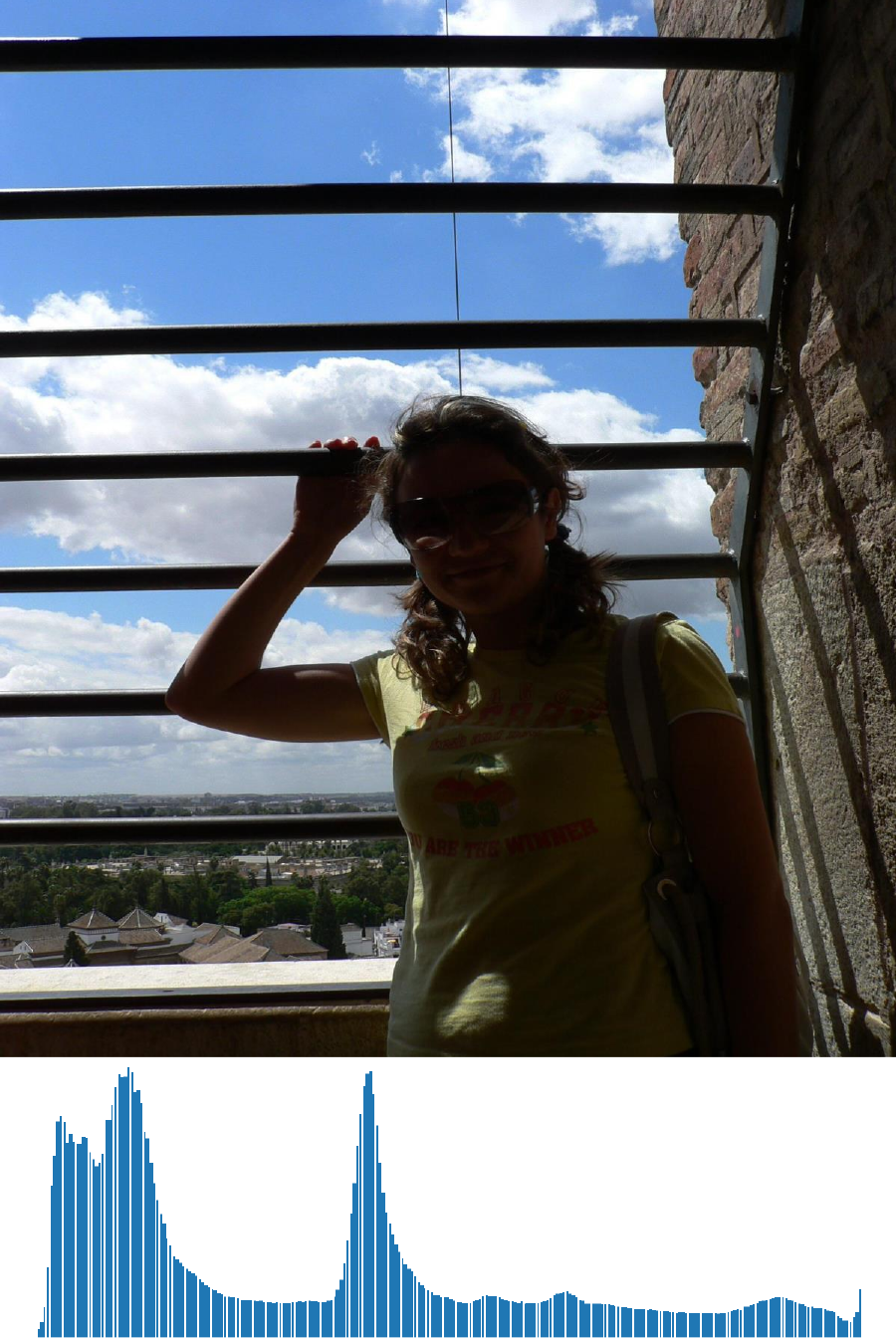}
}\hspace{-2mm}
\subfigure[w/o $L_{sc}$ (SCL-LLE)]{
\includegraphics[width=3.0cm]{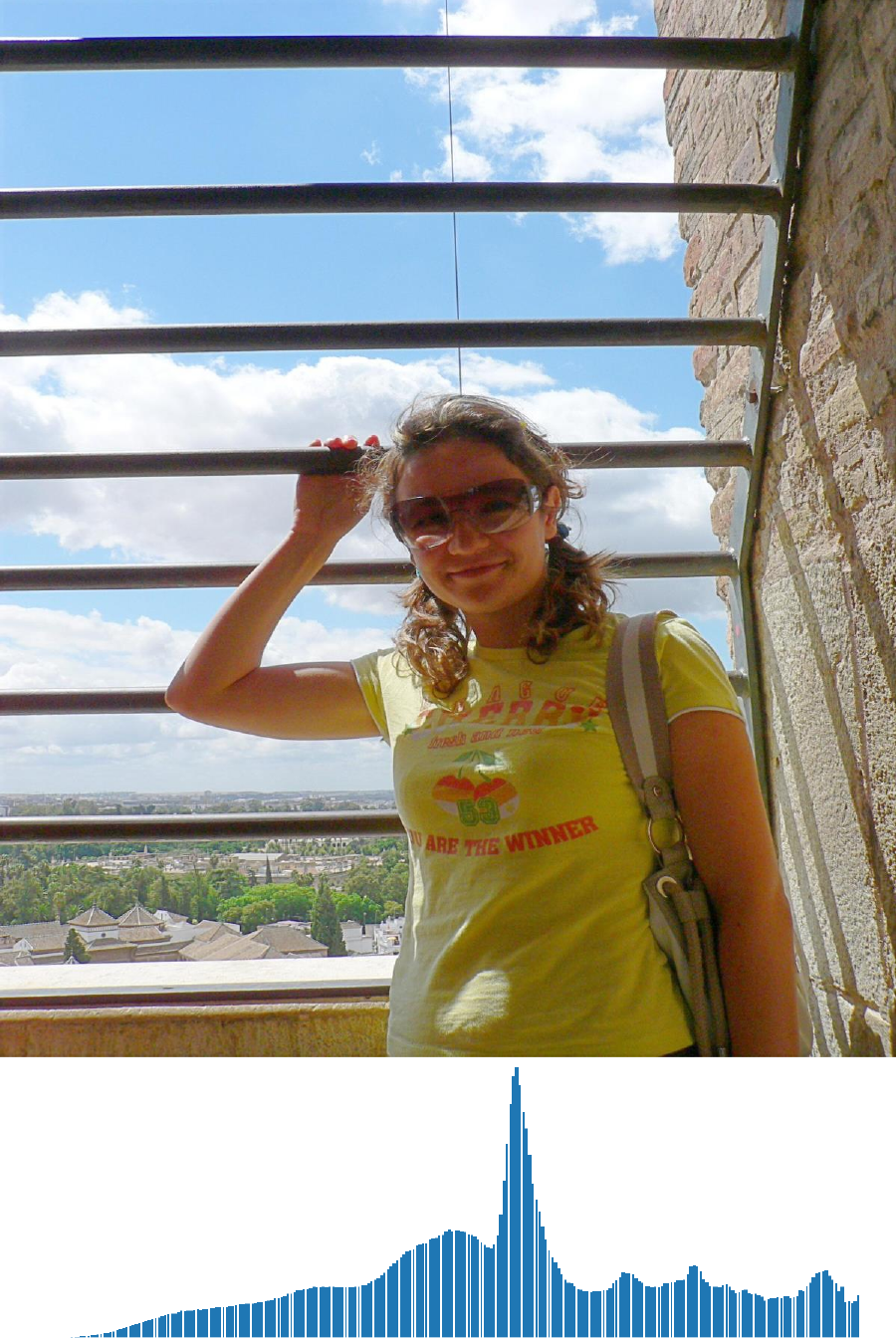}     
}\hspace{-2mm}
\subfigure[w/o $L_{fr}$ (SCL-LLE)]{
\includegraphics[width=3.0cm]{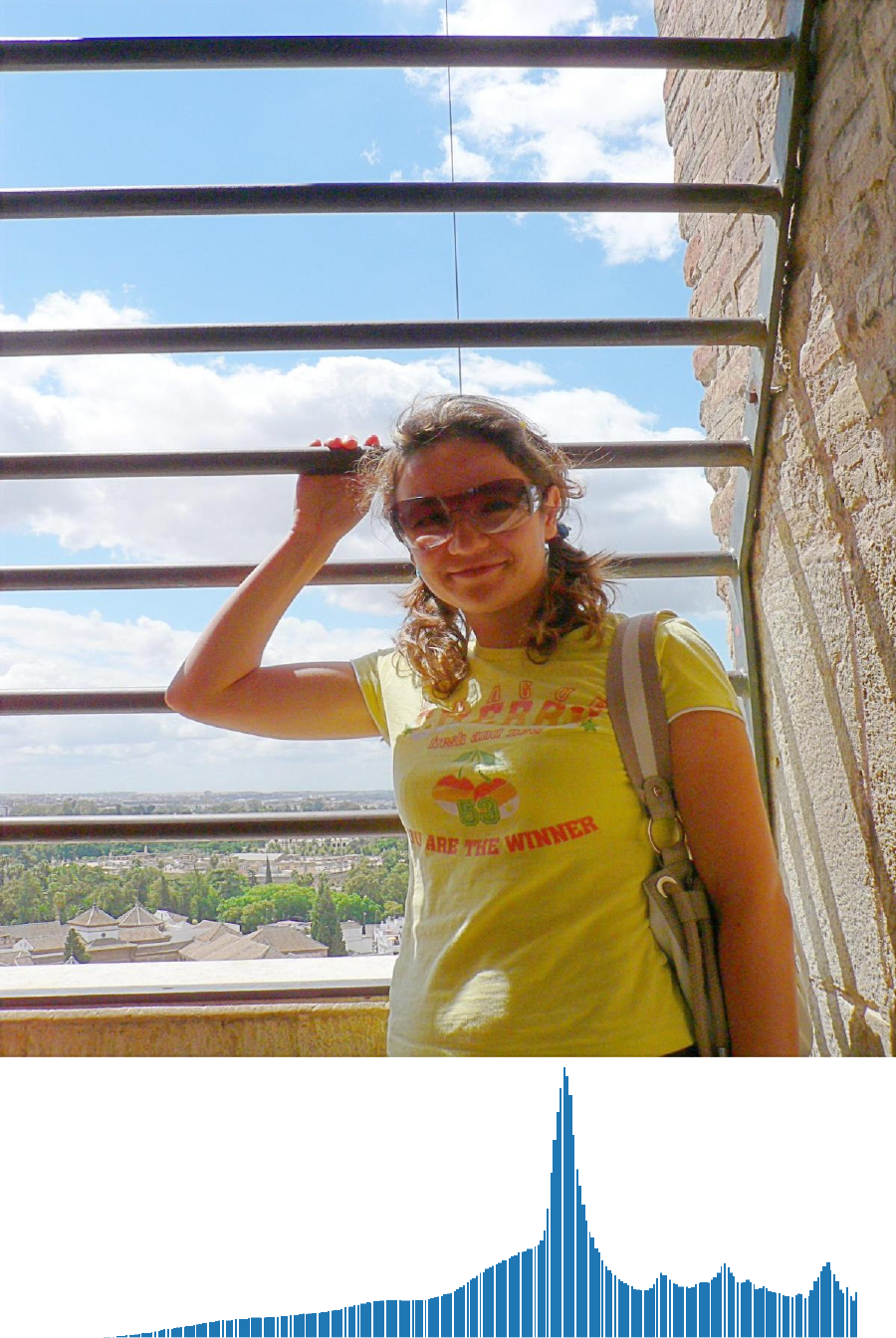}
}\hspace{-2mm}
\subfigure[SCL-LLE]{
\includegraphics[width=3.0cm]{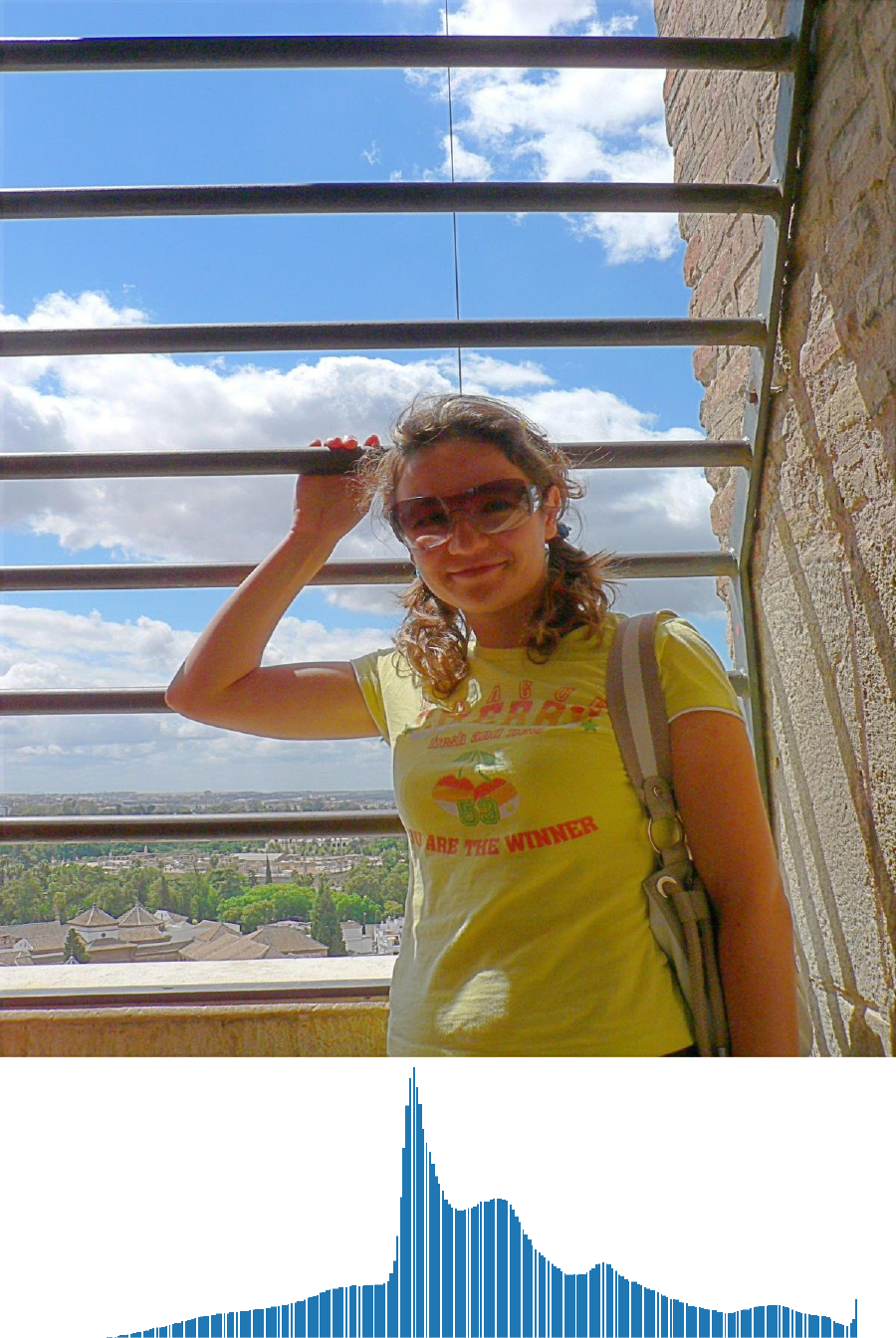}
}\hspace{-2mm}\quad
\\
\subfigure[w/o $L_{c}$ (PIE)]{
\includegraphics[width=3.0cm]{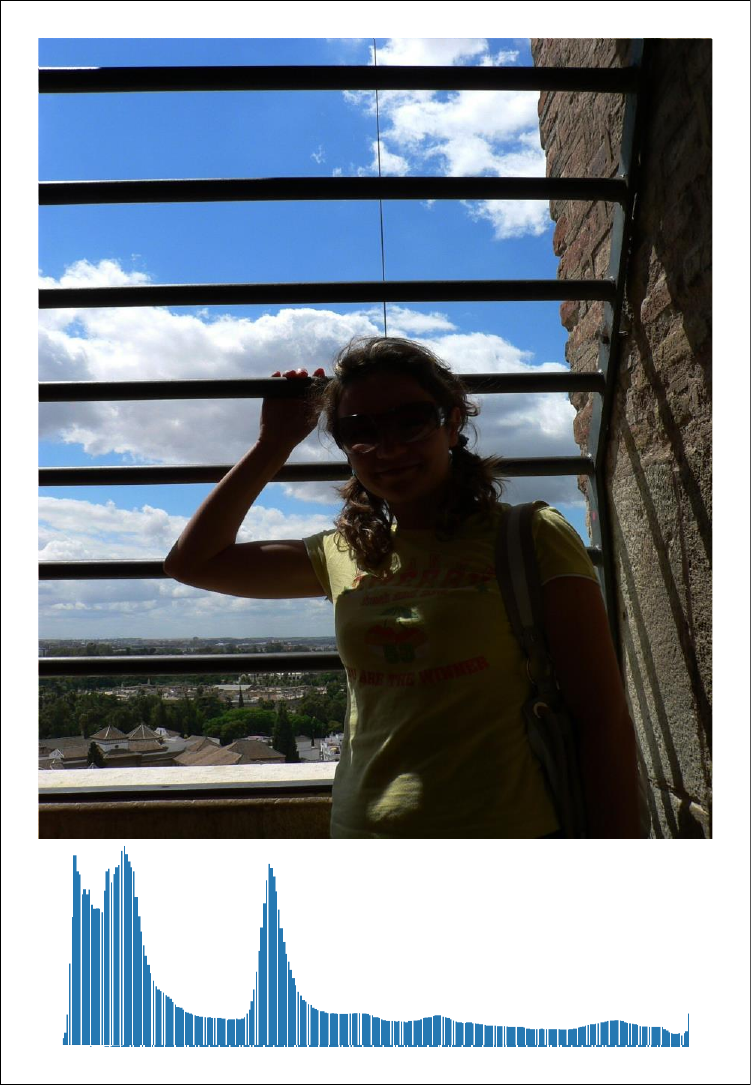}
}\hspace{-2mm}
\subfigure[w/o $L_{rc}$ (PIE)]{
\includegraphics[width=3.0cm]{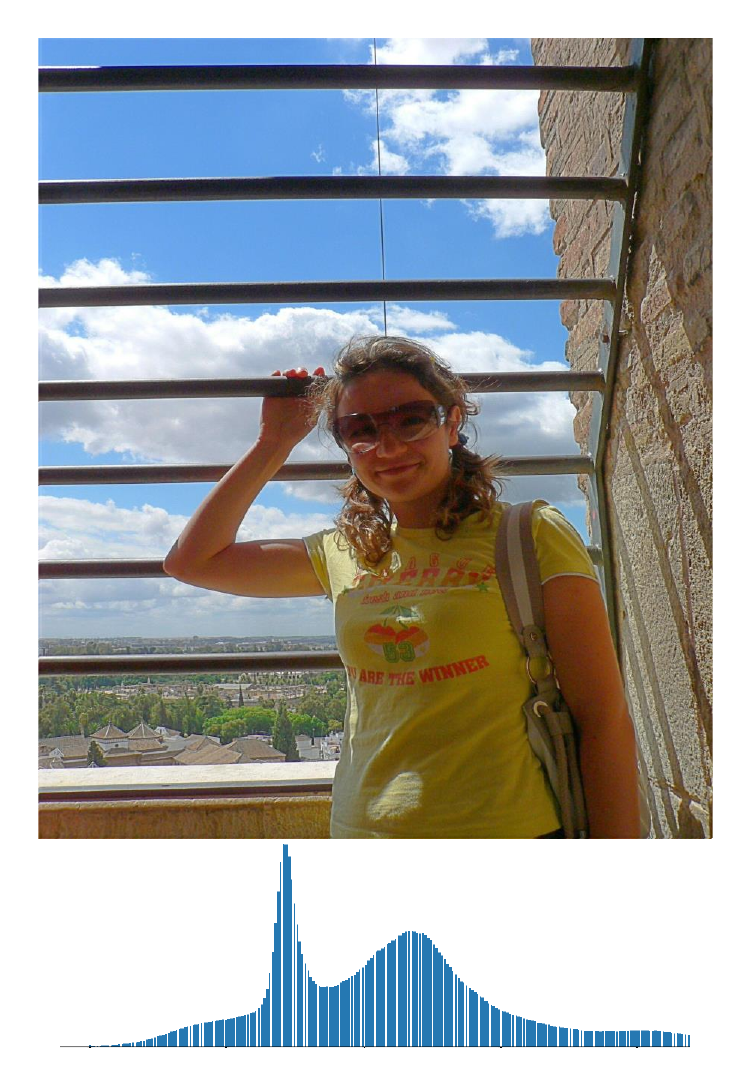}     
}\hspace{-2mm}
\subfigure[w/o $L_{fr}$ (PIE)]{
\includegraphics[width=3.0cm]{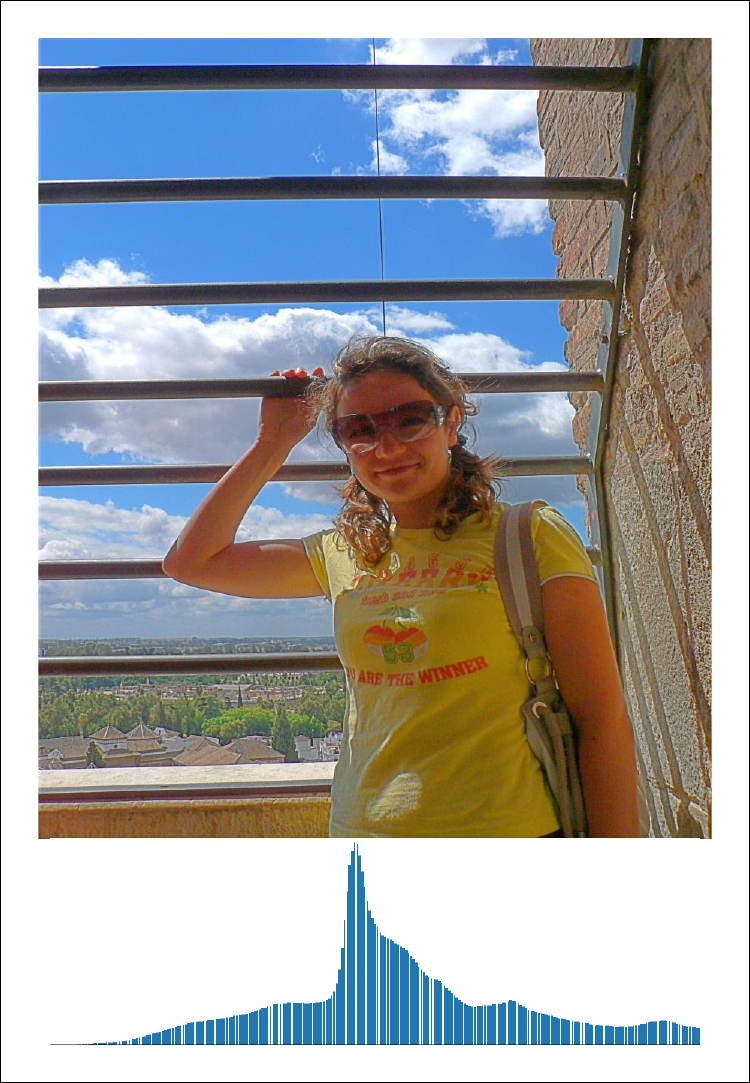}
}\hspace{-2mm}
\subfigure[PIE]{
\includegraphics[width=3.0cm]{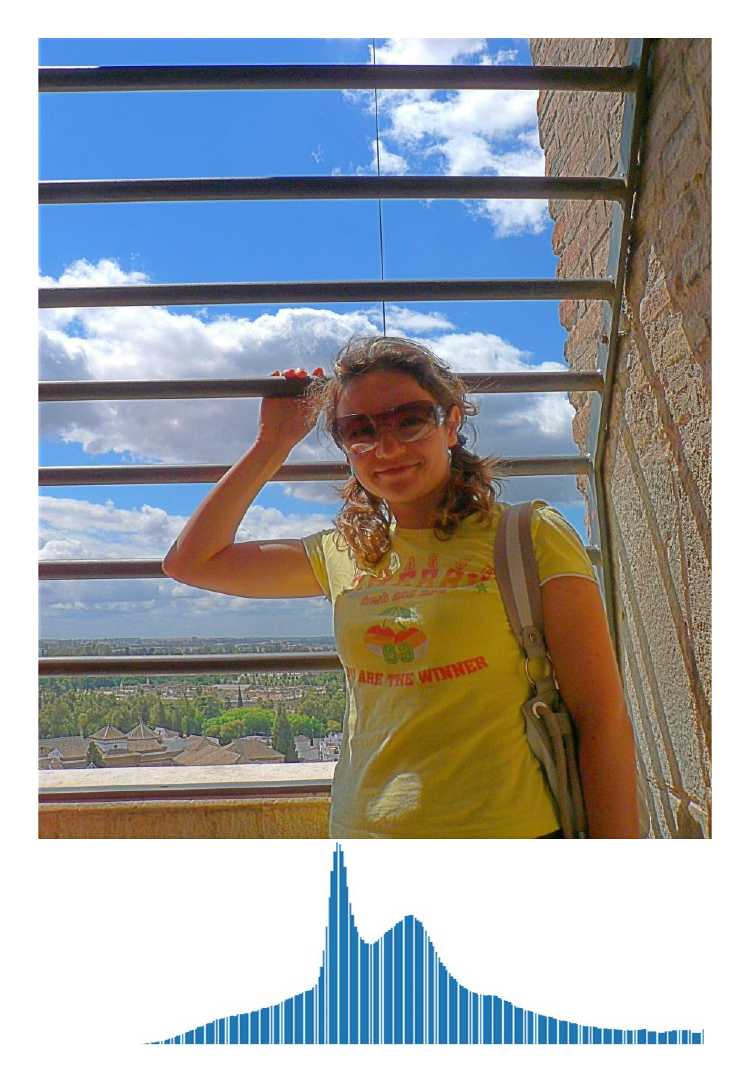}
}\hspace{-2mm}\quad
\caption{Ablation study on the contribution of each loss for PIE and SCL-LLE~\citep{liangaaai}. 
 }
\label{img2a}
\end{figure*}

\subsection{Ablation Study}
In this section, we perform ablation studies to demonstrate the effectiveness of each component of PIE. 
\subsubsection{Contribution of each loss}
In this section, we explore the impact of each loss on PIE and compare them with the loss in our baseline SCL-LLE~\citep{liangaaai}. 
We consider the color consistency item $L_{cc}$, initially proposed and tested in Zero-DCE~\citep{Guo_2020_CVPR}, as a baseline item without conducting an ablation study. Thus, we test the feature preservation loss $L_{fp}$ using the first item $L_{fr}$. 

We perform ablation studies to demonstrate the effectiveness of three different loss functions in PIE: the contrastive learning loss $L_{c}$, the feature retention loss $L_{fr}$, and the regional brightness consistency loss $L_{rc}$.
Fig.~\ref{img2a}~(f-i) shows the visualized samples with their corresponding histograms of the effects of $L_{c}$, $L_{rc}$, and $L_{fr}$ functions in PIE. In Fig.~\ref{imgseg}, the enhanced result without using the regional segmentation module (b) exhibits color deviations. On the other hand, the enhanced result with the regional segmentation module (c) demonstrates that different regions can maintain their own colors, and there is a better distinction between the foreground and background (e.g., balloon, vehicles, and tower). Table~\ref{tab1a} shows each loss's average NIQE and UNIQUE scores on five test sets. We find that the contrastive learning loss $L_{c}$ significantly controls the exposure level. The results without the regional brightness consistency loss $L_{rc}$ in PIE, and without the feature retention loss $L_{fr}$ have relatively lower contrast (\emph{e.g.}, in the sky region) than the final result.
Furthermore, we compare the performance of PIE with SCL-LLE~\citep{liangaaai}. The contrastive learning loss $L_{c}$ in PIE is more critical and effective in controlling the exposure level than in SCL-LLE. However, due to the use of an unsupervised method, the contribution of the Regional Brightness Consistency Loss $L_{rc}$ in PIE is slightly worse than the Semantic Brightness Consistency Loss $L_{sc}$ in SCL-LLE. 

The losses enhance images with fine details and more naturalistic and perceptually favored quality. The corresponding histograms show that the final losses maintain a smooth mixture-of-Gaussian-like global distribution with rare over or under-saturation areas. In contrast, the undesirable unilateral over or under-saturation areas occur in the histograms of Fig.~\ref{img2a}~(b-d) and (f).

\begin{table*}[!]
\setlength\tabcolsep{2pt} 
\centering
\caption{Ablation study. NIQE $\downarrow$ and UNIQUE (UN.) $\uparrow$ scores on the testing sets.}{
\resizebox{2\columnwidth}{!}{

\begin{tabular}{c|cc|cc|cc|cc|cc|cc}
\hline
 &\multicolumn{2}{c|}{DICM} &\multicolumn{2}{c|}{LIME} &\multicolumn{2}{c|}{MEF} &\multicolumn{2}{c|}{VV} &\multicolumn{2}{c|}{NPE} &\multicolumn{2}{c}{Average} \\ 
Methods &NIQE$\downarrow$ &UN.$\uparrow$ &NIQE$\downarrow$ &UN.$\uparrow$ &NIQE$\downarrow$ &UN.$\uparrow$ &NIQE$\downarrow$ &UN.$\uparrow$ &NIQE$\downarrow$ &UN.$\uparrow$ &NIQE$\downarrow$ &UN.$\uparrow$ \\ \hline
Input                      & 4.26          & 0.72          & 4.36          & 0.70          & 4.26          & 0.72          & 3.52          & \textbf{0.74} & 4.32          & 1.17          & 4.13          & 0.75          \\
w/o $L_c$ (SCL-LLE)        & 4.31          & 0.64          & 4.36          & 0.57          & 4.25          & 0.56          & 4.10          & 0.70          & 4.28          & 1.02          & 4.27          & 0.66          \\
w/o $L_{sc}$ (SCL-LLE)       & 3.53          & 0.83          & 3.85          & 0.76          & 3.32          & 1.18          & 3.21          & 0.50          & 3.98          & 1.02          & 3.49          & 0.82          \\
w/o $L_{fr}$ (SCL-LLE)       & 3.54          & 0.80          & 3.88          & 0.71          & 3.32          & 1.22          & 3.18          & 0.47          & 3.97          & 1.03          & 3.50          & 0.80          \\
\hline
w/o $L_c$ (PIE)          & 4.57          & 0.67          & 4.54          & 0.52          & 4.61          & 0.52          & 3.64          & 0.72          & 4.34          & 1.02          & 4.38          & 0.66          \\
w/o $L_{rc}$ (PIE)         & 3.54          & 0.94          & 3.85          & 0.76          & 3.32          & 1.32          & 3.01          & 0.54          & 3.87          & 1.02          & 3.45          & 0.90          \\
w/o $L_{fr}$ (PIE)         & 3.50          & 0.94          & 4.26          & 0.79          & 3.63          & 1.24          & 3.03          & 0.58          & 4.08          & 1.09          & 3.53          & 0.91          \\
\hline
w/o Neg. samples (PIE)   & 3.55          & 0.81          & 3.84          & 0.72          & 3.36          & 1.14          & 3.14          & 0.38          & 3.95          & 1.01          & 3.49          & 0.78          \\
w/o overexp. Neg. (PIE)  & 3.59          & 0.75          & 3.91          & 0.59          & 3.36          & 1.24          & 3.15          & 0.37          & 4.12          & 0.87          & 3.54          & 0.74          \\
w/o underexp. Neg. (PIE) & 4.58          & 0.57          & 4.52          & 0.48          & 4.69          & 0.46          & 3.58          & 0.65          & 4.36          & 0.86          & 4.38          & 0.58          \\
\hline
Gamma curves only       & 3.52          & 0.96          & 3.79          & 0.81          & \textbf{3.22} & 1.19          & 3.01          & 0.63          & 3.75          & 1.18          & 3.39 & \textbf{0.95} \\
Sigmoid curves only     & 3.51          & 0.97          & \textbf{3.77} & 0.82          & 3.31          & 1.19          & 3.10          & 0.61          & 3.80          & \textbf{1.23} & 3.40          & 0.93          \\
Logarithmic curves only& 3.52          & 0.97          & 3.86          & 0.79          & 3.32          & \textbf{1.33} & 3.14          & 0.63          & 3.84          & 1.18          & 3.47          & 0.94          \\
\hline
SCL-LLE                    & 3.51          & 0.87          & \textbf{3.78} & 0.76          & 3.31          & 1.25          & 3.16          & 0.49          & 3.88          & 1.08          & 3.46          & 0.85          \\
PIE                      & \textbf{3.47} & \textbf{0.99} & \textbf{3.78} & \textbf{0.83} & \textbf{3.22} & \textbf{1.32} & \textbf{2.98} & 0.58          & \textbf{3.72} & \textbf{1.23} & \textbf{3.38} & \textbf{0.95} 
      \\      
\hline
\end{tabular}}}
\label{tab1a}
\end{table*}

\subsubsection{Contribution of the curves in BoC}
In the BoC method, we apply three types of curves - Gamma, Sigmoid, and Logarithmic to adjust the brightness of images. To investigate the contribution of each curve to our method, we separately use each type of curve to generate over/underexposed images.  
As shown in Table~\ref{tab1a}, all three types of curves can produce over/underexposed images as negative samples for contrastive learning, which can help the model better learn the features of the data. 



\begin{figure*}[h]
\centering
\subfigure[Input]{
\includegraphics[width=3.55cm]{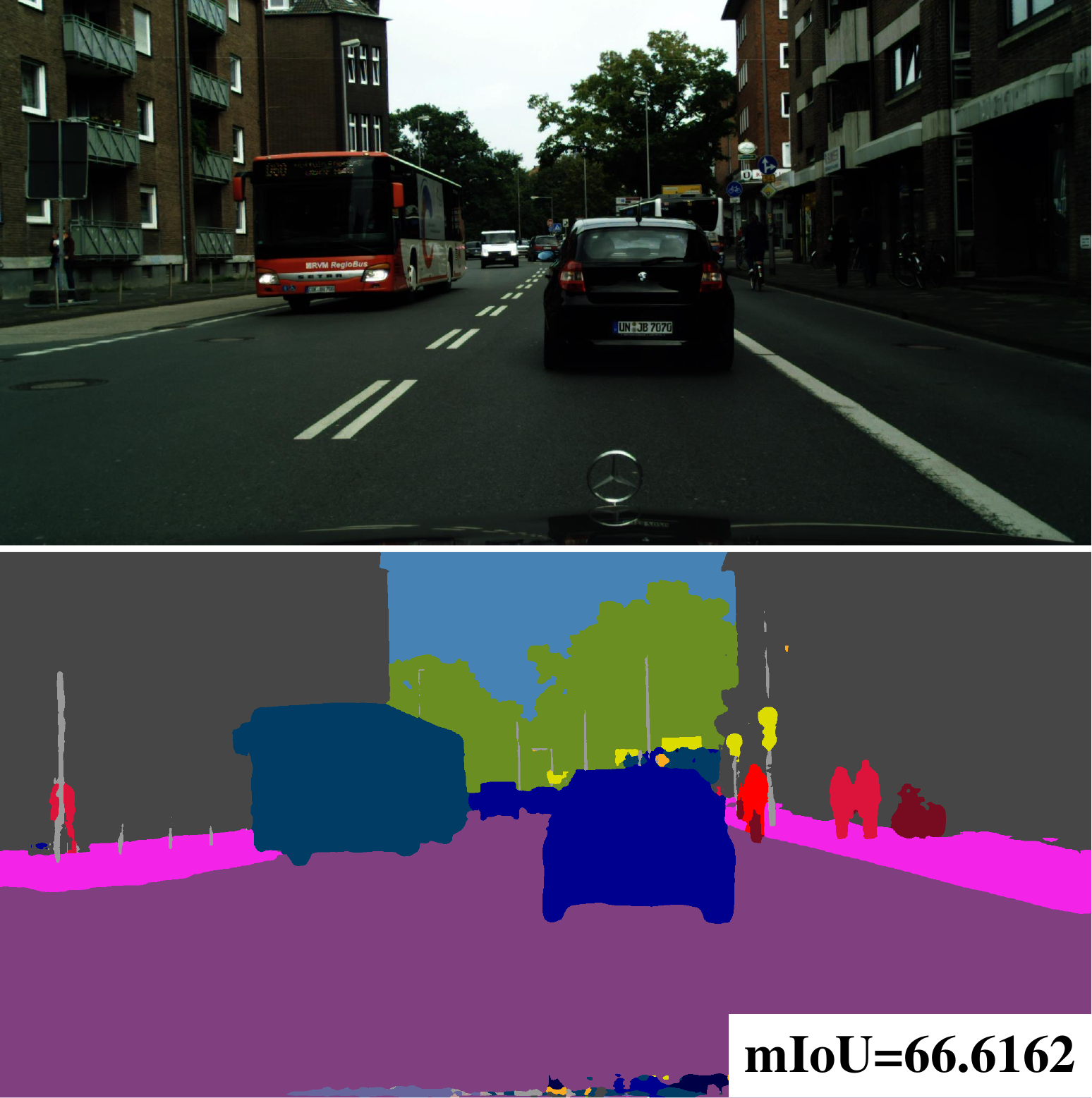}
}
\subfigure[LIME]{
\includegraphics[width=3.55cm]{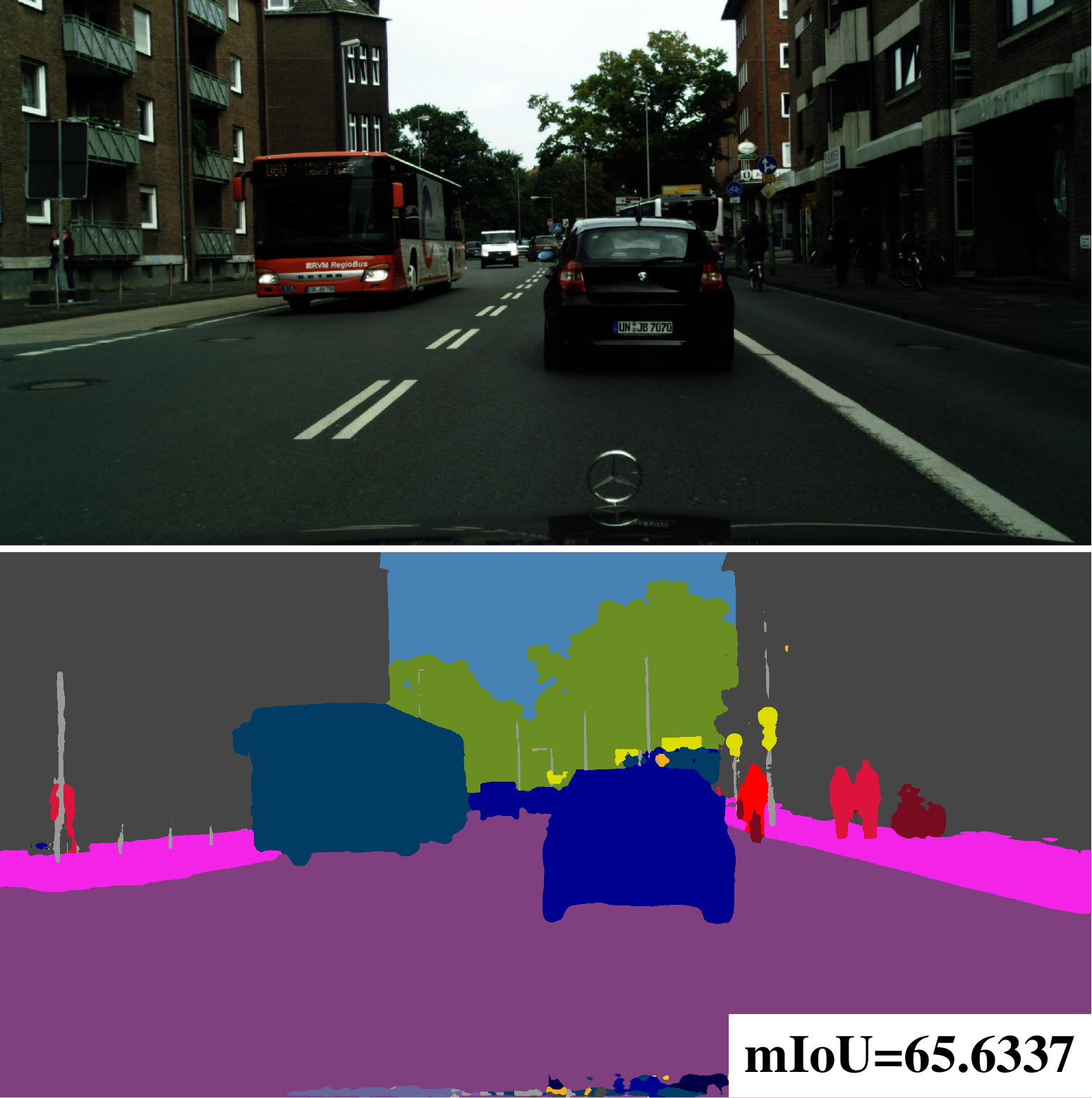}
}
\subfigure[RetinexNet]{
\includegraphics[width=3.55cm]{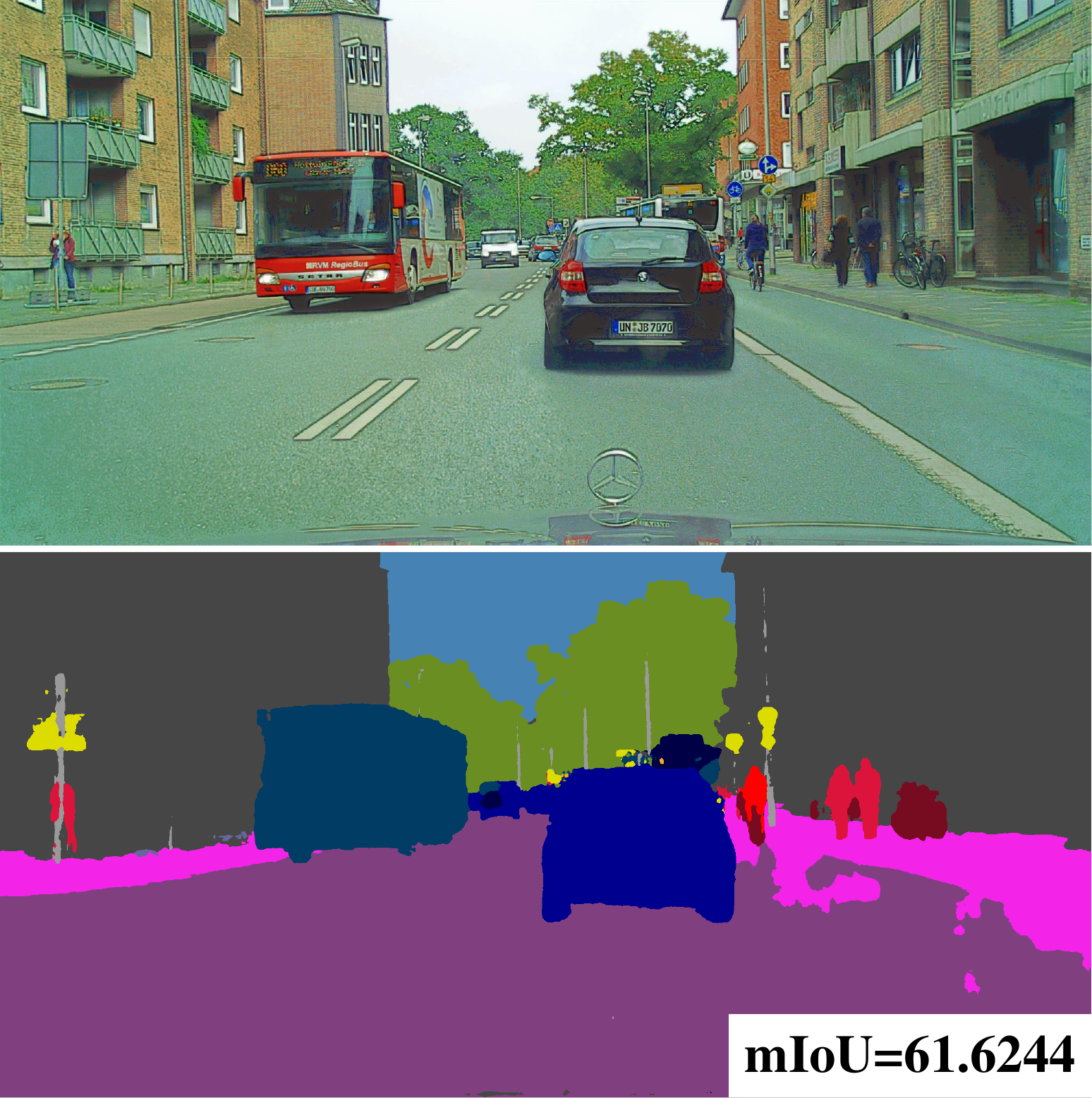}
}
\subfigure[ISSR]{
\includegraphics[width=3.55cm]{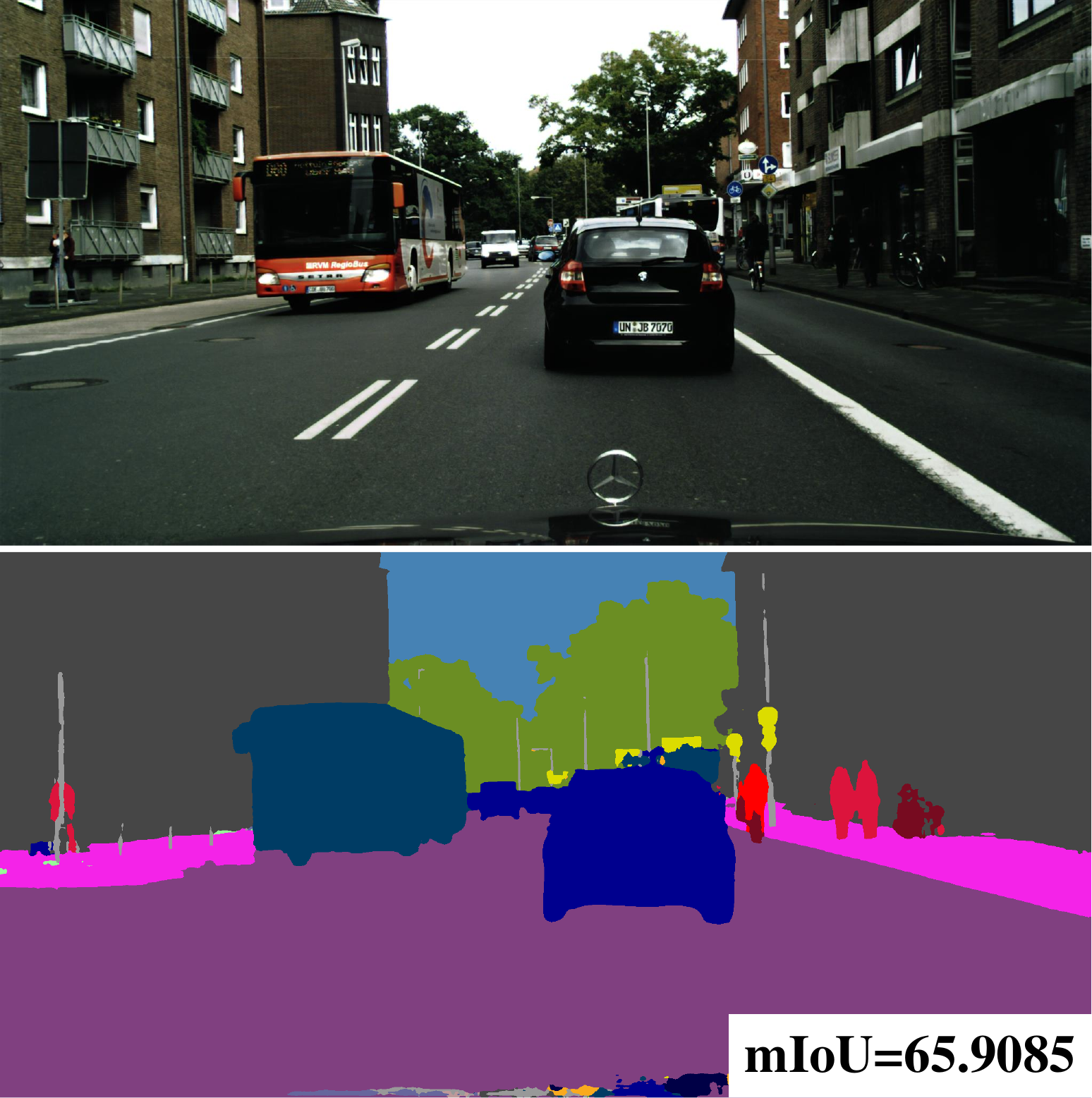}
}\\
\subfigure[Zero-DCE]{
\includegraphics[width=3.55cm]{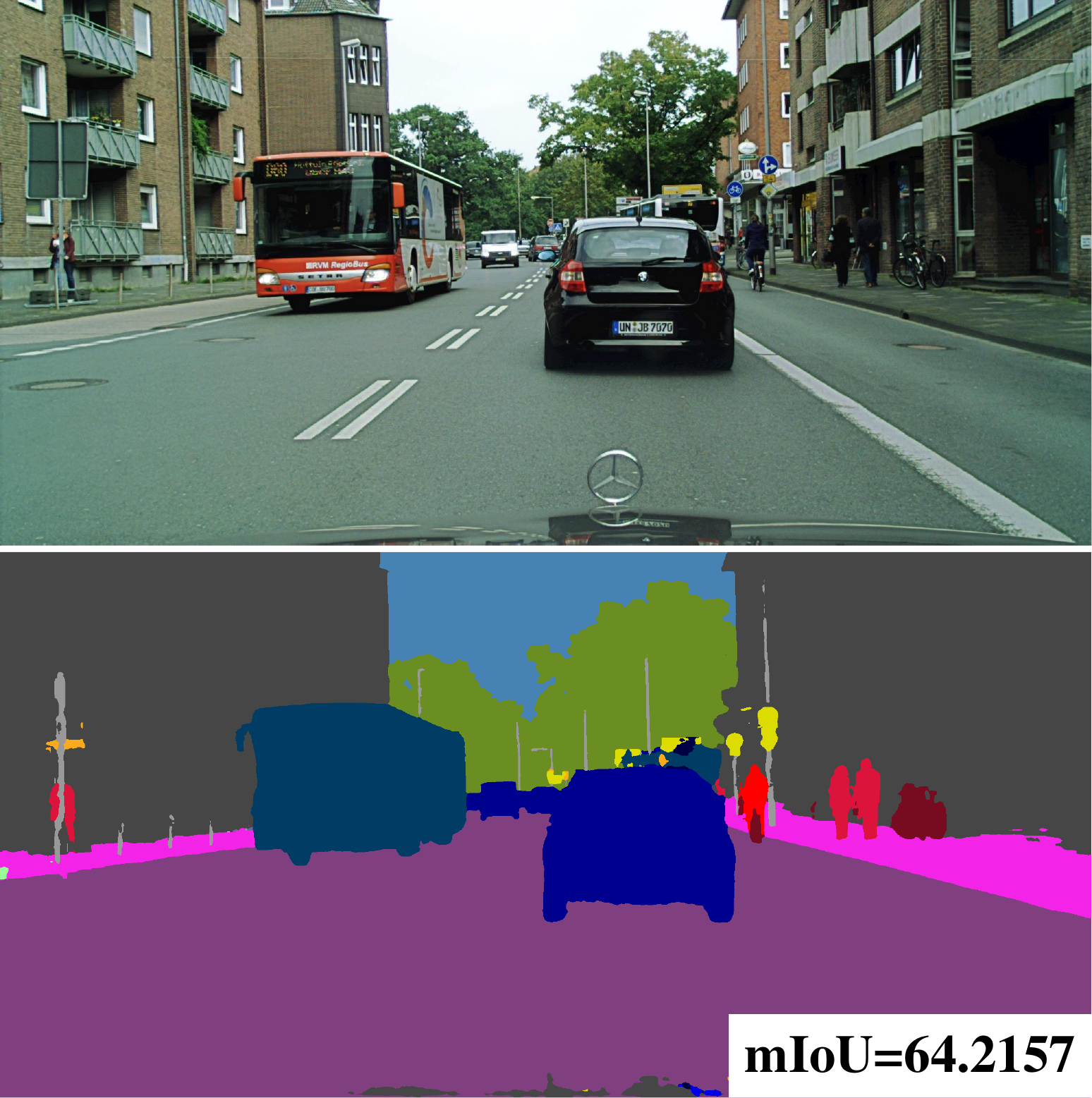}
}
\subfigure[EnlightenGAN]{
\includegraphics[width=3.55cm]{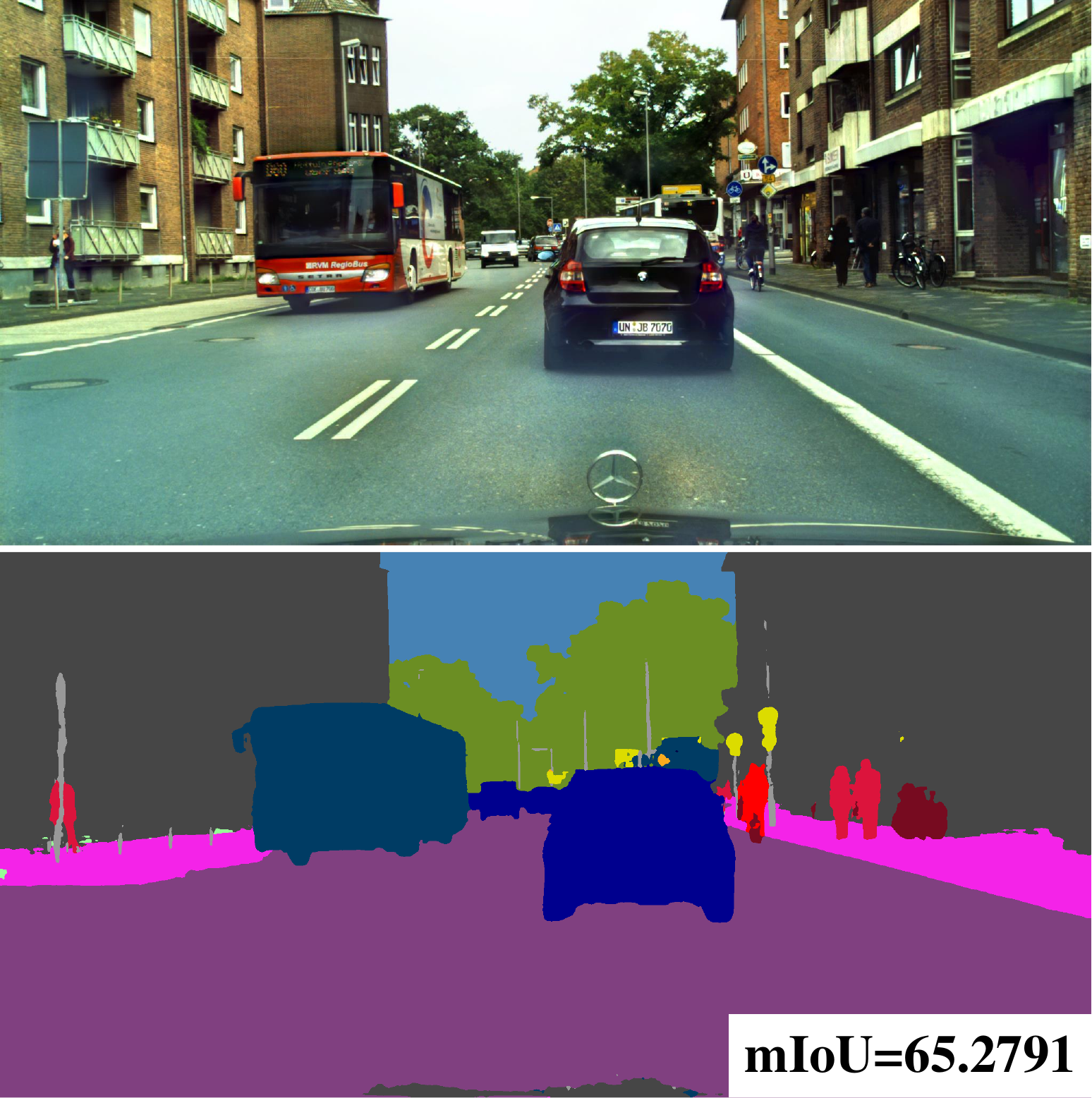}
}
\subfigure[RUAS]{
\includegraphics[width=3.55cm]{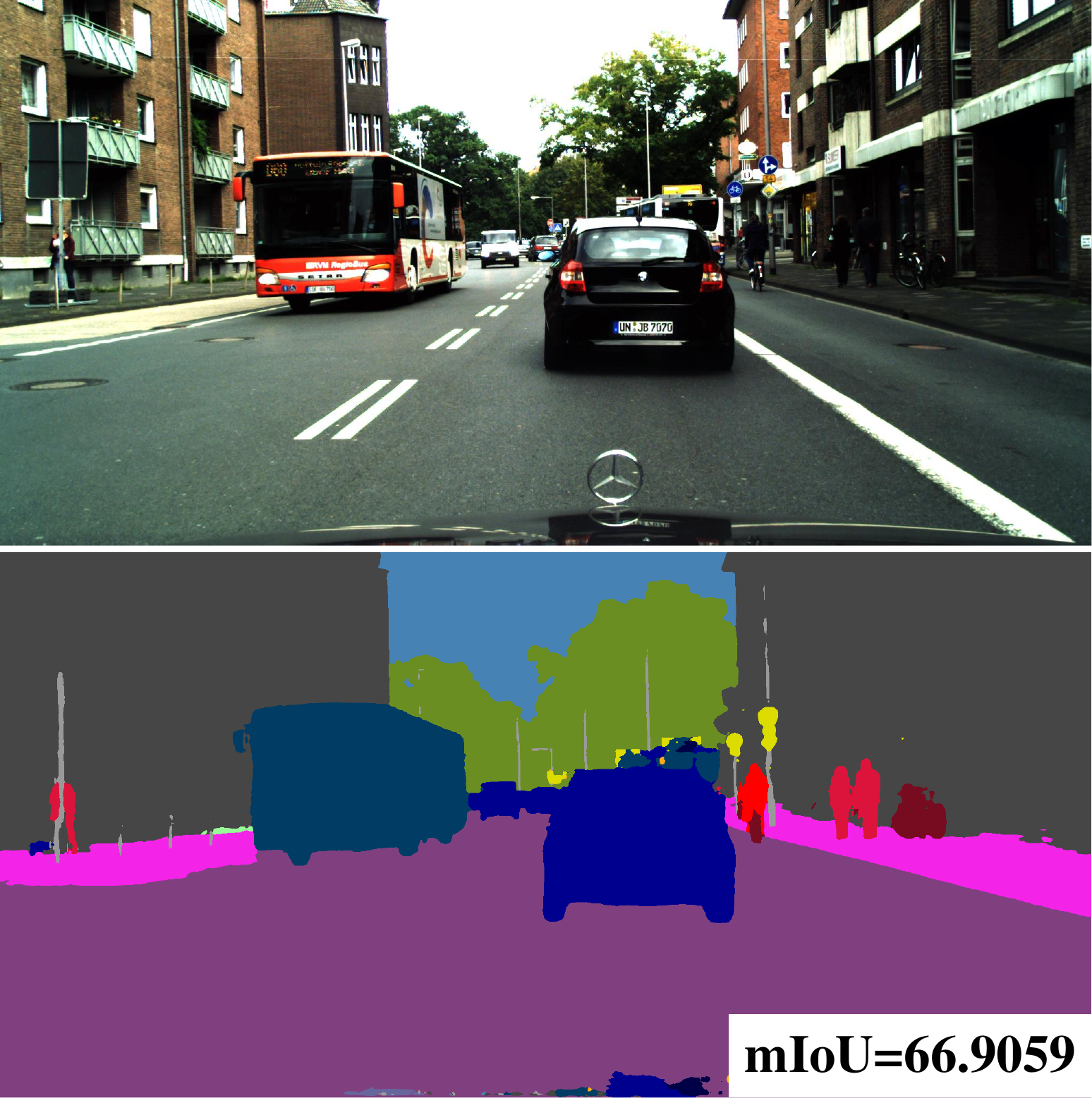}
}
\subfigure[Ours]{
\includegraphics[width=3.55cm]{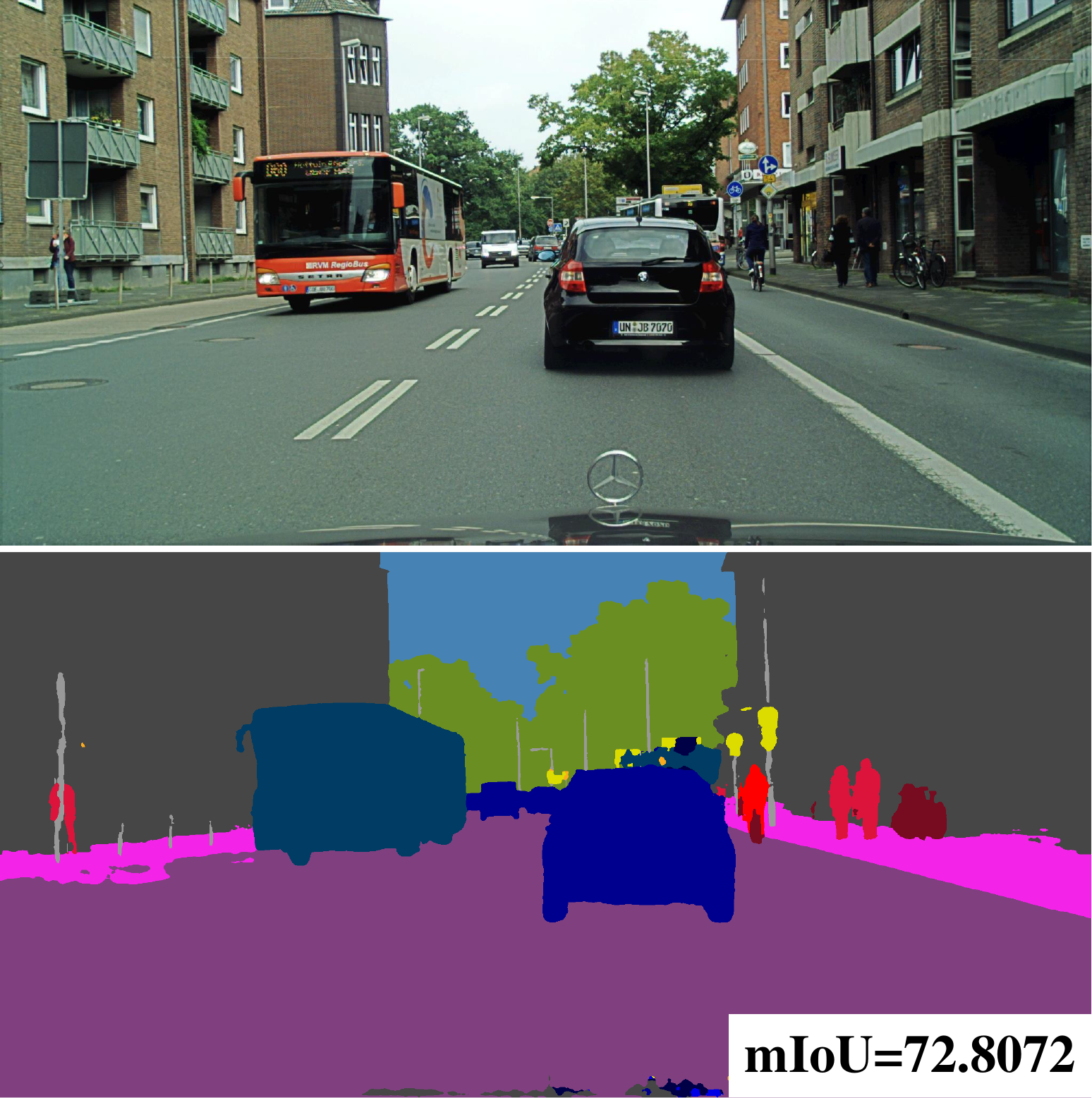}
}

\caption{Visual comparison of seven LLE methods for semantic segmentation.}
\label{seg}
\end{figure*}

\begin{table*}[!]
\centering
\caption{Discussion on the contrastive loss. The NIQE and UNIQUE scores are the average results on five datasets: DICM, LIME, MEF, VV, and NPE. The PSNR and SSIM scores are the results on the Part2 of the SICE dataset.
}{
\begin{adjustbox}{max width=1\textwidth}
\begin{tabular}{ccc|ccc|cccc}
\hline
 \multicolumn{3}{c|}{$L_{cG}$}
 &\multicolumn{3}{c|}{$L_{cE}$} &\multicolumn{4}{c}{$L_{c}$ = $L_{cG}$ + $L_{cE}$}  \\ 

\hline
Triple &   N-pair &  InfoNCE & Triple &   N-pair &  InfoNCE &NIQE$\downarrow$ &UN.$\uparrow$ &PSNR$\uparrow$ &SSIM$\uparrow$   \\ \hline
\checkmark &   &  &\checkmark & & &{3.44} &{0.93} &{18.44} &{0.67} \\ 
\checkmark &   &  & &\checkmark & &{3.65} &{0.83} &{15.08} &{0.62}  \\
\checkmark &   &  & & &\checkmark &\textbf{3.38} &\textbf{0.95} &\textbf{19.79} &\textbf{0.68}  \\
\hline
 &\checkmark   &  &\checkmark & & &{4.26} &{0.67} &{10.69} &{0.44}  \\
 & \checkmark   &  & &\checkmark & &{5.04} &{0.43} &{8.49} &{0.28}  \\
 &\checkmark   &  & & &\checkmark &{4.16} &{0.41} &{12.19} &{0.49}  \\
 \hline
 &   &\checkmark  &\checkmark & & &{5.13} &{0.43} &{8.37} &{0.27}  \\
 &   &\checkmark  & &\checkmark & &{5.34} &{0.21} &{7.26} &{0.18}  \\
 &   &\checkmark  & & &\checkmark &{4.94} &{0.15} &{9.28} &{0.31}  \\

\hline
\end{tabular}
\end{adjustbox}}

\label{tab1eloss}
\end{table*}
\begin{table*}[!]
\centering
\caption{ Comparisons of different positive and negative sample rates. The baseline is PIE with the rate of 1:1.}{
\begin{adjustbox}{max width = 1\textwidth}
\begin{tabular}{cc|cc|cc|cc|cc|cc|cc|c}
\hline
 &
 &\multicolumn{2}{c|}{DICM} &\multicolumn{2}{c|}{LIME} &\multicolumn{2}{c|}{MEF} &\multicolumn{2}{c|}{VV} &\multicolumn{2}{c|}{NPE} &\multicolumn{2}{c|}{Average}&\multicolumn{1}{c}{Average training} \\ 
Positive &  Negative &NIQE$\downarrow$ &UN.$\uparrow$ &NIQE$\downarrow$ &UN.$\uparrow$ &NIQE$\downarrow$ &UN.$\uparrow$ &NIQE$\downarrow$ &UN.$\uparrow$ &NIQE$\downarrow$ &UN.$\uparrow$ &NIQE$\downarrow$ &UN.$\uparrow$ &time/epoch (min.)\\ \hline
1        & 1        & 3.47          & \textbf{0.99} & 3.78          & 0.83          & \textbf{3.22} & 1.32          & 2.98          & \textbf{0.58} & \textbf{3.72} & \textbf{1.23} & 3.38          & \textbf{0.95} & 73.6\\
1        & 5        & 3.47          & \textbf{0.99} & \textbf{3.74} & \textbf{0.85} & \textbf{3.22} & \textbf{1.33} & \textbf{2.96} & 0.57          & 3.74          & 1.20          & \textbf{3.37} & \textbf{0.95} & 131.3\\
5        & 1        & \textbf{3.38} & 0.93          & 3.90          & 0.71          & 3.30          & 1.31          & 3.03          & 0.48          & 3.91          & 1.04          & 3.46          & 0.90          & 126.8\\
5        & 5        & \textbf{3.38} & 0.96          & 3.86          & 0.80          & 3.28          & 1.32          & 2.99          & 0.53          & 3.89          & 1.09          & 3.43          & 0.86          & 158.6\\
\hline
\label{tab1erate}
\end{tabular}
\end{adjustbox}}
\end{table*}
\begin{figure}[h]
\centering
\includegraphics[width=7.5cm]{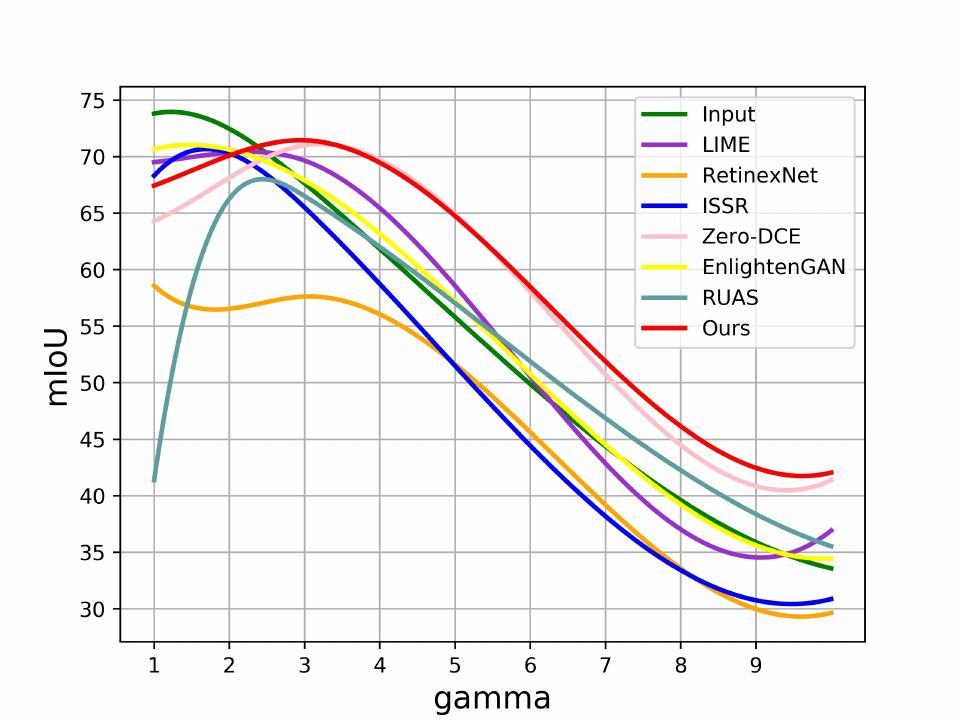}
\caption{The semantic segmentation results of the input low-light images after enhancement. When using the original input ($\gamma$=1), the semantic segmentation with all the enhancement models could not surpass the initial input. When $\gamma$ becomes larger, the mIoU of segmentation after using our method has been significantly better than those using the original images.}
\label{img9}
\end{figure}
\begin{figure}[h]
\centering
\subfigure[RetinaFace]{
\includegraphics[width=7cm]{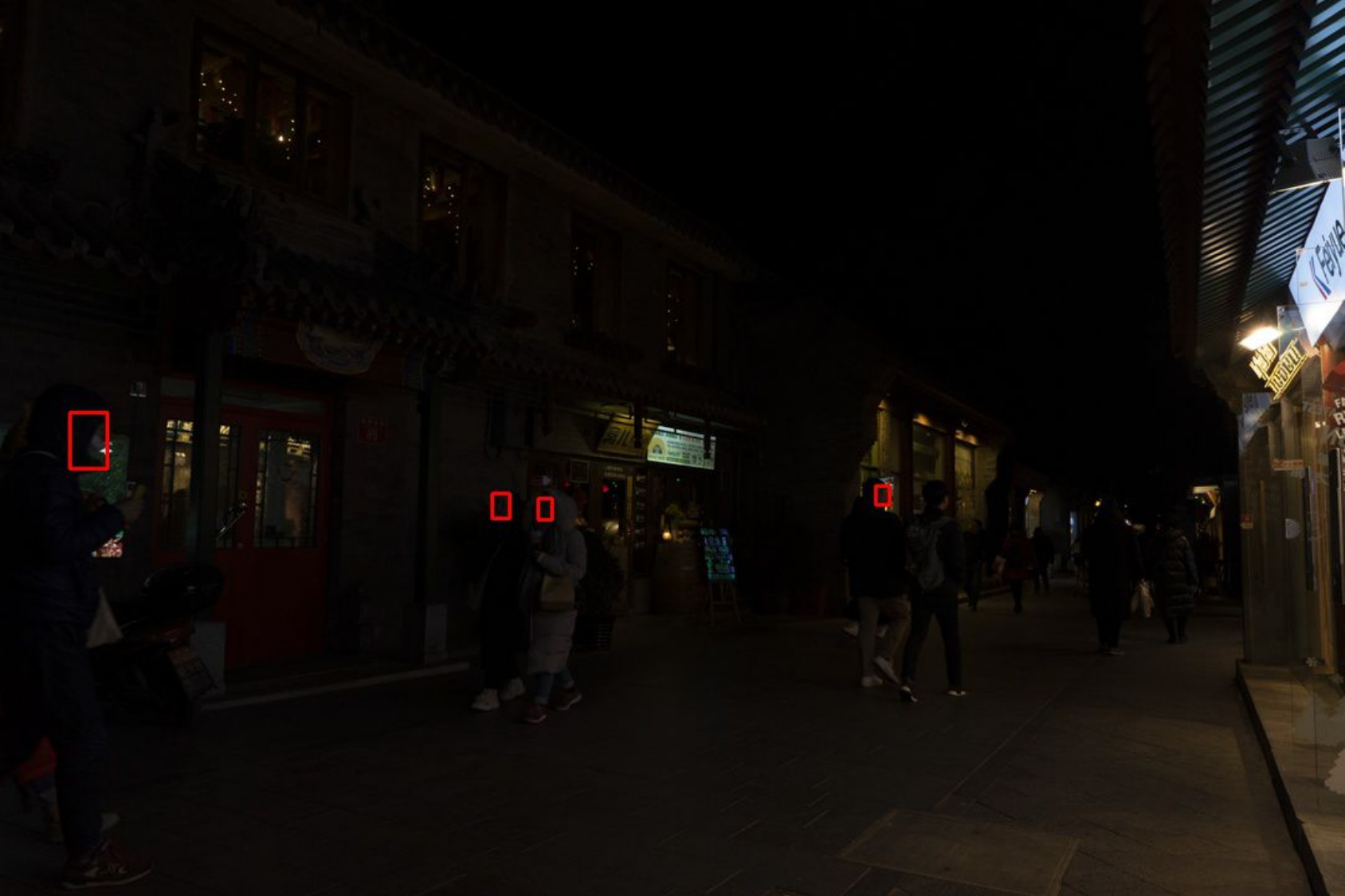}
}
\vspace{-0.2cm}
\subfigure[RetinaFace with PIE$_{det}$]{
\includegraphics[width=7cm]{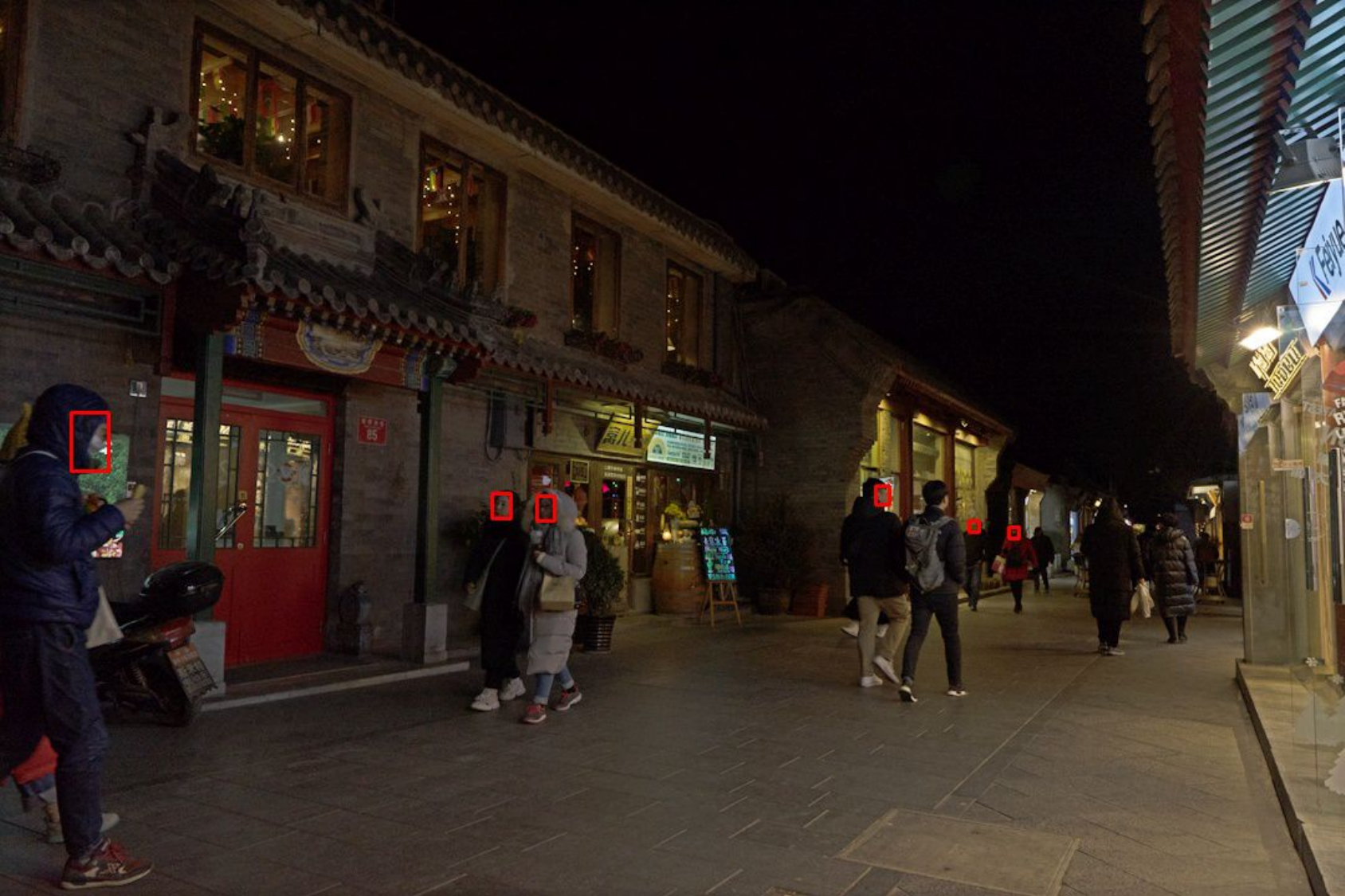}
}
\caption{Qualitative comparison of face detection without and with PIE. The detector is RetinaFace~\citep{deng2020retinaface}}
\label{face}
\end{figure}

\subsubsection{Discussion for contrastive learning}\label{secDIS}
\textbf{Discussion for Contrastive Loss}
Triplet loss~\citep{Hermans_Beyer_Leibe_2017}, N-pair loss~\citep{Sohn_2016}, and InfoNCE loss~\citep{Gutmann_Hyvärinen_2010} are commonly used in contrastive learning to pull the anchor closer to the positive sample and push it away from the negative sample in the latent feature space. In PIE, we apply contrastive losses, $L_{cG}$ and $L_{cE}$, respectively, to the gram matrix $G$ and the expectation $E$ to help the model learn features that represent positive samples and avoid features that represent negative samples. We find that applying different forms of contrastive losses to $L_{cG}$ and $L_{cE}$ results in different gains. Table~\ref{tab1eloss} shows the results of training with different loss forms for $L_{cG}$ and $L_{cE}$. We find that using triplet loss for $L_{cG}$ and infoNCE loss for $L_{cE}$ achieved the best results in terms of four metrics: NIQE, UNIQUE, PSNR, and SSIM. 
\textbf{Numbers of positive and negative samples}
We conduct further experiments to explore the effect of different rates between positive and negative samples. 
For positive samples, we randomly select them from the positive sample dataset. We use our PIE with the rate of 1:1 as the baseline, and all experimental settings are the same as before, except for the number of samples. As the batch size increases, the GPU memory size required for training will increase, and the training time required for training an epoch (\emph{i.e.}, the process of using all samples in the training set to train once) will also significantly increase. Considering these factors, we use at most 5 positive or negative samples. 

As shown in Table~\ref{tab1erate}, adding more negative samples resulted in better performance, while adding more positive samples led to worse results. We conjecture that this is due to the different positive patterns that confuse the low-light image and hinder its capability to learn useful patterns. For negative samples, using more samples helped the model move away from the poor patterns in the over/underexposed images. However, increasing the number of negative samples also increased the training time. When we train using the rate of 1:1, the time required to train an epoch is 73.6 minutes. When we train using the rate of 1:5, the time required to train an epoch increases to 131.3 minutes.
Therefore, in our experiments, we use the rate of 1:1, except for Table~\ref{tab1erate}.


\subsection{Gain for Downstream Tasks}
\label{Gain4Down}
\subsubsection{Semantic segmentation with PIE}
Current low-light image datasets lack semantic annotation, which makes it difficult to evaluate semantic segmentation performance before and after enhancement.
To address this issue, we use subsets of Frankfurt, Lindau, and Munster from the Cityscapes validation set. Additionally, we simulate low-light images with varying brightness levels using the standard positive Gamma transformation with a range of Gamma values.
The trends of mean intersection-over-union (mIoU) with the brightness of the scene are shown in Fig.~\ref{img9}. Among all the methods, the segmentation performance with our method tends to be the best when scenes become dark. In Fig.~\ref{seg}, our method effectively improves semantic segmentation performance compared with LLE state-of-the-art methods. These findings motivate us to explore ways to bridge the gap between current low-light enhancement methods and downstream tasks.

\subsubsection{Face Detection with PIE}
We use RetinaFace~\citep{deng2020retinaface} trained on the WIDER FACE dataset~\citep{yang2016wider} as the face detector. Two thousand images in the DARK FACE dataset~\cite{DARKFACE} are used as test input, and different methods are used to enhance them respectively. 
Then, we feed the results of different low-light image enhancement methods to RetinaFace. 
To evaluate the accuracy of the model, we compare the average precision (AP) under different IoU thresholds (0.5, 0.7, and 0.9). A target is considered detected when the IoU is greater than 50\%.
Table~\ref{FACE_D} shows the AP results, with a focus on the IoU threshold of 0.5. All low-light image enhancement methods except ISSR and RUAS improve the face detection performance on the dataset. However, when we set a higher IoU threshold, the AP scores of all methods decrease.
Our PIE method, which does not require paired training data, achieved a comparable score to the best result produced by LIME at the IoU threshold of 0.5, even without joint training with a face detection model. However, as mentioned earlier, LIME's subjective and quantitative results are not satisfactory. In contrast, our method produces better visual results. Our approach, PIE$_{det}$, achieves the best performance by training with a joint face detection model. Fig.~\ref{face} shows the comparison between the original detection and detection with our PIE.

\begin{table}[!htb]
\centering

\caption{The average precision (AP) for face detection in low-light conditions on the DARK FACE dataset  was evaluated using different IoU thresholds (0.5, 0.7, 0.9). }

\setlength{\tabcolsep}{1.2mm}{
\begin{tabular}{c|ccc}
\hline
 &\multicolumn{3}{|c}{IoU thresholds}\\ 
 Methods& 0.5& 0.7& 0.9\\ \hline
 Input&0.2820 &0.0693 &0.0002 \\
 LIME~\citep{guo2016lime} &0.4221 &0.1068 &\textbf{0.0004} \\
 RetinexNet~\citep{Chen2018Retinex}&0.3874 &0.1065 &0.0002 \\
 ISSR~\citep{FanWY020}&0.2825 &0.0674 &0.0001 \\
 Zero-DCE~\citep{Guo_2020_CVPR}&0.4130 &0.1067 &0.0002 \\
 EnlightenGAN~\citep{jiang2021enlightengan}&0.3762 &0.1009 &0.0003 \\
 RUAS~\citep{liu2021retinex}&0.2782 &0.0659 &0.0002 \\
 ReLLIE~\citep{Zhang2021ReLLIEDR}&0.3583 &0.0958 &0.0001 \\\hline
 PIE&0.4199 &0.1082 &0.0003 \\
 PIE$_{det}$&\textbf{0.4288} &\textbf{0.1104} &0.0003 \\
\hline
\end{tabular}}
\label{FACE_D}
\end{table}

\begin{table}[!htb]
\captionsetup{width=1\textwidth,justification=raggedright}
\caption{Experimental results of test-time cost comparison.}
\setlength{\tabcolsep}{1.2mm}{
\begin{tabular}{c|cc}
\hline
Method & GFLOPs $\downarrow$ &Runtime (s) $\downarrow$\\
\hline
LLNet~\citep{lore2017llnet}& 4124.17 &36.270 \\
MBLLEN~\citep{Lv2018MBLLEN}& 301.12 & 13.995\\
KinD++~\citep{zhang2021beyond}& 12238.02 & 1.068\\
Zero-DCE~\citep{Guo_2020_CVPR} &84.99 & \textbf{0.003}\\
EnlightenGAN~\citep{jiang2021enlightengan} &273.24 &0.008\\
ReLLIE~\citep{Zhang2021ReLLIEDR} &125.13 &1.480\\ 
LIME~\citep{guo2016lime}& (on CPU) & 21.530\\
Retinex-Net~\citep{Chen2018Retinex}& 587.470 &0.120 \\
ISSR~\citep{FanWY020}&  (unavailable) & 9.645\\
RUAS~\citep{liu2021retinex} &1.069 & 0.006\\
IRN~\citep{zhao2021deep} &12438.282 &4.216\\
SCI~\citep{ma2022toward}&\textbf{0.580} &0.010\\
ALL-E~\citep{li2023all}&113.85&1.055\\
SCL-LLE~\citep{liangaaai} &95.21 &0.004\\
\hline
PIE&85.54 &0.004\\
\hline
\end{tabular}}
\label{cost}
\end{table}

\begin{figure}[h!]
\centering
\subfigure[Input]{
\includegraphics[width=3.55cm]{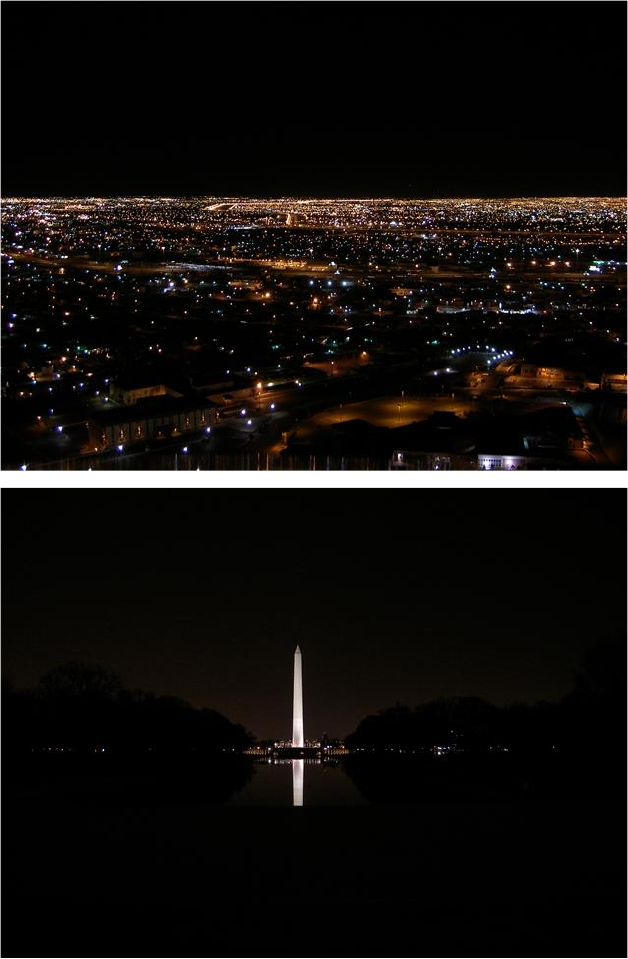}
}
\subfigure[LIME]{
\includegraphics[width=3.55cm]{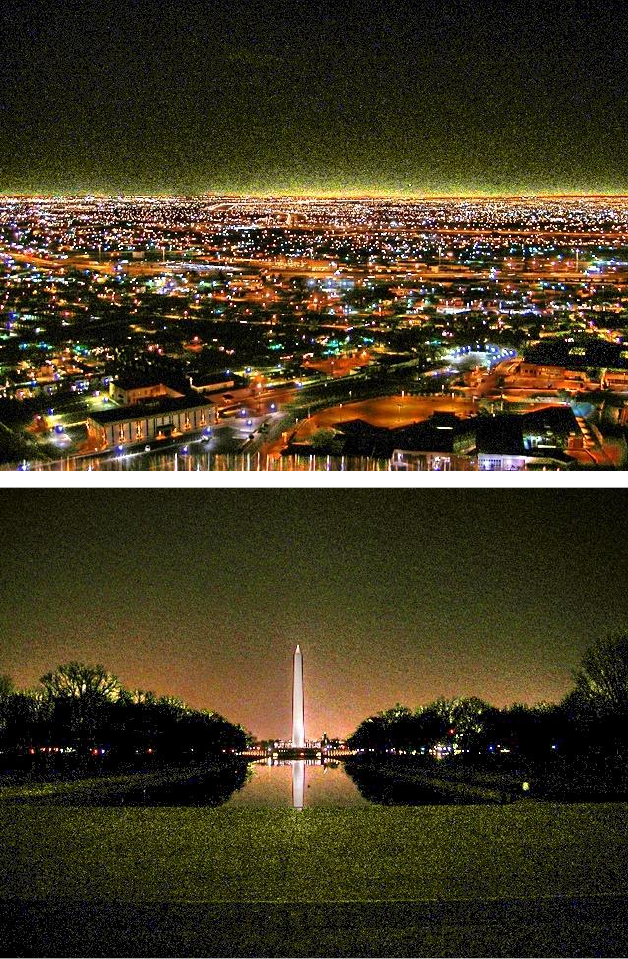}
}
\subfigure[EnlightenGAN]{
\includegraphics[width=3.55cm]{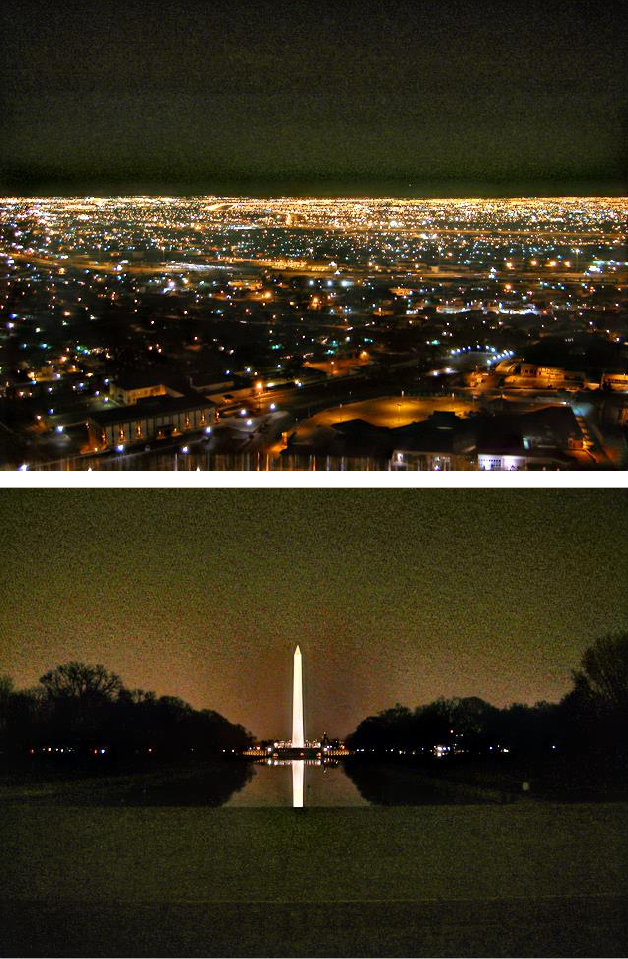}
}
\subfigure[\textbf{Ours}]{
\includegraphics[width=3.55cm]{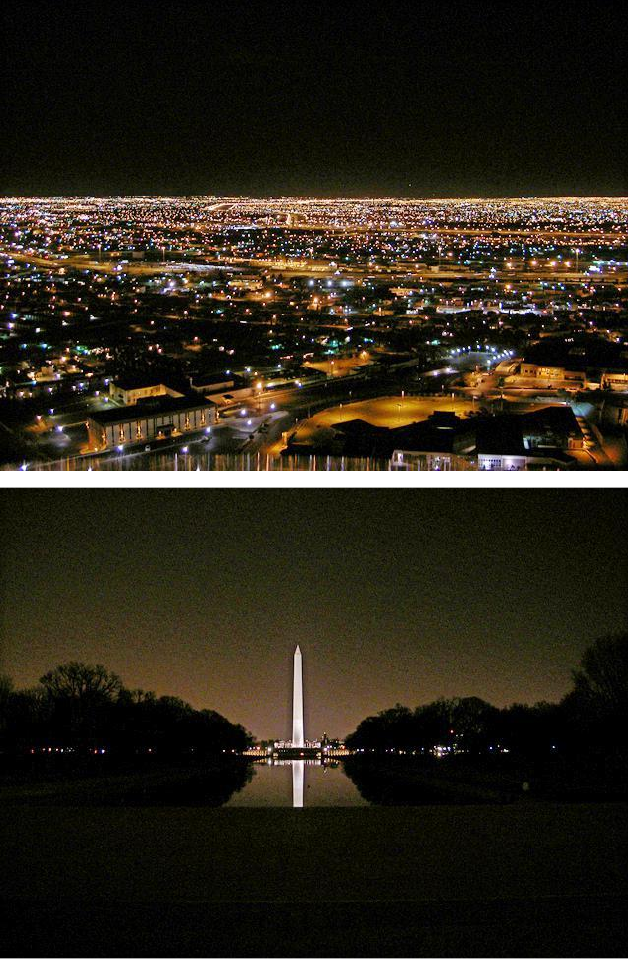}
}
\caption{The failure cases.
}
\label{fc}
\end{figure}

\subsection{Test-time Cost Comparison}
We compare the test runtime and computational costs (measured in Giga floating-point operations per second, GFLOPs) of our method with state-of-the-art methods (LLNet, MBLLEN, 
KinD++, Zero-DCE, EnlightenGAN, ReLLIE, LIME, Retinex-Net, ISSR, RUAS, IRN, SCI, ALL-E, SCL-LLE). All indicators are recorded with full processing for 32 images of size 1200 $\times$ 900 $\times$ 3 using an NVIDIA GTX1080Ti GPU.
As shown in Table~\ref{cost}, our method is only slightly slower than Zero-DCE (our backbone LLE Network), but much faster than most other methods excluding Zero-DCE in terms of runtime. This indicates that PIE has high speed and can quickly process a large number of images. 
Our method achieves a moderate level in terms of computational cost (GFLOPs).
\subsection{Failure Cases}
In Fig.~\ref{fc}, we showcase two scenarios in which the Performance-Improvement Enhancement algorithm encounters failures. These failures arise when the low-light image input contains a substantial amount of noise. Similar to other established approaches like LIME and EnlightenGAN, our method also exhibits the presence of background stripe noise in the enhanced images, which is influenced by the inherent noise in the original image. To address this issue, a module for removing stripe noise can be incorporated as a solution.

\section{Conclusion}
\label{Conclusion}
We propose physics-inspired contrastive learning for low-light image enhancement (PIE). This is achieved by introducing Bag of Curves in contrastive learning, which efficiently generates negative samples that mimic the Gamma correction and Tone mapping processes in the ISP pipeline. BoC generates under/overexposed images aligned with the underlying physical imaging principles. Additionally, the regional segmentation module is an unsupervised method that maintains regional brightness consistency and removes the dependence on semantic ground truths.
Extensive experiments demonstrate that our method outperforms the state-of-the-art LLE models on six independent cross-scene datasets. Furthermore, we conduct experiments combining LLE with semantic segmentation, object detection, and image classification, demonstrating that PIE benefits downstream tasks under extremely dark conditions. The proposed method runs fast with reasonable GFLOPs in test time, making it easy to use on mobile devices.

\section*{Acknowledgements}
 This work was partly supported by NSFC (Grant Nos. 62272229, 62076124, 62222605), the National Key R$\&$D Program of China (2020AAA0107000), the Natural Science Foundation of Jiangsu Province (Grant Nos. BK20222012, BK20211517), and Shenzhen Science and Technology Program JCYJ20230807142001004. The authors would like to thank all the anonymous reviewers for their constructive comnments.
\bibliography{sn-bibliography.bib}

\end{document}